%% file: main.tex
\documentclass[letterpaper,11pt]{article}

\usepackage{diagbox}
\usepackage{makecell}
\usepackage{booktabs}
\usepackage{tikz}
\usepackage{amsmath}
\usepackage{amsfonts}
\usepackage{amssymb}
\usepackage{amsthm}
\usepackage{url,ifthen}
\usepackage{enumitem}
\usepackage{srcltx}
\usepackage{multirow}
\usepackage{boxedminipage}
\usepackage[margin=1in]{geometry}
\usepackage{nicefrac}
\usepackage{xspace}
\usepackage{graphicx}
\usepackage{color}
\usepackage{colortbl}
\usepackage{setspace}
\usepackage{natbib}
\usepackage{subcaption}
\usepackage{dsfont}
\usepackage{authblk}
\usepackage{soul}
\setcitestyle{square, authoryear}
\definecolor{DarkGreen}{rgb}{0.1,0.5,0.1}
\definecolor{DarkRed}{rgb}{0.5,0.1,0.1}
\definecolor{DarkBlue}{rgb}{0.1,0.1,0.5}
\definecolor{Gray}{rgb}{0.2,0.2,0.2}

\usepackage{listings}
\lstdefinestyle{mystyle}{
    commentstyle=\color{DarkBlue},
    keywordstyle=\color{DarkRed},
    numberstyle=\tiny\color{Gray},
    stringstyle=\color{DarkGreen},
    basicstyle=\footnotesize,
    breakatwhitespace=false,         
    breaklines=true,                 
    captionpos=b,                    
    keepspaces=true,                 
    numbers=left,                    
    numbersep=5pt,                  
    showspaces=false,                
    showstringspaces=false,
    showtabs=false,                  
    tabsize=2
}
\usepackage[pdftex]{hyperref}
\hypersetup{
    unicode=false,          
    pdftoolbar=true,        
    pdfmenubar=true,        
    pdffitwindow=false,      
    pdfnewwindow=true,      
    colorlinks=true,       
    linkcolor=DarkBlue,          
    citecolor=DarkGreen,        
    filecolor=DarkRed,      
    urlcolor=DarkBlue,          
    pdftitle={},
    pdfauthor={},
}

\usepackage[utf8]{inputenc}
\usepackage[T1]{fontenc}
\usepackage{microtype}
\usepackage{csquotes}
\usepackage{mathpazo}
\RequirePackage{helvet}
\RequirePackage{beramono}
\RequirePackage{textcomp}

\usepackage{macros}

\usepackage{thmtools,thm-restate}

\theoremstyle{definition}

\newtheorem*{property*}{Property}

\usepackage{apptools}
\AtAppendix{\counterwithin{proposition}{section}}
\AtAppendix{\counterwithin{corollary}{section}}
\AtAppendix{\counterwithin{lemma}{section}}

\title{On the Out-Of-Distribution Generalization of Multimodal Large Language Models}

\author[1]{Xingxuan Zhang$^{\dag}$}

\author[1]{Jiansheng Li$^{\dag}$}
\author[2]{Wenjing Chu}
\author[2]{Junjia Hai}
\author[1]{Renzhe Xu}
\author[1]{Yuqing Yang}
\author[1]{Shikai Guan}
\author[1]{Jiazheng Xu}
\author[1]{Peng Cui*}

\affil[1]{Department of Computer Science, Tsinghua University}
\affil[2]{School of Computer and Information Technology, Beijing Jiaotong University}
\affil[ ]{}
\affil[ ]{\small \texttt{xingxuanzhang@hotmail.com, cuip@tsinghua.edu.cn}}

\newcommand{\picl}{P^{\text{ICL}}}
\newcommand{\ptest}{P^{\text{test}}}

\usepackage{color}
\usepackage{xcolor}

\makeatletter
\def\maketag@@@#1{\hbox{\m@th\normalfont\normalsize#1}}
\makeatother

\date{}

\begin{document}
\maketitle



\input{paragraphs/abstract.tex}

\renewcommand{\thefootnote}{\fnsymbol{footnote}}
\footnotetext[2]{Equal contribution}
\footnotetext[1]{Corresponding Author}

\input{paragraphs/intro}

\input{paragraphs/zero-shot}

\input{paragraphs/errer_analysis}
\input{paragraphs/ICL}

\input{paragraphs/app-related_works}
\input{paragraphs/app-exp-details}

\input{paragraphs/conclusion}


\clearpage

\appendix
\input{paragraphs/app-error-anly}
\input{paragraphs/app-showcase-zero-shot}

\input{paragraphs/app-showcase-icl}

\clearpage
\bibliographystyle{plainnat}
\bibliography{references}

\end{document}

%% file: paragraphs/abstract.tex
\begin{abstract}
    We investigate the generalization boundaries of current Multimodal Large Language Models (MLLMs) via comprehensive evaluation under out-of-distribution scenarios and domain-specific tasks. We evaluate their zero-shot generalization across synthetic images, real-world distributional shifts, and specialized datasets like medical and molecular imagery. Empirical results indicate that MLLMs struggle with generalization beyond common training domains, limiting their direct application without adaptation. To understand the cause of unreliable performance, we analyze three hypotheses: semantic misinterpretation, visual feature extraction insufficiency, and mapping deficiency. Results identify mapping deficiency as the primary hurdle. To address this problem, we show that in-context learning (ICL) can significantly enhance MLLMs' generalization, opening new avenues for overcoming generalization barriers. We further explore the robustness of ICL under distribution shifts and show its vulnerability to domain shifts, label shifts, and spurious correlation shifts between in-context examples and test data.
    
\end{abstract}

%% file: paragraphs/intro.tex
\section{Introduction}

Supported by the exponential growth of both data availability and computational prowess, the field of large language models (LLMs)~\citep{touvron2023llama,chowdhery2022palm} has witnessed a surge in interest and development, demonstrably achieving generalization across diverse natural language tasks~\citep{openai2023gpt4,zhu2023minigpt4}. 
Pushing the boundaries of general artificial intelligence further, the rise of Multimodal Large Language Models (MLLMs)~~\citep{alayrac2022flamingo,dai2023instructblip} aims to integrate LLM strengths with additional sensory modalities, such as image, audio, or 3D data, enabling them to condition their output on this broader range of inputs~\citep{team2023gemini,zhang2023llamaadapter}.
Some works extend LLMs with visual understanding through end-to-end tuning, such as GPT-4V~\citep{yang2023dawn}, Gemini~\citep{team2023gemini}, Flamingo~\citep{alayrac2022flamingo}, and Qwen~\citep{bai2023qwen}. Alternatively, fueled by powerful LLMs like LLaMA~\citep{touvron2023llama}, modular combinations of LLMs and image-to-text models such as LLaVA~\citep{liu2023visual}, MiniGPT-4~\citep{zhu2023minigpt}, InstructBLIP~\citep{dai2023instructblip} and many more~\citep{zhang2023llamaadapter,awadalla2023openflamingo} are also being explored. The burgeoning field of MLLMs has also witnessed a surge in benchmark development~\citep{fu2023mme,bai2023touchstone,lu2023mathvista}, which demonstrate MLLMs' intriguing ability to solve complex multimodal tasks like open-world recognition and multimodal commonsense reasoning. 

Despite the considerable progress in MLLMs, a comprehensive understanding of their generalization capabilities remains elusive, particularly when applied to specialized domains or confronted with distributional shifts~\citep{yuan2023revisiting}. 
Recently, researchers have highlighted the potential impact of distributional bias on the reliability of LLMs~\citep{wang2023robustness,yuan2023revisiting}. Notably, recent works~\citep{yang2023dawn,han2023well} demonstrate that GPT-4V, a state-of-the-art MLLM, exhibits susceptibility to generating erroneous outputs when encountering distribution shifts. Yet comprehensive evaluations of current MLLMs under out-of-distribution scenarios and on domain-specific tasks rarely addressed during training are scarce. Furthermore, existing studies often lack in-depth analysis and explanation of the underlying causes driving MLLMs' susceptibility to errors in these situations.
This knowledge gap hinders the practical implementation of MLLMs and necessitates further research efforts in diverse application areas and under varying data distributions. 

In this paper, we aim to delineate the boundaries of the current MLLMs' generalization beyond the confines of its training data distribution. We first evaluate the zero-shot generalization capabilities of MLLMs across diverse test data including synthetically generated images to probe model resilience beyond their training distribution, naturally occurring distribution shifts to simulate real-world data variations, and domain-specific imagery, including medical and molecular datasets, to assess the usability to specialized tasks. 
We find that current MLLMs struggle with zero-shot generalization beyond domains closely resembling their training data. Direct application of MLLMs to specific domains without further adaptation or fine-tuning is likely to
be insufficiently accurate and reliable. 

To uncover the root cause of unreliability, we conducted a comprehensive error analysis investigating three plausible hypotheses:
(1) Semantic misinterpretation: MLLMs may struggle to grasp the nuanced meaning and implications of specific scientific categories within prompts, harming their understanding of the task.
(2) Visual feature extraction insufficiency: The inherent characteristics of medical and molecular data, like high dimensionality or intricate image features, might challenge the model's internal encoding and processing mechanisms, hindering accurate visual information extraction.
(3) Mapping deficiency: Restricted training data in specialized domains hinders the development of robust mappings between semantic meanings and visual features in MLLMs.

Our analysis identifies that mapping deficiency can be the primary hindrance to model generalization. To address this limitation, we investigate the potential of in-context learning (ICL) to enhance the model's acquisition and utilization of critical relationships between semantic descriptions and visual features. Surprisingly, we find that incorporating in-context examples (ICE) not only from the target distribution but also those with domain shifts significantly enhances the generalization capabilities of MLLMs. This opens exciting avenues for using designed ICL to break through generalization barriers.
Furthermore, we show that some severe distribution shifts between ICE and test samples, including label and spurious correlation shifts, can introduce performance decline and instability rising. Thus promising data selection of ICE approaches or reinforcement of generalization is required for reliable applications of MLLMs on specific tasks.

The contribution of the paper is summarized as follows.

1. We evaluate the zero-shot generalization of 14 current multi-modal large language models (MLLMs) on 20 datasets under various distributional shifts. We demonstrate that the OOD generalization performance of MLLMs can significantly diverge from their performance on current public generic benchmarks. 

2. We investigate the scaling law of MLLMs on OOD generalization ability for the potential issues of visual feature extraction failure caused by high data complexity within visual input. 

3. Our analysis identifies that mapping deficiency other than semantic misinterpretation within text input and data complexity within visual input can be the primary hindrance to model generalization.

4. We validate the potential of ICL to enhance the model's acquisition and utilization of critical relationships between semantic descriptions and visual features with ICE both from the target distribution and biased distributions. 


%% file: paragraphs/zero-shot.tex
\section{Zero-shot Generalization}
\label{sec:zero-shot}
\begin{figure}[ht]
    \centering
    \includegraphics[width=0.8\linewidth]{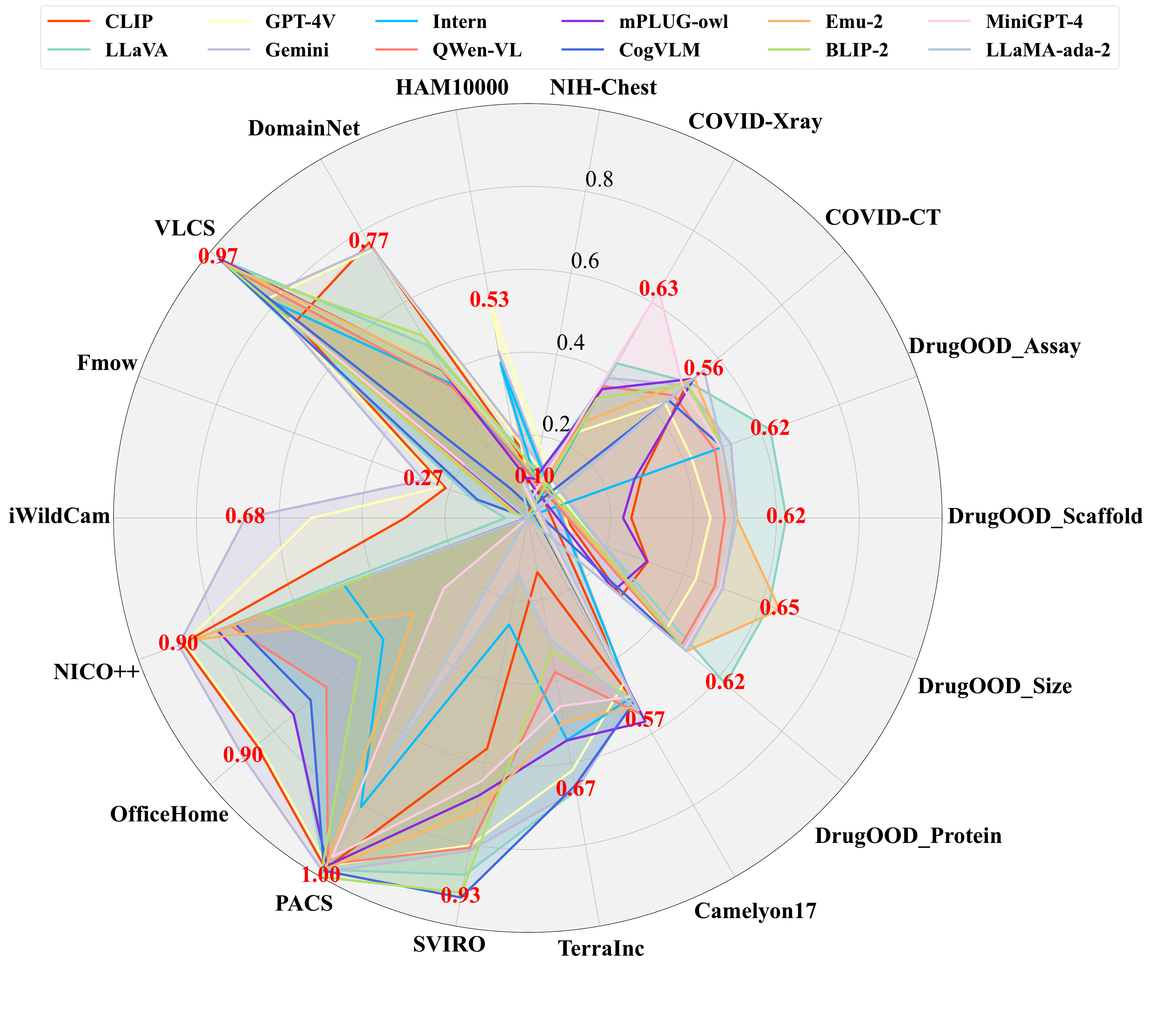}
    \caption{Zero-shot generalization of current MLLMs on OOD generalization or domain-specific tasks.}
    \label{fig:zero-shot}
\end{figure}

This section delves into the zero-shot generalization performance of MLLMs on visually shifted data encompassing three categories: synthetically generated images, naturally occurring distribution shifts, and domain-specific imagery like medical and molecular datasets. This investigation aims to elucidate the generalization limitations and strengths of MLLMs under distinct data variations. We consider MLLMs including LLaVA~\citep{liu2023visual},QWen~\citep{bai2023qwen}, CogVLM~\citep{wang2023cogvlm}, mPLUG-owl~\citep{ye2023mplug}, MiniGPT-4~\citep{zhu2023minigpt}, LLaMA-adapter V2~\citep{gao2023llama},  CLIP~\citep{radford2021learning}, BLIP~\citep{li2023blip}, InstructBLIP~\citep{dai2023instructblip},kosmos-2~\citep{peng2023kosmos}, Emu-2~\citep{sun2023generative},Intern~\citep{zhang2023internlm}, Gemini~\citep{team2023gemini} and GPT-4V~\citep{openai2023gpt4} for our experiments.

\subsection{Zero-shot Generalization on Synthetic and Natural Images}
\label{sec:zero-shot-common}
We present an evaluation of MLLMs on imagery with significant synthetic and natural shifts in data distribution. The observed performance provides insights into the models' capabilities for perceiving and understanding everyday objects and concepts, while also exposing potential limitations in generalization beyond controlled settings. We leverage a total of 11 datasets, each with its distinct characteristics and challenges: CMNIST~\citep{arjovsky2019invariant}, RMNIST~\citep{ghifary2015domain}, PACS~\citep{li2017deeper}, VLCS~\citep{fang2013unbiased}, OfficeHome~\citep{venkateswara2017deep}, TerraInc~\citep{beery2018recognition}, DomainNet~\citep{peng2019moment}, Fmow~\citep{christie2018functional}, iWildCam~\citep{beery2021iwildcam}, SVIRO~\citep{cruz2020sviro}, and NICO++~\citep{zhang2023nico++}. The introduction of these datasets can be found in Section \ref{app:exp-details}. 

\textbf{MLLMs achieve outstanding performance on most of the well-used datasets for domain generalization or OOD generalization, such as PACS, VLCS, OfficeHome, DomainNet, and NICO++.} 
The performance on these regular datasets approaches or even surpasses that of the existing optimal algorithms, suggesting that the generalization of these models to recognize common objects in the domains contained in these datasets is strong. This may be because these domains have been included in the training data for these large models.
The prediction accuracy of both open-source and API-accessible models including CLIP, LLaVA, QWen, mPLUG-owl, Emu-2, BLIP2, GPT-4V, and Gemini exceed 96\% on PACS and 80\% on VLCS, with the performance of LLaVA, QWen-VL, mPLUG-owl, and BLIP-2 exceed 90\% on VLCS.

Notably, the usability of current common object recognition is also evidenced by the exceptional performance of GPT-4V and Gemini on challenging benchmarks like DomainNet (74.8\% and 75.3\% accuracy, respectively) and iWildCam (52.3\% and 68.2\%). This surpasses the previous state-of-the-art approaches~\citep{rame2023model,cho2023promptstyler}.
The superior performance of GPT-4V and Gemini on benchmarks with numerous categories (DomainNet: 345, iWildCam: 206) suggests these models excel at handling complex prompts.

\begin{table*}[th]
\centering
\caption{Zero-shot generalization results of current state-of-the-art MLLMs on OOD generalization datasets.}
\resizebox{\textwidth}{!}{
\begin{tabular}{lccccccccccc|c}
\toprule
Dataset & CMNIST & RMNIST& DomainNet & Fmow & iWildCam & NICO++ & OfficeHome & PACS & SVIRO & TerraInc & VLCS & Average\\ 

\midrule
LLaVA & 0.622& 0.696& 0.480 & 0.148 & 0.054 & 0.849 & 0.736 & 0.980 & 0.874 & 0.668 & 0.975 &  0.644 \\
QWen-VL & 0.198 & 0.212 & 0.365 & 0.038 & 0.000 & 0.763 & 0.635 & 0.964 & 0.808 & 0.378 & 0.943 & 0.482 \\
CogVLM & 0.117 & 0.317 & 0.081 & 0.013 & 0.000 & 0.066 & 0.684 & 0.983 & 0.930 & 0.656 & 0.965 & 0.437  \\
mPLUG-owl & 0.356& 0.411& 0.412 & 0.020 & 0.003 & 0.796 & 0.738 & 0.969 & 0.680 & 0.545 & 0.972 & 0.537 \\
MiniGPT-4 & 0.285& 0.294& 0.000& 0.062& 0.000 & 0.631& 0.266 & 0.945 & 0.645 & 0.461 & 0.750 & 0.394\\
LLaMA-adapter V2 & 0.314& 0.252& 0.389& 0.022& 0.003& 0.394 & 0.000 & 0.690 & 0.137 & 0.294 & 0.197 & 0.245\\
CLIP & 0.468& 0.446& 0.767 & 0.211 & 0.296 & 0.887 & 0.854 & 0.977 & 0.565 & 0.133 & 0.734 & 0.576 \\
BLIP-2 & 0.3392& 0.376& 0.508 & 0.043 & 0.000 & 0.673 & 0.527 & 1.000 & 0.918 & 0.328 & 0.937 & 0.514 \\
InstructBLIP & 0.5098& 0.6112& 0.477 & 0.020& 0.000 & 0.483& 0.003 & 0.723 & 0.000 & 0.376 & 0.257 & 0.315 \\
kosmos-2 & 0.000& 0.000& 0.227 & 0.035 & 0.000 & 0.163 & 0.000 & 0.792 & 0.413 & 0.365 & 0.569 & 0.233 \\
Emu-2 & 0.622 & 0.668 & 0.415 & 0.027 & 0.000 & 0.860 & 0.360 & 0.977 & 0.720 & 0.500 & 0.957 & 0.555 \\
Intern & 0.198 & 0.285 & 0.375 & 0.015 & 0.000 & 0.472 & 0.456 & 0.805 & 0.261 & 0.545 & 0.822 & 0.385 \\
Gemini & 0.729& 0.933& 0.753 & 0.268 & 0.682 & 0.897 & 0.897 & 0.987 & 0.816 & 0.668 & 0.832 & 0.769 \\
GPT-4V & 0.646& 0.469& 0.748& 0.220 & 0.523 & 0.880 & 0.848 & 0.969 & 0.802 & 0.619 & 0.872 & 0.691 \\

\bottomrule
\end{tabular}
}
\label{tab:domainbed-datasets}
\end{table*}

\subsection{Zero-shot Generalization on Domain-Specific Data}

We consider medical and molecular images for the evaluation of domain-specific generalization ability for the following reasons. 1) Analyzing MLLMs' performance across these two disparate domains, with their unique knowledge requirements and data distributions, provides valuable insights into their generalization capabilities and potential for broader real-world applications. 2) They help investigate the potential of incorporating auxiliary inputs, such as domain-specific information and in-context examples, for enhancing the model's performance in specific application areas. By systematically evaluating the efficacy of these additional inputs, we can gain valuable insights into the factors influencing model generalization and identify promising avenues for further improvement.
3) Failure analysis on these tasks can highlight areas where further development and domain-specific adaptation might be necessary. We consider diverse medical and molecular image datasets including Camelyon17~\citep{sun2022camelyon}, CT-XCOV~\citep{elbouknify2023ct}, XCOVFour~\citep{han2021semi}, HAM10000~\citep{tschandl2018ham10000}, NIH-Chest~\citep{national2017nih}, DrugOOD\_Assay~\citep{ji2022drugood}, DrugOOD\_Scaffold~\citep{ji2022drugood}, DrugOOD\_Size~\citep{ji2022drugood}, and DrugOOD\_Protein~\citep{ji2022drugood}.

While MLLMs excel at zero-shot generalization on OOD datasets of synthetic and natural images, their performance plummets when transitioning to medical and molecular data. 
Specifically, all MLLMs make near-random predictions on all medical and molecular datasets including Camelyon17, CT-XCOV, XCOVFour, HAM10000, NIH-Chest, DrugOOD\_Assay, DrugOOD\_Scaffold, DrugOOD\_Size, and DrugOOD\_Protein. Please note that the average number of categories in these domain-specific datasets is significantly lower than that of common object datasets in Section \ref{sec:zero-shot-common}, e.g., there are only 2 categories in Camelyon17, CT-XCOV, DrugOOD\_Assay, DrugOOD\_Scaffold, DrugOOD\_Size, and DrugOOD\_Protein. 

\textbf{This significant drop in performance suggests the crucial fact that MLLMs struggle with zero-shot generalization beyond domains closely resembling their training data.} Thus direct application of these models to medical and molecular domains without further adaptation or fine-tuning is insufficiently accurate and reliable. We study the underlying causes of the observed limitations of MLLM generalization on specific domains and potential solutions to enhance MLLM adaptability and robust performance across diverse domains in Section \ref{sec:error-analysis} and \ref{sec:icl}.

\begin{table*}[th]
\centering
\caption{Zero-shot generalization results of MLLMs on domain-specific datasets.}
\resizebox{\textwidth}{!}{
\begin{tabular}{lccccccccc|c}
\toprule
Dataset & Camelyon17 & HAM10000 & NIH-Chest & XCOVFour & CT-XCOV & DrugOOD\_A & DrugOOD\_Sc & DrugOOD\_Si & DrugOOD\_P & Average\\ 

\midrule
LLaVA & 0.508 & 0.363 & 0.089 & 0.431 & 0.509 & 0.623 & 0.624 & 0.623 & 0.620 & 0.488 \\
QWen-VL & 0.546 & 0.095 & 0.042 & 0.367 & 0.460 & 0.481 & 0.476 & 0.481 & 0.481 & 0.381 \\
CogVLM & 0.513 & 0.061 & 0.066 & 0.040 & 0.443 & 0.500 & 0.500 & 0.500 & 0.500 & 0.347 \\
mPLUG-owl & 0.568 & 0.101 & 0.096 & 0.359 & 0.523 & 0.274 & 0.230 & 0.306 & 0.272 & 0.303 \\
MiniGPT-4 & 0.499 & 0.090 & 0.048 & 0.633 & 0.489 & 0.496 & 0.502 & 0.495 & 0.493 & 0.416 \\
LLaMA-adapter V2 & 0.536 & 0.117 & 0.000 & 0.000 & 0.555 & 0.499 & 0.500 & 0.500 & 0.500 & 0.356 \\
CLIP & 0.502 & 0.219 & 0.050 & 0.000 & 0.500 & 0.292 & 0.250 & 0.308 & 0.294 & 0.268 \\
BLIP-2 & 0.503 & 0.073 & 0.006 & 0.333 & 0.503 & 0.500 & 0.500 & 0.500 & 0.500 & 0.380 \\
kosmos-2 & 0.189 & 0.012 & 0.000 & 0.000 & 0.000 & 0.000 & 0.000 & 0.000 & 0.000 & 0.022 \\
Emu-2 & 0.525 & 0.103 & 0.047 & 0.267 & 0.525 & 0.500 & 0.504 & 0.647 & 0.500 & 0.402 \\
Intern & 0.500 & 0.376 & 0.098 & 0.328 & 0.008 & 0.500 & 0.500 & 0.500 & 0.500 & 0.368 \\
Gemini  & 0.502 & 0.410 & 0.078 & 0.390 & 0.498 & 0.522 & 0.505 & 0.499 & 0.489 & 0.433 \\
GPT-4V  & 0.462 & 0.530 & 0.057 & 0.239 & 0.432 & 0.421 & 0.441 & 0.432 & 0.430 & 0.383 \\

\bottomrule
\end{tabular}
}
\label{tab:Medical-images}
\end{table*}


%% file: paragraphs/errer_analysis.tex
\section{Failure Analysis}
\label{sec:error-analysis}

We consider three potential causes of the failure on domain-specific data: 
(1) Semantic understanding failure: MLLMs may struggle to grasp the nuanced meaning and implications of specific scientific concepts within prompts, leading to semantic misinterpretation within text inputs.
(2) Visual feature extraction failure: The inherent characteristics of medical and molecular data, such as high dimensionality and intricate image features, can challenge the model's encoding and processing mechanisms, hampering accurate visual information extraction.
(3) Mapping deficiency: limited training data in specialized domains restricts the development of robust and generalizable mappings between semantic meanings and visual features.

\paragraph{Potential shared biases among MLLMs.}
To uncover potential shared biases arising from model architecture and training methodologies in various MLLMs, we begin with the analysis of individual MLLM error patterns across diverse datasets. We present an overall analysis of error in Appendix \ref{app-error} and extensive case studies of zero-shot generalization of MLLMs in Appendix \ref{sec:app-casestudy}.
We reveal a lack of consistent, dataset-specific biases. Instead, errors appeared scattered across most datasets. 
This suggests that despite similar architectural and training characteristics, different MLLMs do not exhibit consistent biases in semantic or visual feature extraction, such as persistently associating specific semantic features with incorrect visual features. 
Moreover, this observation raises the intriguing possibility of constructing multimodal models by strategically combining MLLMs to leverage their diverse error profiles and achieve enhanced generalization capabilities.

\paragraph{Semantic misinterpretation.}
To analyze the impact of semantic misinterpretation within text input on model failure, we use informative and distinct prompts that provide additional scientific context and guidance for each dataset. These prompts aim to address the potential limitations of MLLMs in understanding specific meanings and implications of certain categories within the input prompt, particularly in specialized scientific contexts. 
Specifically, to enhance model adaptation and domain awareness with text information, we crafted introductory prompts for each dataset. These prompts comprised three key elements: (1) domain specification and context, outlining the relevant field and study goals; (2) expected expertise, guiding the model towards domain-specific knowledge; and (3) category-specific introductions, providing clear explanations of each category's meaning and potential characteristics. Please see Section \ref{app-new-prompt} for details of the prompt design.

Table \ref{tab:new_prompt} shows that augmenting prompts with detailed domain-specific background information yields negligible performance improvements for MLLMs in these tasks. This suggests that: (1) semantic misinterpretation within text input might not be the main limiting factor, or (2) the model's internal mechanisms may be inadequate for efficiently integrating and leveraging the auxiliary information provided. Given the sophisticated language understanding capabilities of LLMs like GPT-4 and Gemini, a simple inability to comprehend the provided auxiliary information seems improbable as the sole explanation for their performance limitations in these domains. \textbf{Therefore, semantic understanding failure is unlikely to severely affect MLLMs generalization.}

\begin{table*}[th]
\centering
\caption{Prompt contextual information results of current state-of-the-art MLLMs on domain-specific datasets. Performance differences versus the common prompt are indicated parenthetically, with positive values in red reflecting improvements in accuracy and negative values in green indicating drops.}
\resizebox{\textwidth}{!}{
\begin{tabular}{lccccccccc|c}
\toprule
Dataset & Camelyon17 & HAM10000 & NIH-Chest & XCOVFour & CT-XCOV & DrugOOD\_A & DrugOOD\_Sc & DrugOOD\_Si & DrugOOD\_P & Average\\ 

\midrule
LLaVa & 0.555\textcolor{red}{(+4.7\%)} & 0.621\textcolor{red}{(+25.8\%)} & 0.097\textcolor{red}{(+0.8\%)} & 0.327\textcolor{teal}{(-10.4\%)} & 0.333\textcolor{teal}{(-17.6\%)} & 0.501\textcolor{teal}{(-12.2\%)} & 0.529\textcolor{teal}{(-9.5\%)} & 0.504\textcolor{teal}{(-11.9\%)} & 0.512\textcolor{teal}{(-10.8\%)} & 0.442\textcolor{teal}{(-4.6\%)} \\
QWen-VL & 0.523\textcolor{teal}{(-2.3\%)} & 0.101\textcolor{red}{(+0.6\%)} & 0.078\textcolor{red}{(+3.6\%)} & 0.329\textcolor{teal}{(-3.8\%)} & 0.500\textcolor{red}{(+4.0\%)} & 0.547\textcolor{red}{(+6.6\%)} & 0.502\textcolor{red}{(+2.6\%)} & 0.538\textcolor{red}{(+5.7\%)} & 0.504\textcolor{red}{(+2.3\%)} & 0.402\textcolor{red}{(+2.1\%)} \\
CogVLM & 0.500\textcolor{teal}{(-1.3\%)} & 0.179\textcolor{red}{(+11.8\%)} & 0.000\textcolor{teal}{(-6.6\%)} & 0.100\textcolor{red}{(+6.0\%)} & 0.500\textcolor{red}{(+5.7\%)} & 0.500\textcolor{red}{(+0.0\%)} & 0.001\textcolor{teal}{(-49.9\%)} & 0.647\textcolor{red}{(+14.7\%)} & 0.629\textcolor{red}{(+12.9\%)} & 0.340\textcolor{teal}{(-0.7\%)} \\
MiniGPT-4 & 0.502\textcolor{red}{(+0.3\%)} & 0.223\textcolor{red}{(+13.3\%)} & 0.000\textcolor{teal}{(-4.8\%)} & 0.008\textcolor{teal}{(-62.5\%)} & 0.486\textcolor{teal}{(-0.3\%)} & 0.000\textcolor{teal}{(-49.6\%)} & 0.000\textcolor{teal}{(-50.2\%)} & 0.000\textcolor{teal}{(-49.5\%)} & 0.000\textcolor{teal}{(-49.3\%)} & 0.135\textcolor{teal}{(-28.1\%)} \\
LLaMa-adapter V2 & 0.439\textcolor{teal}{(-9.7\%)} & 0.004\textcolor{teal}{(-11.3\%)} & 0.000\textcolor{red}{(+0.0\%)} &0.022\textcolor{red}{(+2.2\%)} & 0.441\textcolor{red}{(+44.1\%)} & 0.500\textcolor{teal}{(-5.5\%)} & 0.500\textcolor{red}{(+0.1\%)} & 0.500\textcolor{red}{(+0.0\%)} & 0.500\textcolor{red}{(+0.0\%)} & 0.363\textcolor{red}{(+0.7\%)} \\
BLIP-2 & 0.500\textcolor{teal}{(-0.3\%)} & 0.004\textcolor{teal}{(-6.9\%)} & 0.010\textcolor{red}{(+4.0\%)} & 0.333\textcolor{red}{(+0.0\%)} & 0.500\textcolor{teal}{(-0.3\%)} & 0.500\textcolor{red}{(+0.0\%)} & 0.500\textcolor{red}{(+0.0\%)} & 0.500\textcolor{red}{(+0.0\%)} & 0.500\textcolor{red}{(+0.0\%)} & 0.372\textcolor{teal}{(-0.8\%)}\\
Emu-2 & 0.575\textcolor{red}{(+5.0\%)} & 0.098\textcolor{teal}{(-0.5\%)} & 0.000\textcolor{teal}{(-4.7\%)}  & 0.333\textcolor{red}{(+6.6\%)}  & 0.500\textcolor{teal}{(-2.5\%)}  & 0.717\textcolor{red}{(+21.7\%)}  & 0.473\textcolor{teal}{(-3.1\%)}  & 0.504\textcolor{teal}{(-14.3\%)}  & 0.512\textcolor{red}{(+1.2\%)}  & 0.489\textcolor{red}{(+8.7\%)} \\
Intern & 0.500\textcolor{red}{(+0.0\%)} & 0.000\textcolor{teal}{(-37.6\%)} & 0.087\textcolor{teal}{(-1.1\%)}  & 0.205\textcolor{teal}{(-12.3\%)}  & 0.500\textcolor{red}{(+0.0\%)} & 0.500\textcolor{red}{(+0.0\%)} & 0.496\textcolor{teal}{(-0.4\%)}  & 0.500\textcolor{red}{(+0.0\%)} & 0.500\textcolor{red}{(+0.0\%)} & 0.365\textcolor{teal}{(-0.8\%)} \\
Gemini  & 0.506\textcolor{red}{(+0.3\%)} & 0.410\textcolor{red}{(+0.0\%)} & 0.104\textcolor{red}{(+2.6\%)} & 0.427\textcolor{red}{(+3.7\%)} & 0.697\textcolor{red}{(+19.9\%)} & 0.534\textcolor{red}{(+1.2\%)} & 0.506\textcolor{red}{(+0.1\%)} & 0.528\textcolor{red}{(+2.9\%)} & 0.519\textcolor{red}{(+3.0\%)} & 0.470\textcolor{red}{(+3.7\%)} \\
GPT-4V  & 0.430\textcolor{teal}{(-3.2\%)}  & 0.291\textcolor{teal}{(-23.9\%)}  & 0.000\textcolor{teal}{(-5.7\%)}  & 0.100\textcolor{teal}{(-13.9\%)}  & 0.534\textcolor{red}{(+10.2\%)}  & 0.050\textcolor{teal}{(-37.1\%)}  & 0.023\textcolor{teal}{(-41.8\%)}  & 0.060\textcolor{teal}{(-37.2\%)}  & 0.295\textcolor{teal}{(-13.5\%)} & 0.198\textcolor{teal}{(-18.5\%)} \\

\bottomrule
\end{tabular}
}
\label{tab:new_prompt}
\end{table*}

\paragraph{Visual feature extraction insufficiency.}
To analyze the impact of data complexity within visual input, we investigate the sufficiency of visual extraction capabilities in MLLMs for generalizing to specific domains. We employ a linear probing approach, utilizing the publicly available CLIP model as a feature extractor. Extracted features from CLIP are then fed into a separate linear classifier trained with corresponding category labels. By analyzing the classifier's accuracy on unseen test data, we assess the quality of visual features extracted by MLLMs for generalization.

Table \ref{tab:linear_prob} shows the performance of CLIP with linear probing, zero-shot CLIP, GPT-4V, and Gemini on challenging specific datasets, where CLIP with linear probing shows markedly superior performance. Since CLIP has a weaker visual feature extraction capability compared to most modern MLLMs, \textbf{the results suggest that the visual feature extraction is unlikely the bottleneck hindering MLLMs generalization to these specific tasks.}

\begin{table}[th]
\centering
\caption{The comparison of CLIP with linear probing (donated as CLIP(LP)) and zero-shot generalization of CLIP, GPT-4V, and Gemini.}
\resizebox{0.65\linewidth}{!}{
\begin{tabular}{lcccc}
\toprule
 & CLIP (LP) & CLIP & GPT-4V & Gemini\\ 

\midrule
COVID-CT & 0.830 & 0.500 (\textcolor{teal}{-33 .0\%}) & 0.432 (\textcolor{teal}{-47.95\%}) & 0.498 (\textcolor{teal}{-40.00\%}) \\
DrugOOD\_Assay & 0.760 & 0.292 (\textcolor{teal}{-61.58\%}) & 0.421 (\textcolor{teal}{-44.61\%}) & 0.505 (\textcolor{teal}{-33.55\%}) \\
DrugOOD\_Size & 0.771 & 0.308 (\textcolor{teal}{-60.05\%}) & 0.432 (\textcolor{teal}{-43.97\%}) & 0.499 (\textcolor{teal}{-35.28\%}) \\
Fmow & 1.000 & 0.211 (\textcolor{teal}{-78.90\%}) & 0.220 (\textcolor{teal}{-78.00\%}) & 0.268 (\textcolor{teal}{-73.20\%}) \\
HAM10000 & 0.840 & 0.219 (\textcolor{teal}{-73.93\%}) & 0.530 (\textcolor{teal}{-36.90\%}) & 0.410 (\textcolor{teal}{-51.19\%}) \\
NIH-Chest & 0.740 & 0.050 (\textcolor{teal}{-93.24\%}) & 0.057 (\textcolor{teal}{-92.30\%}) & 0.078 (\textcolor{teal}{-89.46\%}) \\
COVID-Xray & 0.970 & 0.000 (\textcolor{teal}{-97.00\%}) & 0.239 (\textcolor{teal}{-75.36\%}) & 0.390 (\textcolor{teal}{-59.79\%}) \\

\bottomrule
\end{tabular}
}

\label{tab:linear_prob}
\end{table}

\paragraph{The scaling of zero-shot generalization.}

We explore the potential of enhanced input encoding capabilities in large language models to quantify and potentially mitigate the limitations in generalization arising from insufficient encoding capacity of both text and visual inputs. 
Extensive research has established the performance gains achieved through scaling neural networks, often governed by reliable scaling laws related to training data, model size, and computational resources\citep{openai2023gpt4,cherti2023reproducible}. This understanding proves invaluable in mitigating the escalating cost of large-scale experiments. However, existing scaling laws primarily apply to in-distribution performance or general tasks, neglecting OOD generalization. We investigate the scaling laws for MLLMs under OOD scenarios, shedding light on their behavior in real-world contexts. 

Due to the limitations in publicly available model sizes for other MLLMs, we focus on CLIP as a representative model to investigate the scalability of zero-shot generalization. Specifically, we explore CLIP models with different sizes of ViT architecture. We evaluate the performance of these CLIP models across five datasets: DomainNet, Terra Incognita, Office-Home, NICO, and SVIRO. These datasets were chosen based on their suitability for showcasing effective CLIP performance while minimizing potential data leakage biases, thus maximizing the reliability of our conclusions regarding the effectiveness of scaling.

\begin{figure}[ht]
    \centering
    \includegraphics[width=0.6\linewidth]{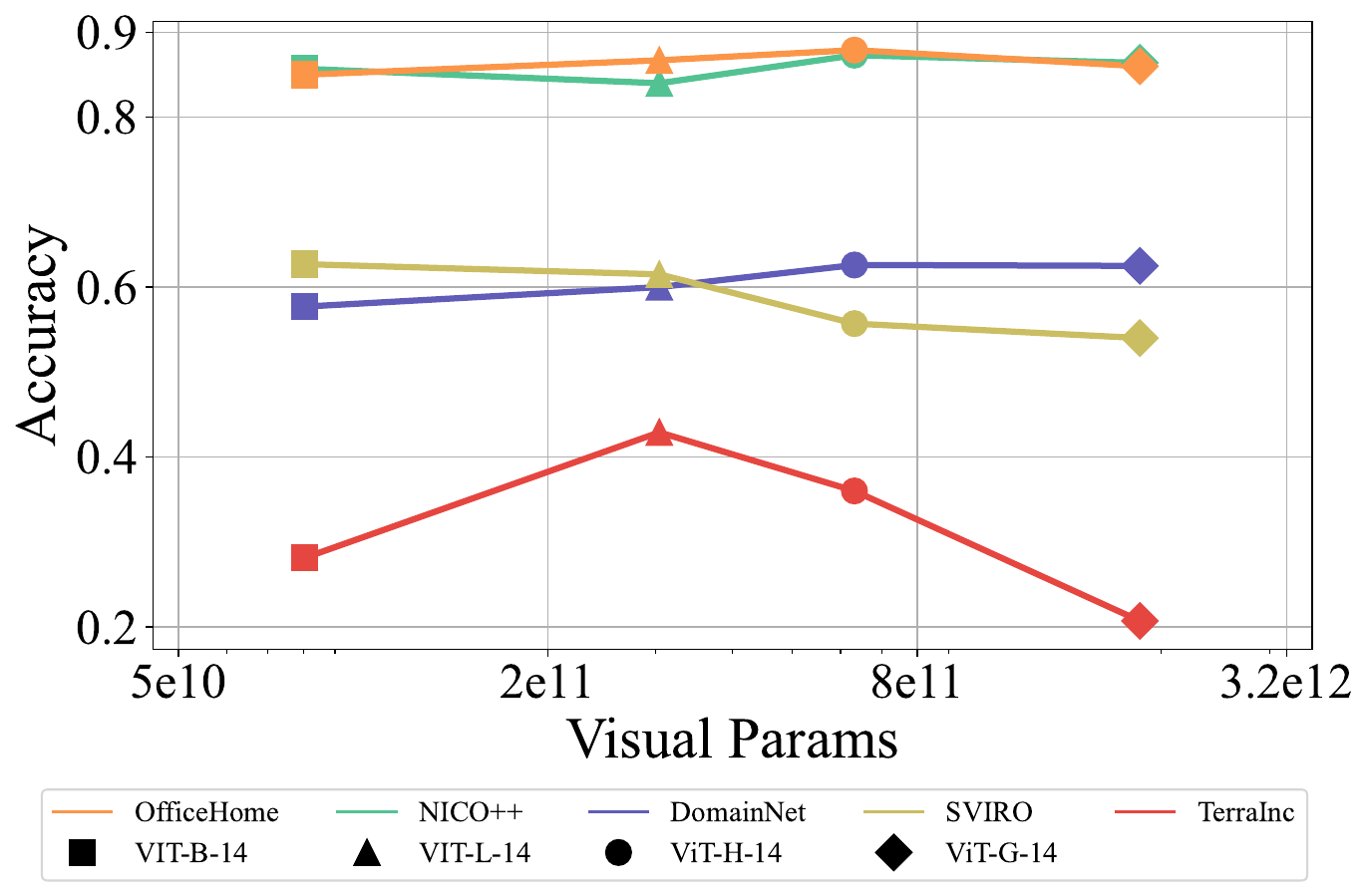}
    \caption{Scaling law of zero-shot generalization under distribution shifts.}
    \label{fig:scaling}
\end{figure}

Unlike the typical scaling law observed on in-distribution (ID) tasks, where larger models with enhanced encoding capabilities tend to perform better, Figure \ref{fig:scaling} reveals a different trend for OOD tasks, even exhibiting a notable decline on TerraInc and SVIRO. \textbf{This suggests that simply scaling up model size and improving its in-domain encoding skills may not be sufficient to guarantee improved generalization ability in domain-specific scenarios.}

We then analyze the impact of mapping deficiency and potential solutions with in-context learning in Section \ref{sec:icl}.

%% file: paragraphs/ICL.tex
\section{In-context Learning Generalization}
\label{sec:icl}
Building upon the failure analysis in Section \ref{sec:error-analysis}, this section investigates the potential of in-context learning (ICL) to bridge the gap between MLLMs and domains not represented in their pre-training and fine-tuning data by addressing the mapping deficiency.

ICL has recently garnered significant attention as a potential paradigm shift for large vision models~\citep{bar2022visual,zhang2023makes}. It shows its ability to adapt to new tasks or distributions without updating model parameters by simply prepending domain-specific input-output pairs (in-context examples, ICE) to the test example. This augmented input acts as a guiding framework, directing the model towards delivering desired outputs for previously unseen tasks. Thus ICL offers a flexible and efficient mechanism for continual adaptation without incurring the computational costs of retraining. 

The following sub-sections study the potential of ICL for enhancing MLLM generalization by exploring two avenues: 1) Utilizing target domain data directly to inform the selection and construction of ICE. Here we explore the extent to which explicitly incorporating information about the target domain within the ICE can guide the model to adapt to new tasks within the domain.
2) Employing ICE that differ from the target domain data. 
This focuses on evaluating the effectiveness of ICL when bridging the gap between known and target data distributions. In other words, we aim to assess the model's performance when the distribution of test data is not precisely known when designing ICE.

\paragraph{Notations.} 
Let $X$ denote the inputs and $Y$ denote the outputs. Suppose the test distribution is $\ptest(X, Y)$ and ICE $\{(x_i, y_i)\}_{i=1}^n$ are sampled from a distribution $\picl(X, Y)$.

\subsection{ICL with Data from Target Distribution}
\label{subsec: icl with iid data}
This subsection explores the effectiveness of ICL to MLLMs in an ideal scenario: when examples are sampled directly from the target distribution, i.e., $\picl(X, Y) = \ptest(X, Y)$. 
This simulates a setting where the target domain's data distribution is known and readily available for crafting ICE. By investigating the model's performance under such optimal conditions, we aim to determine the upper bound of potential performance improvements achievable through ICL. This theoretical understanding can inform real-world applications by illuminating the limitations and optimal capabilities of this approach when adapting MLLMs to specific domains.

Given the previously observed underperformance of MLLMs on specific domain-specific datasets, we select challenging and diverse datasets for our ICL experiments. 
This choice prioritizes datasets where MLLMs have demonstrated limitations, allowing us to investigate the potential of ICL to bridge this performance gap. We analyze GPT-4V and Gemini due to their superior ICL capabilities observed in our preliminary experiments, which enables the isolation and analysis of the effectiveness of ICL without confounding factors arising from models' inherent limitations in understanding the guidance provided within such examples.

To further control for potential confounding factors like prompt understanding and category imbalance, we implemented a simplified experimental design for evaluating ICL's impact on MLLM performance on new domains and tasks. Across all datasets, we first randomly partitioned both domain and category into two balanced groups for the test data, ensuring an equal number of samples per category. This controlled setting mitigated the influence of category imbalance and potentially improved the consistency and reliability of our conclusions. We then investigate the effectiveness of ICE by systematically varying their number (0, 2, 4, and 8). Notably, the ICE were drawn from each category within the chosen domain with equal probability, mirroring the balanced class distribution observed in the test data. This stratified sampling approach ensures that the ICE can be considered independently and identically distributed (i.i.d.), statistically resembling the expected test data distribution.

\begin{figure*}[t]
    \centering
    \begin{subfigure}[b]{0.49\linewidth}
        \centering
        \includegraphics[width=\linewidth]{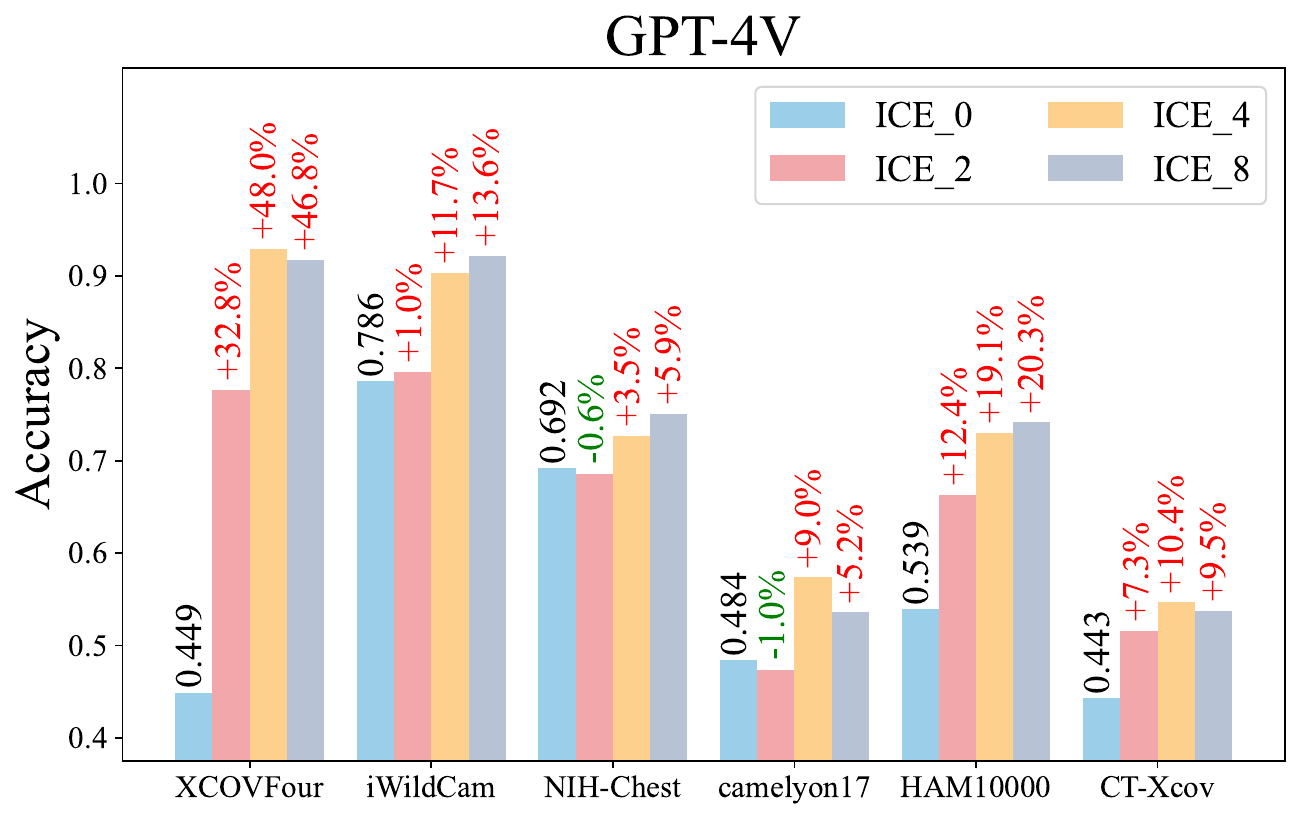}
        \label{fig:icl_iid_gpt4}
    \end{subfigure}
    \hfill
    \begin{subfigure}[b]{0.49\linewidth}
        \centering
        \includegraphics[width=\linewidth]{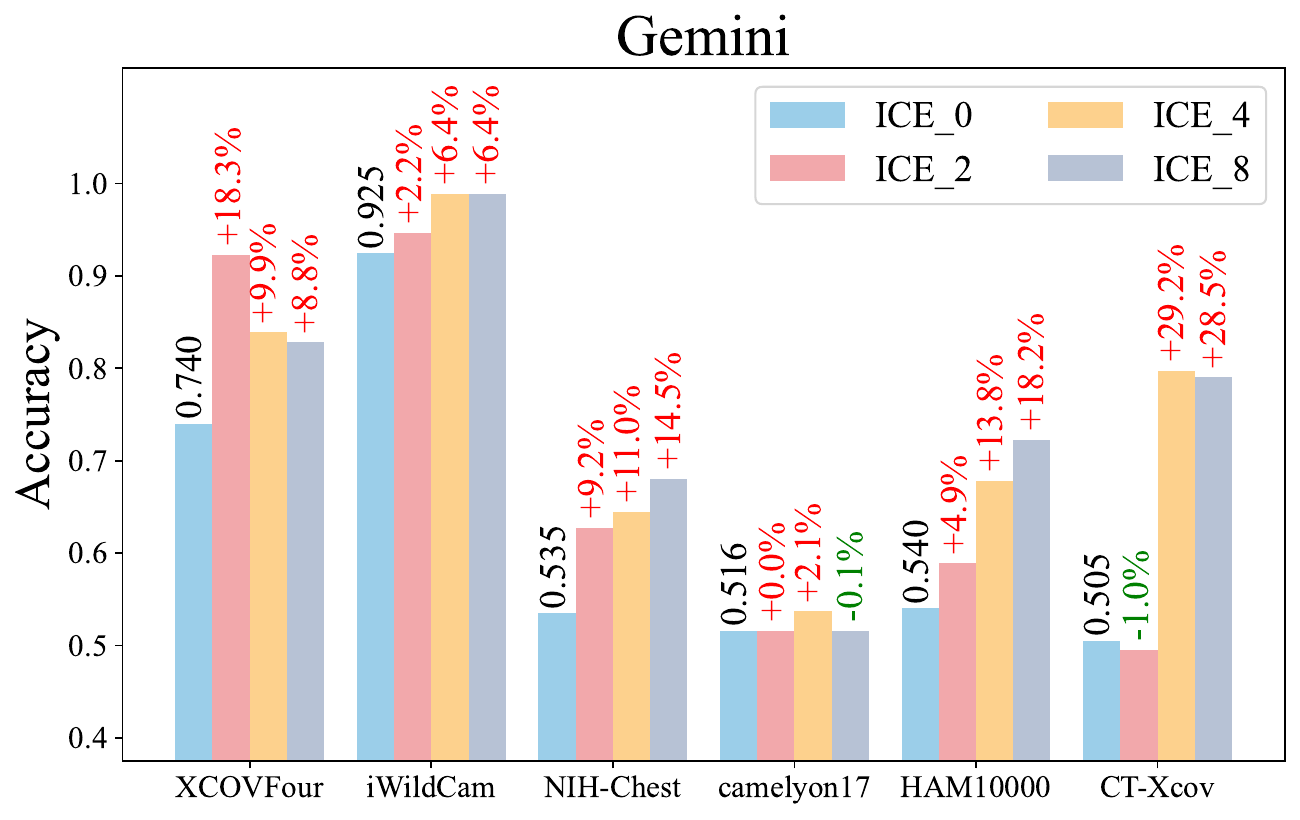}
        \label{fig:icl_iid_gemini}
    \end{subfigure}
    \caption{ICL Generalization of GPT-4V and Gemini with ICE from the target distribution.}
    \label{fig:icl_iid}
\end{figure*}

Both GPT-4V and Gemini exhibited substantial performance gains in ICL across XCOVFour, iWildCam, NIH-Chest, and HAM10000 datasets. This trend consistently held, with increasing numbers of ICE correlating with improved model performance, as shown in Figure \ref{fig:icl_iid}. Notably, on iWildCam, GPT-4V achieved a 14.6\% performance boost with just 2 ICE and a 36.6\% boost with 8. Similarly, on HAM10000, both models demonstrate performance improvements, with GPT-4V achieving 12.5\% (2 examples) and 20.3\% (8 examples), while Gemini achieve a 4.9\% (2 examples) and 18.2\% (8 examples) increase. 
On Camelyon17 and DomainNet, while ICL does not significantly enhance Gemini's performance, GPT-4V exhibits consistent performance gains and achieves up to 9\% performance improvement on Camelyon17 and 4.2\% improvement on DomainNet. 
On CT-XCOV, while GPT-4V's performance with ICL plateaued, Gemini exhibited a marked improvement, demonstrating a substantial peak gain of 29.2\%. 

\textbf{These findings demonstrate that when ICE closely mirror the distribution of test data, MLLMs can achieve substantial performance gains on many new tasks, where the degree of improvement scales with the number of examples provided. This further suggests that the primary obstacle to generalization in these tasks likely stems from deficiencies in task-specific knowledge mapping within the MLLMs.}

We show the whole results of GPT-4V and Gemini after ICL with data from target distributions in Table \ref{tab:app-icl-1}.
Intriguingly, both GPT-4V and Gemini failed to achieve performance gains on molecular datasets. 
This may be due to the inherent complexity of predicting molecular activity. Accurately classifying a molecule as ``active" or not often requires extensive domain knowledge and substantial training data, which mere ICE might not adequately provide. This observation suggests that ICL may not be a universally effective approach, particularly for tasks demanding specialized knowledge beyond readily learnable patterns. 
Further investigation is needed to elucidate the specific factors hindering its efficacy in such cases and potentially refine ICL strategies for tasks requiring domain expertise or introduce other methods to improve the adaptation of models such as retrieval-augmented generation (RAG)~~\citep{lewis2020retrieval}.

\begin{table}[th]
\centering
\caption{Results of ICL with data from target distribution.}
\resizebox{\linewidth}{!}{
\begin{tabular}{lccccccccc}
\toprule
& XCOVFour & iWildCam & NIH-Chest & CT-XCOV & DrugOOD\_Protein & DrugOOD\_Size & DrugOOD\_Assay & Camelyon17 & HAM10000 \\ 

\midrule
\multirow{4}{*}{\rotatebox{90}{GPT-4V}} & 0.449 & 0.634 & 0.692 & 0.443 & 0.459 & 0.282 & 0.447 & 0.484 & 0.539 \\
& 0.777 & 0.780 & 0.686 & 0.516 & 0.468 & 0.157 & 0.139 & 0.474 & 0.663 \\
& 0.929 & 0.923 & 0.727 & 0.547 & 0.468 & 0.247 & 0.253 & 0.574 & 0.730\\
& 0.917 & 1.000 & 0.751 & 0.438 & 0.468 & 0.287 & 0.272 &  0.536 & 0.742 \\

 \midrule
 \multirow{4}{*}{\rotatebox{90}{Gemini}}& 0.740 & 0.925 & 0.535 & 0.505 & 0.496 & 0.557 & 0.503 & 0.516 & 0.540 \\
& 0.923 & 0.946 & 0.627 & 0.495 & 0.560 & 0.467 & 0.396 & 0.516 & 0.589 \\
& 0.839 & 0.989 & 0.645 & 0.797 & 0.568 & 0.495 & 0.437 & 0.537 & 0.678 \\
& 0.828 & 0.989 & 0.680 & 0.790 & 0.579 & 0.489 & 0.453 & 0.515 & 0.722 \\

\bottomrule
\end{tabular}
}
\label{tab:app-icl-1}
\end{table}

\subsection{ICL under Distribution Shifts}
This subsection investigates the generalization of MLLMs with ICL under realistic conditions where the target data distribution may be unknown or deviate from that of the ICE. We focus on three key challenges: (1) Domain shifts: ICE and test samples are drawn from distinct domains defined by each dataset. (2) Label shifts: While the underlying domain remains, the distribution of labels differs between ICE and test data. (3) Spurious correlation shifts: If ICE contain misleading correlations not present in the target domain, the model may be prone to biased decisions, indicating the careful design of ICE is needed to avoid such pitfalls.
Please note that these three types of shifts are not orthogonal to each other, but they all represent the common and effective utilization of MLLMs in real-world applications. 
While real-world domain shifts may potentially encompass both label and spurious correlations shifts, our experiments confine such shifts within the predefined domains of the dataset. This limitation precludes isolating and quantifying the individual effects of each factor on model generalizability. Thus we study label and spurious correlation shifts through distinct analyses tailored to their specific characteristics.

\subsubsection{ICL with Domain Shifts}
We leverage the inherent domain diversity within existing datasets to investigate ICL's efficacy under realistic conditions. Prior research has established that domain shifts pose significant challenges for generalization~\citep{gulrajani2020search}. Therefore, we deliberately construct scenarios where ICE and test samples are drawn from different domains within these datasets, replicating natural or artificially induced distribution shifts, i.e., $\picl(X,Y)$ and $\ptest(X,Y)$ are the distributions of different domains predefined in each dataset, respectively.  
This is the basic setting for domain generalization (DG) and out-of-distribution (OOD) generalization tasks, which enables us to observe how effectively ICL can bridge these gaps and enhance MLLM generalization to unknown domains.
In the following experiments, we sample ICE and test samples from 2 domains within each dataset.
For datasets with multiple domains, we randomly sample 2 of them with adequate diversity of categories and sufficient samples.

\begin{figure*}[t]
    \centering
    \includegraphics[width=\linewidth]{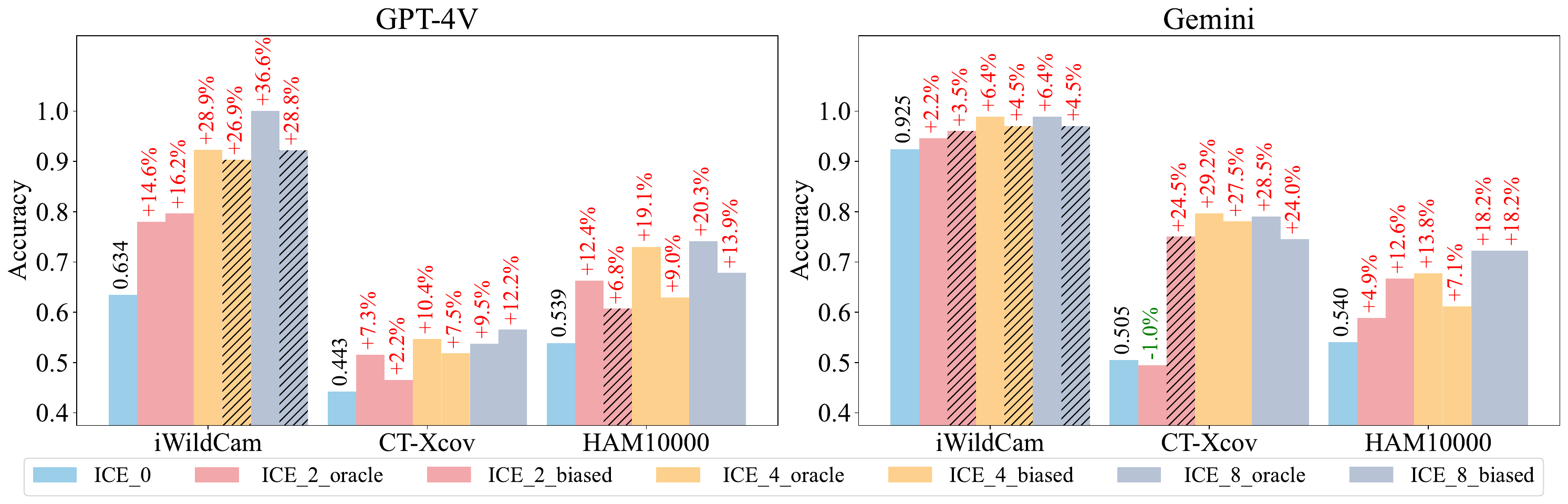}
    \caption{Generalization of GPT-4V and Gemini with ICL under domain shifts. ICE\_2\_oracle indicates ICL with 2 ICE from target distribution, while ICE\_2\_biased indicates 2 ICE that have domain shifts with target distribution.}
    \label{fig:icl_domain}
\end{figure*}

As shown in Figure \ref{fig:icl_domain}, ICE with domain shifts consistently significantly outperform zero-shot performance and show surprisingly robust performance, only marginally falling short of the gains achieved from examples directly drawn from the target distribution. This observation highlights the remarkable OOD generalization capabilities of MLLMs with ICL in traditional OOD settings when presented with non-ideal examples.

Specifically, across datasets including iWildCam, DomainNet, CT-XCOV, and HAM10000, ICL under domain shifts significantly bolsters both GPT-4V and Gemini's performance, as shown in Figure \ref{fig:icl_domain}. We also observe a consistent trend of increasing accuracy with the number of provided examples. Notably, on iWildCam, GPT-4V achieves a remarkable 16.2\% performance improvement with just 2 ICE, reaching a 28.8\% improvement with 8 examples. Similarly, Gemini demonstrably benefits from ICL on CT-XCOV, exhibiting a 24.5\% increase with 2 examples and a further 3\% with 4. On HAM10000, GPT-4V and Gemini achieve 13.9\% and 18.2\% improvement with 8 examples, respectively.

\textbf{While ICL with domain shifts slightly underperforms compared to target-distribution examples, the performance remains indistinguishable and suggests a remarkable capacity of MLLMs to generalize across domains via ICL.}

\subsubsection{ICL with Label Shifts}
We study another common challenge for generalization: label shift~~\citep{azizzadenesheli2018regularized}. In this scenario, the ground-truth output (labels) between ICE and the test data exhibit distributional differences. Specifically, this means that $\picl(Y) \ne \ptest(Y)$ while $\picl(X|Y) = \ptest(X|Y)$. Unlike domain shifts, which alter the entire underlying data structure, label shifts primarily affect the output label space.  Here we investigate whether the model's output preference can be influenced by ICE, leading to potentially misleading predictions. Specifically, both ICE and test samples are sampled from the same domain of each dataset with different distributions on $Y$.

\begin{table}[th]
\centering
\caption{ICL Generalization under label shifts.}
\resizebox{0.5\textwidth}{!}{
\begin{tabular}{lcccc}
\toprule
& Ratio & XCOVFour & NIH-Chest & CT-XCOV \\ 

\midrule
\multirow{5}{*}{\rotatebox{90}{GPT-4V}} & 1:7 & 0.908\textcolor{teal}{(-0.9\%)} & 0.719\textcolor{teal}{(-3.8\%)} & 0.445\textcolor{teal}{(-9.5\%)} \\
& 2:6 & 0.820\textcolor{teal}{(-9.7\%)} & 0.707\textcolor{teal}{(-5.0\%)} & 0.471\textcolor{teal}{(-6.9\%)} \\
& 4:4 & 0.917\textcolor{red}{(+0.0\%)} & 0.757\textcolor{red}{(+0.0\%)} & 0.540\textcolor{red}{(+0.0\%)} \\
& 6:2 & 0.958\textcolor{red}{(+4.1\%)} & 0.675\textcolor{teal}{(-8.2\%)} & 0.502\textcolor{teal}{(-3.8\%)} \\
& 7:1 & 0.950\textcolor{red}{(+3.3\%)} & 0.663\textcolor{teal}{(-9.4\%)} & 0.486\textcolor{teal}{(-5.4\%)} \\

 \midrule
\multirow{5}{*}{\rotatebox{90}{Gemini}} & 7:1 & 0.699\textcolor{teal}{(-12.9\%)} & 0.603\textcolor{teal}{(-7.7\%)} & 0.742\textcolor{teal}{(-4.8\%)} \\
& 6:2 & 0.699\textcolor{teal}{(-12.9\%)} & 0.627\textcolor{teal}{(-5.3\%)} & 0.732\textcolor{teal}{(-5.8\%)} \\
& 4:4 & 0.828\textcolor{red}{(+0.0\%)} & 0.680\textcolor{red}{(+0.0\%)} & 0.790\textcolor{red}{(+0.0\%)} \\
& 2:6 & 0.745\textcolor{teal}{(-8.3\%)} & 0.676\textcolor{teal}{(-0.4\%)} & 0.659\textcolor{teal}{(-13.1\%)} \\
& 1:7 & 0.732\textcolor{teal}{(-9.6\%)} & 0.641\textcolor{teal}{(-3.9\%)} & 0.717\textcolor{teal}{(-7.3\%)} \\

\bottomrule
\end{tabular}
}
\label{tab:label-shift}
\end{table}

Results are shown in Table \ref{tab:label-shift}. Changing the proportion of categories in the context demonstrably alters the predicted category proportions, subsequently influencing overall performance. Maintaining a fixed test data distribution while varying the in-context categories can lead to declining performance and rising instability. \textbf{This suggests that label shifts in real-world applications can significantly impact MLLM generalizability. Therefore, the judicious selection of ICE considering both the task specifics and the desired target data distribution is crucial for stable model performance.}

\subsubsection{ICL with Spurious Correlation Shifts}
We explore the impact of spurious correlations on ICL's efficacy. Spurious correlations arise when training data across domains exhibits statistically unreliable associations between input features and labels. Specifically, we divide the features $X$ into invariant $X_v$ and spurious $X_s$ components, as outlined in~\citet{lin2022zin, arjovsky2019invariant}. The output $Y$ depends solely on the invariant features $X_v$ in both $\picl$ and $\ptest$ distributions, represented as $Y = g(X_v, \epsilon)$, with $g(X_v, \epsilon)$ being the labeling function on $X_v$ and $\epsilon$ as exogenous noise. The spurious correlation shift is characterized by the difference $\picl(Y|X_s) \ne \ptest(Y|X_s)$.

These misleading correlations between $Y$ and $X_s$, often superficially stronger than invariant dependencies between $Y$ and $X_v$, can be readily acquired by models, hindering their ability to generalize beyond the training data's specific domain context. 
Specifically, we leverage datasets containing two distinct domains ($D_1$ and $D_2$) and two categories ($C_1$ and $C_2$). For ICE, we deliberately craft examples with contrasting domain-category combinations: $C_1$ data from $D_1$ and $C_2$ data from $D_2$. Conversely, test samples mirror the opposite scenario, presenting $C_2$ data from $D_1$ and $C_1$ data from $D_2$. This controlled setting ensures that ICE and test samples exhibit opposing domain-category associations, allowing us to isolate the influence of spurious correlation shifts between ICE and test samples on MLLM generalization beyond other distribution shifts. 

The results, as illustrated in Table \ref{tab:spurious-correlation}, support that spurious correlations can negatively impact the effectiveness of ICL. Case in point, GPT-4V exhibits a marked drop in prediction accuracy across both CMNIST and DomainNet when subject to ICL. Similar observations are evident in Gemini's experiments on iWildCam and DomainNet. \textbf{This indicates that although ICL can contribute to generalization compared with zero-shot prediction in some scenarios, the presence of significant spurious correlations within ICE can pose a substantial threat to the generalization of MLLMs on domains not included in their pre-training data. Thus promising data selection approaches or reinforcement of generalization against spurious correlation is required for reliable applications of MLLMs on these tasks.}

\begin{table}[th]
\centering
\caption{Generalization with ICL under spurious correlation shifts. ICE indicates the number of ICE.}
\resizebox{0.65\linewidth}{!}{
\begin{tabular}{lcccccc}
\toprule
& ICE & CMNIST & DomainNet & iWildCam & Camelyon17 & HAM10000 \\ 

\midrule
\multirow{4}{*}{\rotatebox{90}{GPT-4V}} & 0 & 0.896 & 0.850 & 0.800 & 0.500 & 0.490  \\
 & 2 & 0.490 & 0.730 & 0.860 & 0.530 & 0.530  \\ 
 & 4 & 0.784 & 0.930 & 0.810 & 0.550 & 0.520  \\
 & 8 & 0.803 & 0.970 & 0.880 & 0.470 & 0.510  \\

 \midrule
 \multirow{4}{*}{\rotatebox{90}{Gemini}} & 0 & 0.469 & 0.939 & 0.929  & 0.510 & 0.587  \\
 & 2 & 0.641 & 0.919 & 0.786 & 0.520 & 0.626  \\ 
 & 4 & 0.563 & 0.940 & 0.854 & 0.521 & 0.643   \\
 & 8 & 0.583 & 0.860 & 0.936 & 0.541 & 0.646  \\ 

\bottomrule
\end{tabular}
}
\label{tab:spurious-correlation}
\end{table}

%% file: paragraphs/app-related_works.tex
\section{Related Works} \label{sect:related-works}

\paragraph{Multimodal Large Language Model.}
Recent advancements in Large Language Models (LLMs) have demonstrated remarkable generality and performance ~\citep{brown2020language,openai2023gpt4,chowdhery2022palm,touvron2023llama}, which has motivated researchers to investigate Multimodal Large Language Models (MLLMs) that effectively and cohesively combine various visual and linguistic modalities in addressing intricate and diverse multimodal tasks.
A prevailing methodology in this domain entails integrating vision encoders (e.g., ~\citep{dosovitskiy2021an}) within the architecture of established LLMs, exemplified by LLaMA-adapter ~\citep{zhang2023llamaadapter} and LLaVA ~\citep{liu2023improved, liu2023visual}.
Another prominent research avenue involves extending the capabilities of LLMs by incorporating multi-sensory perception, such as GPT-4V~\citep{yang2023dawn}, Gemini~\citep{team2023gemini}, Frozen ~\citep{tsimpoukelli2021multimodal}, Flamingo~\citep{alayrac2022flamingo}, PaLM-E ~\citep{driess2023palme}, and Qwen~\citep{bai2023qwen}
Concurrently, open-source LLMs ~\citep{zhang2022opt,touvron2023llama,peng2023instruction} has spurred numerous projects, such as OpenFlamingo ~\citep{awadalla2023openflamingo}, MiniGPT-4 ~\citep{zhu2023minigpt4}, Otter ~\citep{li2023otter} and InstructBLIP ~\citep{dai2023instructblip}. The rapid advancement of multi-modal large language models (MLLMs) has been accompanied by a proliferation of new benchmarks~\citep{fu2023mme,bai2023touchstone,lu2023mathvista}. These benchmarks serve to evaluate and illuminate the increasingly sophisticated capabilities of LLMs, particularly their ability to tackle complex multimodal tasks such as open-world recognition and multimodal commonsense reasoning. While recent studies~\citep{yang2023dawn,han2023well} highlight the vulnerability of GPT-4V to specific tasks and distribution shifts, a more comprehensive understanding of their robustness and generalizability remains elusive. Current research largely overlooks thorough evaluations of MLLMs in out-of-distribution scenarios and on domain-specific tasks outside their training domains. Furthermore, existing studies often lack in-depth analysis and explanation of the underlying causes driving MLLMs' susceptibility to errors in these situations. 

\paragraph{In-context Learning.}
The recent progress in LLMs has also highlighted their remarkable capability for in-context learning (ICL) ~\citep{brown2020language, chowdhery2022palm, ahuja2023closer}, enabling them to effectively adapt to novel tasks through the utilization of a limited number of contextually-relevant examples.                     
This emergent phenomenon has been strategically leveraged to tackle intricate challenges like mathematical reasoning ~\citep{wei2023chainofthought}, thereby manifesting previously unseen capabilities within the models' behavioral repertoire ~\citep{wei2022emergent}.
Initial forays into ICL in vision and multimodality include Flamingo ~\citep{alayrac2022flamingo}, which fuses visual inputs with LLMs for in-context adaptation to visual-linguistic tasks via language interfaces, and early investigations by SegGPT ~\citep{wang2023seggpt} and Painter ~\citep{wang2023images} that explore this capability in visual domains.

Meanwhile, a multitude of studies have meticulously examined the effect of the inductive biases imbued during the pretraining phase of a model on its ICL capabilities ~\citep{chan2022transformers}, alongside delving into the key determinants shaping such inductive biases. 
For example, 
~\citet{wei2023larger} recent work asserts the equal significance of model size, revealing that substantially scaled models, exemplified by PaLM-540B, uniquely demonstrate the capability to selectively supersede inherent semantic priors when required, a capability that remains elusive to smaller-sized analogues. ~\citet{kirsch2024generalpurpose} finds that state size is a more decisive factor influencing the inductive bias pertinent to ICL performance over model size. 

\paragraph{Out-of-distribution Generalization.}
Out-of-distribution (OOD) generalization constitutes a pivotal area of research aiming to enhance the model's aptitude for generalizing to hitherto unseen domains ~\citep{ghifary2015domain, koh2021wilds}. A prevailing strategy involves the extraction of domain-invariant or causally informative features across numerous source domains ~\citep{hu2019domain, Piratla2020EfficientDG, Seo2019LearningTO, Zhang2021DeepSL, Xu2021WhySL, Zhang2021TowardsPD, zhang2023free}, or alternatively, aggregating domain-specific modules to build adaptive architectures ~\citep{Mancini18, Mancini2018RobustPC}. Notably, augmenting the input space through data augmentation techniques during training has also proven effective for OOD generalization ~\citep{Carlucci2019DomainGB, Qiao2020LearningTL, Xu2021AFF, Zhou2020DeepDI, Zhou2020LearningTG}. Furthermore, researchers have employed regularization methods coupled with meta-learning ~\citep{Dou2019DomainGV, Li2019EpisodicTF} and adopted the Invariant Risk Minimization (IRM) framework ~\citep{Arjovsky2019InvariantRM} to further bolster generalization performance.
Several studies have explored the concept of ensemble modeling, suggesting the aggregation of diverse model weights to optimize generalization capabilities ~\citep{Arpit2021EnsembleOA, Cha2021SWADDG, Chu2022DNADG, Li2023SIMPLESM, Ram2022RecyclingDM, Ram2022DiverseWA, Wortsman2022ModelSA}, and garnered considerable attention. More recently, several works have shown that flatness-aware optimization improves the generalization of deep models ~\citep{foret2020sharpness,zhang2023gradient,Zhang_2023_ICCV}.

%% file: paragraphs/app-exp-details.tex
\section{Experimental Details}
\label{app:exp-details}
\subsection{Details of the MLLMs Used in Our Experiemnts}
We list the versions and sizes of MLLMs used in the experiments in Table \ref{tab:mllms}. 

\begin{table*}[th]
\centering
\caption{The versions and sizes of current state-of-the-art MLLMs}
\resizebox{0.5\textwidth}{!}{
\begin{tabular}{lcc}
\toprule
Model & Version & Size \\ 

\midrule
LLaVA & LLaVA-1.5 & 13B \\
QWen-VL & QWen-VL & 7B \\ 
CogVLM & cogvlm-chat-v1.1 & 17B \\ 
mPLUG-owl & mPLUG-owl & 7B \\ 
MiniGPT-4 & Vicuna & 13B \\ 
LLaMA-adapter V2 & LLaMA-Adapter V2 multimodal & 7B \\ 
CLIP & clip-vit-large-patch14-336 & 384M\\ 
BLIP-2 & BLIP-2 ViT-g Flan-T5-xxl & 12.1B\\ 
InstructBLIP & Flan-T5-xxl & 11B\\ 
kosmos-2 & kosmos-2-patch14-224  & 1.6B\\ 
Emu-2 & Emu2-Chat & 37B \\ 
Intern & InternLM-XComposer & 7B \\ 
GPT-4V & gpt-4-vision-preview & - \\
Gemini & gemini-pro-vision & - \\

\bottomrule
\end{tabular}
}
\label{tab:mllms}
\end{table*}

\subsection{Details of the Datasets Used in Our Experiemnts}

The introduction of the datasets used in the experiments is as follows.

\begin{itemize}
  \item \textbf{Colored MNIST}\citep{Arjovsky2019InvariantRM} is a variant of the MNIST handwritten digit classification dataset\citep{deng2012mnist}. The domain space $\mathcal{D} = \{0.1, 0.3, 0.9\}$ contains a disjoint set of digits colored either red or blue. The label is a noisy function of the digit and color. This dataset contains 70,000 examples and 2 classes.
  
  \item \textbf{Rotated MNIST}\citep{ghifary2015domain} is a variant of MNIST where the domain space $\mathcal{D} = \{0, 15, 30, 45, 60, 75\}$ contains digits rotated by $ d $ degrees. The dataset contains 10 classes and 70,000 examples.
  
  \item \textbf{PACS}\citep{li2017deeper} comprises four domains $\mathcal{D} = \{\text{art}, \text{cartoons}, \text{photos}, \text{sketches}\} $. This dataset contains 7 classes and  9,991 examples.
  
  \item \textbf{VLCS}\citep{fang2013unbiased} comprises photographic domains $ \mathcal{D} = \{\text{Caltech101}, \text{LabelMe}, \text{SUN09}, \text{VOC2007}\} $. It contains 5 classes and 10,729 examples.
  
  \item \textbf{Office-Home}\citep{venkateswara2017deep} includes four domains $\mathcal{D} = \{\text{art}, \text{clipart}, \text{product}, \text{real}\} $. It contains 65 classes and 15,588 examples.
  
  \item \textbf{Terra Incognita}\citep{beery2018recognition} contains photographs of wild animals taken by camera traps at various locations. 
  
  \item \textbf{DomainNet}\citep{peng2019moment} has six domains $ \mathcal{D} = \{\text{clipart}, \text{infograph}, \text{painting}, \text{quickdraw}, \text{real}, \text{sketch}\} $. DomainNet contains 345 classes and 586,575 examples.

  \item \textbf{Fmow}\citep{peng2019moment} consists of over 1 million images from over 200 countries. For each image, at least one bounding box annotation containing one of 63 categories, including a "false detection" category is provided. 

  \item \textbf{iWildCam}\citep{beery2021iwildcam} is a dataset for counting the number of animals of each species that appear in sequences of images captured with camera traps. The images are from different cameras spread across the globe. The training set contains 201,399 images from 323 locations, and the test set contains 60,029 images from 91 locations. These 414 locations are spread across the globe.

  \item \textbf{SVIRO}\citep{cruz2020sviro} is a synthetic dataset for vehicle interior rear seat Occupancy detection and classification. The dataset consists of 25.000 sceneries across 10 different vehicles and provides several simulated sensor inputs and ground truth data.  

  \item \textbf{NICO++}\citep{zhang2023nico++} is dedicatedly designed for OOD (Out-of-Distribution) image classification. The contexts in NICO++ are divided into two types: 1) 10 common contexts that are aligned across all categories, containing nature, season, humanity, and light; 2) 10 unique domains specifically for each of the 80 categories, including attributes (e.g. action, color), background, camera shooting angle, and accompanying objects and so on. Totally there are more than 230,000 images with both category and domain labels in NICO++.
  
  \item\textbf{Camelyon17}\citep{sun2022camelyon} contains whole-slide images (WSI) of hematoxylin and eosin (H\&E) stained lymph node sections. The data set for CAMELYON17 is collected from 5 medical centers in the Netherlands. The data are provided as TIFF images. Lesion-level annotations are provided as XML files. For training, 100 patients will be provided, and another 100 patients for testing.

  \item \textbf{CT-XCOV}\citep{elbouknify2023ct} is a two-class chest CT dataset (32, including 349 COVID-19 CT scans and 387 non-COVID-19 CT scans.

  \item \textbf{XCOVFour}\citep{han2021semi}  is a four-class chest X-ray dataset, including 239 COVID-19 X-rays, 1,619 healthy X-rays, 1300 bacterial pneumonia X-ravs, and 1.273 viral pneumoniaX-rays images.

  \item \textbf{HAM10000}\citep{tschandl2018ham10000} is a dataset of 10000 training images for detecting pigmented skin lesions. The dataset consists of 10,015 dermatoscopic images. Cases include a representative collection of all important diagnostic categories in the realm of pigmented lesions: Actinic keratoses and intraepithelial carcinoma / Bowen's disease (akiec), basal cell carcinoma (bcc), benign keratosis-like lesions (solar lentigines / seborrheic keratoses and lichen-planus like keratoses, bkl), dermatofibroma (df), melanoma (mel), melanocytic nevi (nv) and vascular lesions (angiomas, angiokeratomas, pyogenic granulomas and hemorrhage, vasc).

  \item \textbf{NIH-Chest}\citep{national2017nih} comprises 112,120 frontal-view X-ray images of 30,805 unique patients with the text-mined fourteen disease image labels (where each image can have multi-labels) including Atelectasis, Consolidation, Infiltration, Pneumothorax, Edema, Emphysema, Fibrosis, Effusion, Pneumonia, Pleur\_thickening, Cardiomegaly, Nodule, Mass and Hernia.

  \item \textbf{DrugOOD\_Assay}\citep{ji2022drugood} is a systematic OOD dataset curator and benchmark for AI-aided drug discovery,  all the datasets are based on ChEMBL\citep{mendez2019chembl}. Domains are split according to the assay. The validation split contains 19,028 molecules from the next largest 314 assays with an average of 60.6 molecules per assay. The test split contains 19,302 molecules from the smallest 314 assays with an average of 27.6 molecules per assay.

  \item \textbf{DrugOOD\_Scaffold}\citep{ji2022drugood} is a systematic OOD dataset curator and benchmark for AI-aided drug discovery,  all the datasets are based on ChEMBL\citep{mendez2019chembl}. Domains are split according to the molecular scaffold. The validation split is from the next largest 6,345 scaffolds, with a total of 11,683 molecules, and an average of 1.84 molecules per scaffold. The test is from the smallest 4,350 scaffolds, with a total of 19,048 molecules, and an average of 4.42 molecules per scaffold.

  \item \textbf{DrugOOD\_Size}\citep{ji2022drugood} is a systematic OOD dataset curator and benchmark for AI-aided drug discovery,  all the datasets are based on ChEMBL\citep{mendez2019chembl}. Domains are split according to the number of atoms. The validation set is from the next largest 4 size groups in addition to ID data, with 17,660 molecules, and an average of 4,415 molecules per group. The test is from the smallest 18-size groups, with 19,048 molecules, and an average of 1,058 molecules per group.

  \item \textbf{DrugOOD\_Protein}\citep{ji2022drugood} is a systematic OOD dataset curator and benchmark for AI-aided drug discovery, all the datasets are based on ChEMBL\citep{mendez2019chembl}. Samples with the same target protein are in the same domain.
  
\end{itemize}

\subsection{Our Prompt for Zero-Shot Generalization}
\label{sec:app-prompt-zero-shot}
Inspired by \citep{han2023well}, we adopt structured prompts to drive MLLMs response in a controlled way. We show our prompt for zero-shot generalization in Table \ref{tab:prompt-zero-shot}.

\begin{table*}[th]
\centering
\caption{The prompt for zero-shot generalization. We take the prompt for the VLCS dataset as an example.}
\begin{tabular}{l}
\toprule

\textless Image\textgreater \\
Given the image above, answer the following question using the specified format. \\
Question: What is in the image above? \\
Choices: [bird, car, chair, dog, person]. \\
Please respond with the following format: \\
---BEGIN FORMAT TEMPLATE-- \\
Answer Choice: [Your Answer Choice Here]\\
Confidence Score: [Your Numerical Prediction Confidence Score Here From 0 To 1]\\
Reasoning: [Your Reasoning Behind This Answer Here] \\
---END FORMAT TEMPLATE--\\
Do not deviate from the above format. Repeat the format template for the answer. \\

\bottomrule
\end{tabular}
\label{tab:prompt-zero-shot}
\end{table*}

\subsection{Our Prompt for ICL Generalization}
\label{sec:app-prompt-icl}

Table \ref{tab:prompt-icl} shows our prompt for ICL generalization.

\begin{table*}[th]
\centering
\caption{The prompt for ICL. We take the prompt for the VLCS dataset as an example.}
\begin{tabular}{l}
\toprule

\textless ICE\_Image\textgreater \\
Given the image above, answer the following question using the specified format. \\
Question: What is in the image above? \\
Choices: [bird, car, chair, dog, person]. \\
Please respond with the following format: \\
---BEGIN FORMAT TEMPLATE-- \\
Answer Choice: [Your Answer Choice Here]\\
---END FORMAT TEMPLATE--\\
Do not deviate from the above format. Repeat the format template for the answer. \\
Answer Choice: ICE\_Label \\
\\
$\cdots$ (optional for other ICE)\\
\\
\textless Image\textgreater \\
Given the image above, answer the following question using the specified format. \\
Question: What is in the image above? \\
Choices: [bird, car, chair, dog, person]. \\
Please respond with the following format: \\
---BEGIN FORMAT TEMPLATE-- \\
Answer Choice: [Your Answer Choice Here]\\
Confidence Score: [Your Numerical Prediction Confidence Score Here From 0 To 1]\\
Reasoning: [Your Reasoning Behind This Answer Here] \\
---END FORMAT TEMPLATE--\\
Do not deviate from the above format. Repeat the format template for the answer. \\

\bottomrule
\end{tabular}
\label{tab:prompt-icl}
\end{table*}

%% file: paragraphs/conclusion.tex
\vspace{-10pt}
\section{Conclusion} 
\label{sect:conclusion}
We investigated the generalization of current MLLMs and found that they showed unreliable performance beyond common training domains. We found that the mapping deficiency could be the major hurdle and showed the potential of ICL to largely enhance MLLMs' generalization. We further presented ICL's vulnerability to label and spurious correlation shifts.

%% file: paragraphs/app-error-anly.tex
\section{Failure Analysis}
\label{app-error}
\subsection{Potential shared biases among MLLMs}
\begin{figure*}[ht]
    \centering
    \begin{subfigure}[b]{0.29\linewidth}
        \centering
        \includegraphics[width=\linewidth]{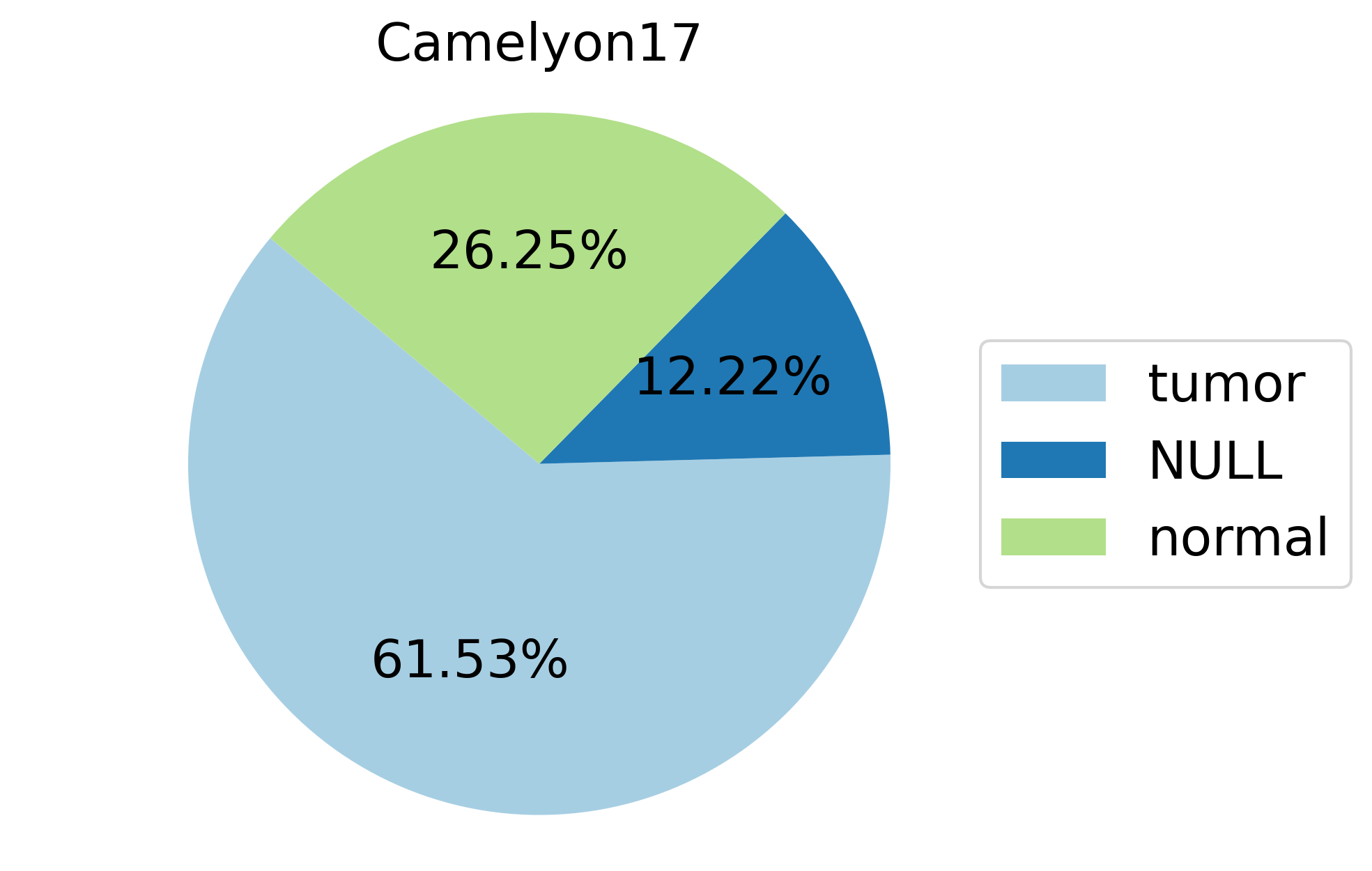}
        \caption{}
        \label{fig:The Proportion of incorrect answers-Camelyon17}
    \end{subfigure} 
    \hfill%
    \begin{subfigure}[b]{0.32\linewidth}
        \centering
        \includegraphics[width=\linewidth]{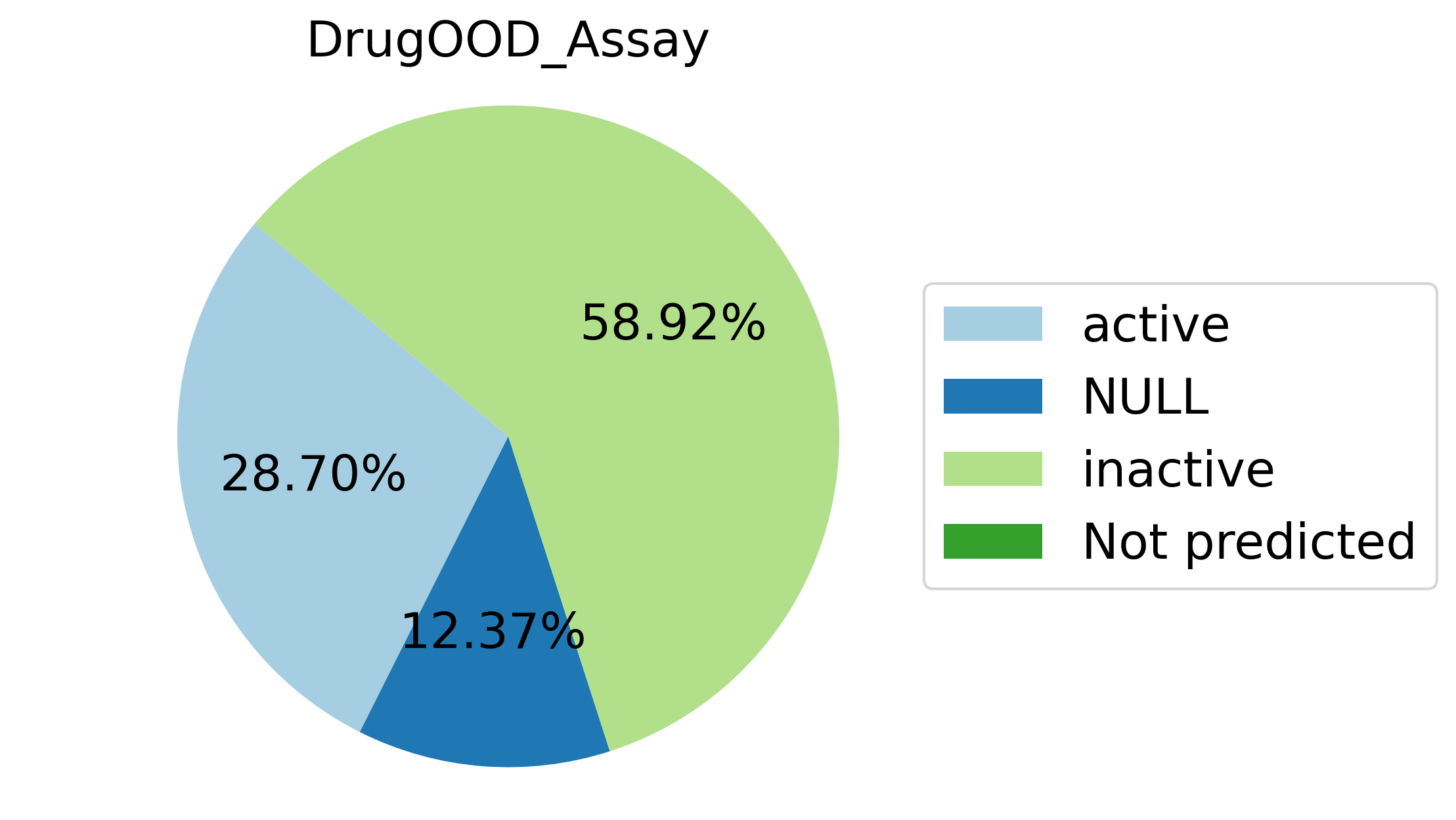} 
        \caption{} 
        \label{fig:The Proportion of incorrect answers-DrugOOD_Assay} 
    \end{subfigure}
    \hfill%
    \begin{subfigure}[b]{0.32\linewidth}
        \centering
        \includegraphics[width=\linewidth]{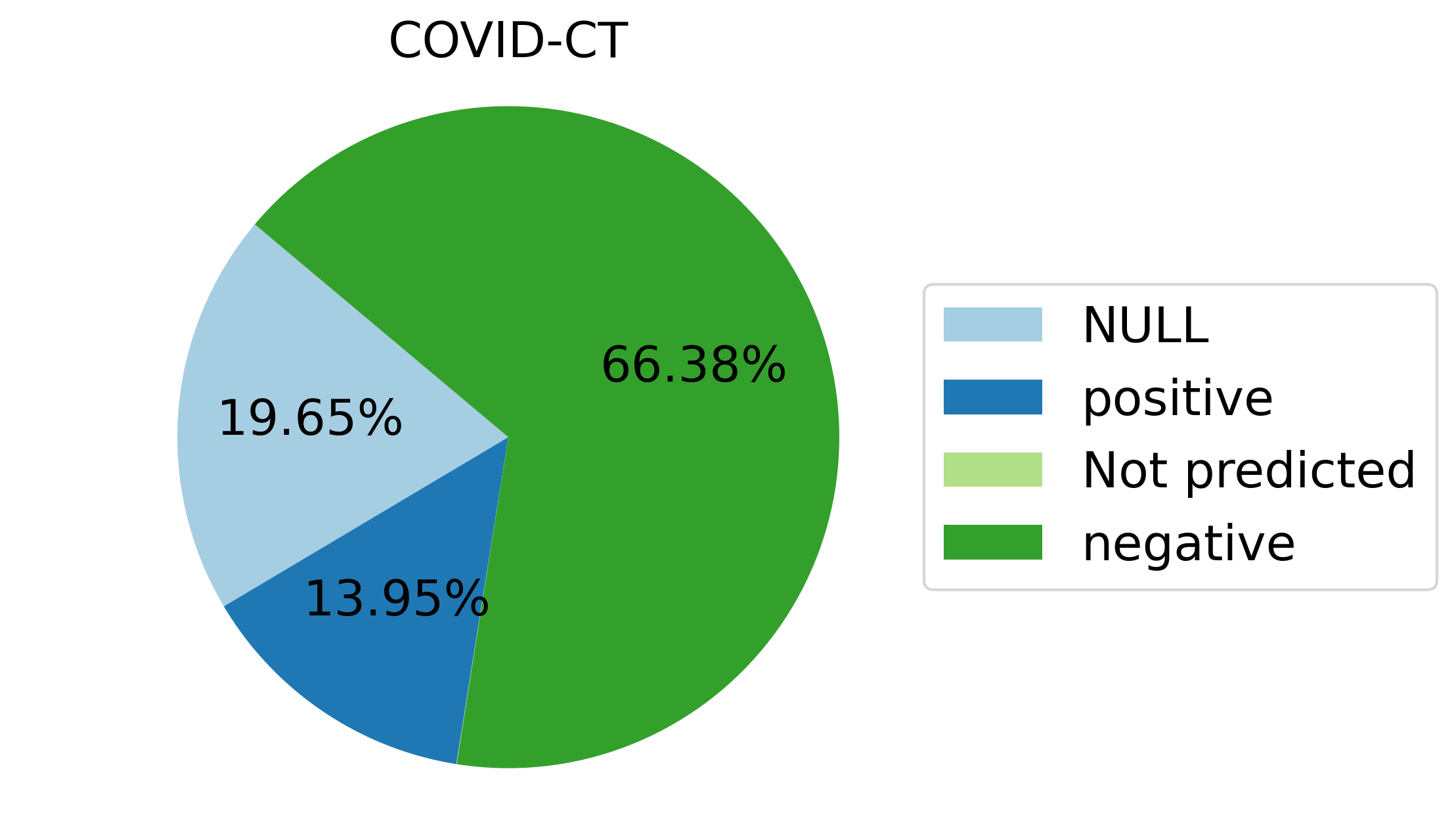}
        \caption{}
        \label{fig:The Proportion of incorrect answers-COVID-CT}
    \end{subfigure}
    \begin{subfigure}[b]{0.32\linewidth}
        \centering
        \includegraphics[width=\linewidth]{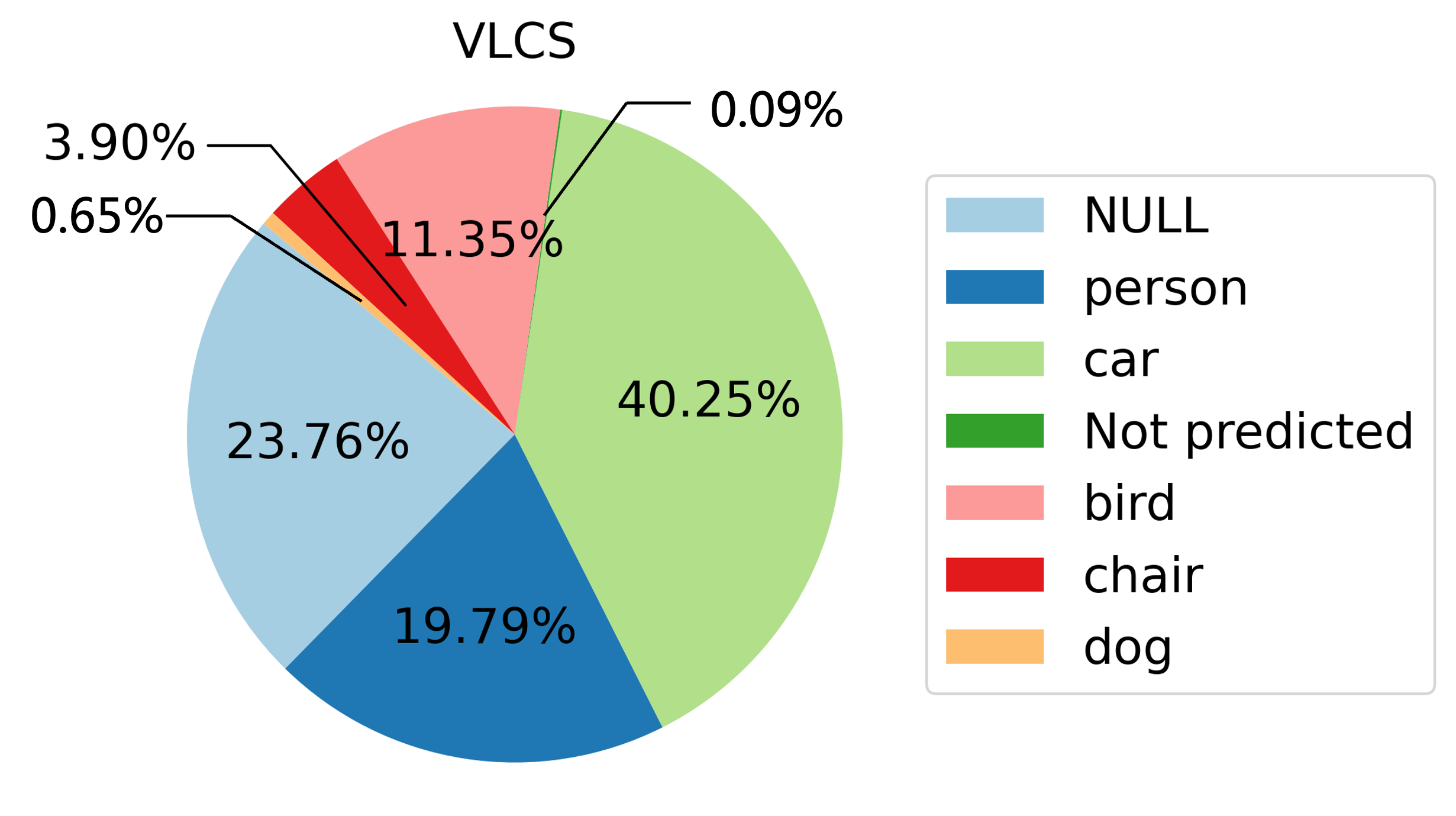}
        \caption{}
        \label{fig:The Proportion of incorrect answers-VLCS}
    \end{subfigure}
    \begin{subfigure}[b]{0.32\linewidth}
        \centering
        \includegraphics[width=\linewidth]{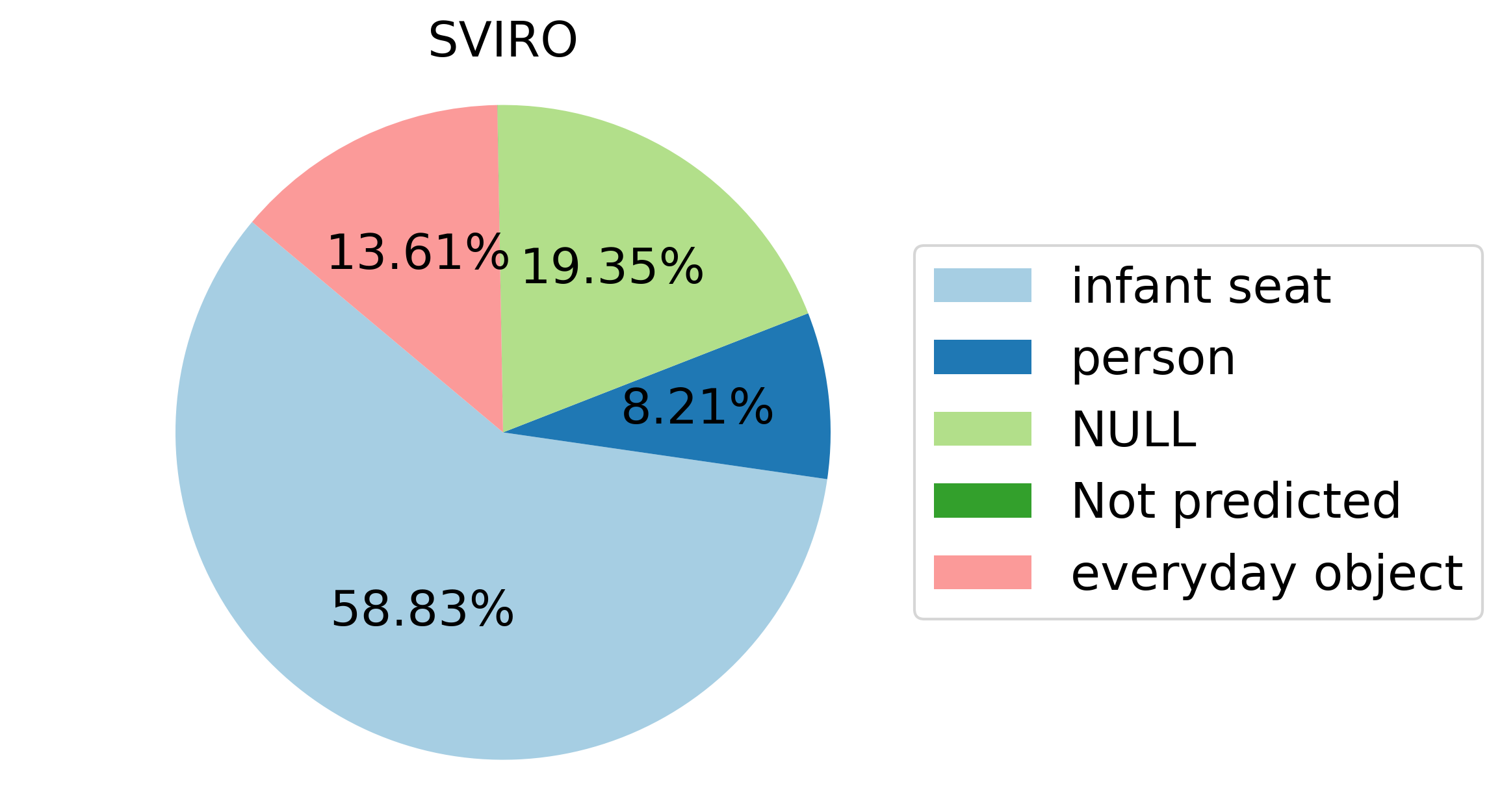}
        \caption{}
        \label{fig:The Proportion of incorrect answers-SVIRO}
    \end{subfigure}
    \begin{subfigure}[b]{0.32\linewidth}
        \centering
        \includegraphics[width=\linewidth]{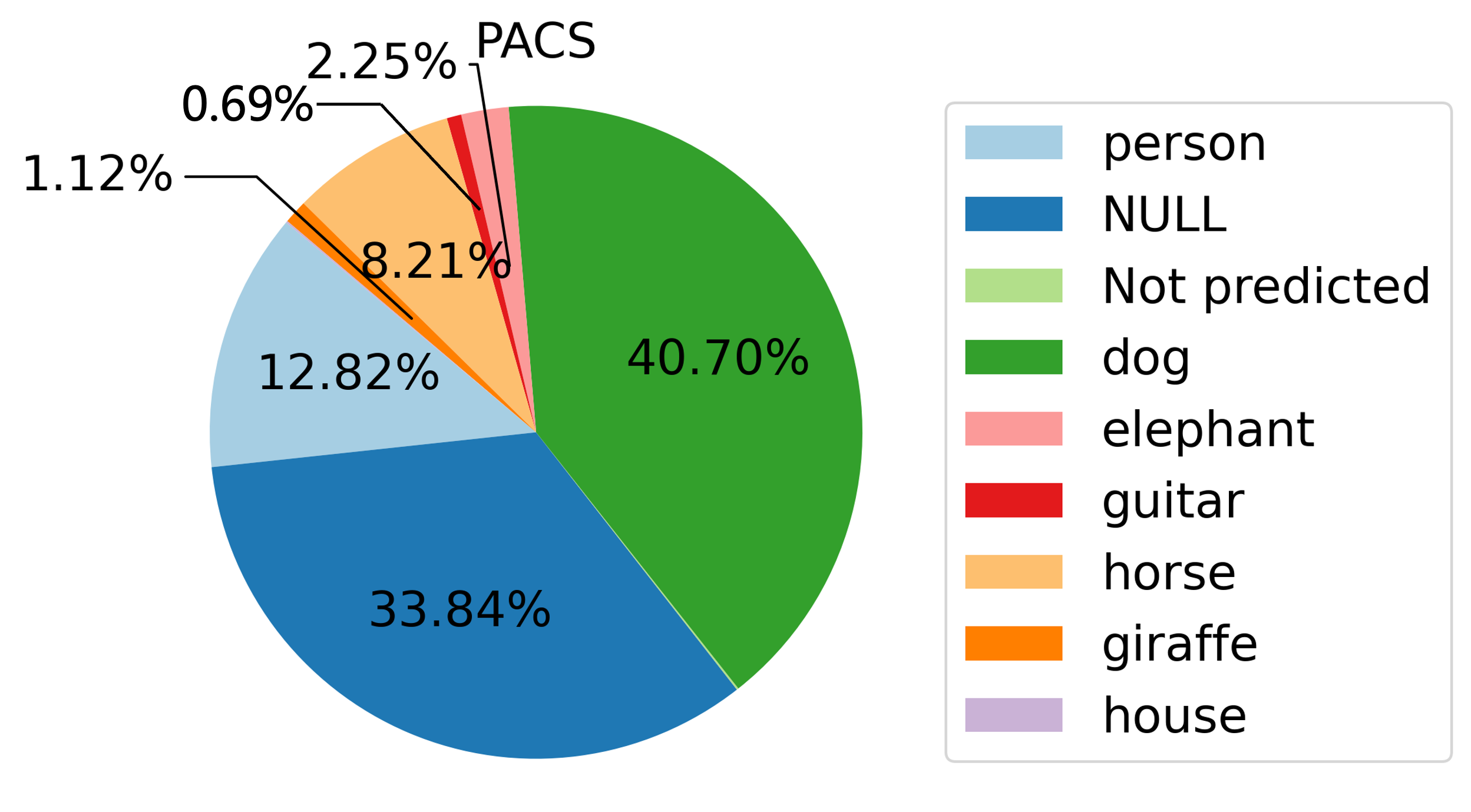}
        \caption{}
        \label{fig:The Proportion of incorrect answers-PACS}
    \end{subfigure}
    \caption{The Proportion of categories that MLLMs incorrectly predicted into on OOD generalization and domain-specific datasets.}
    \label{fig:app-error-1}
\end{figure*}

\begin{figure*}[ht]
    \centering
    \label{fig:The Proportion of incorrect models on datasets}
    \includegraphics[width=0.7\textwidth]{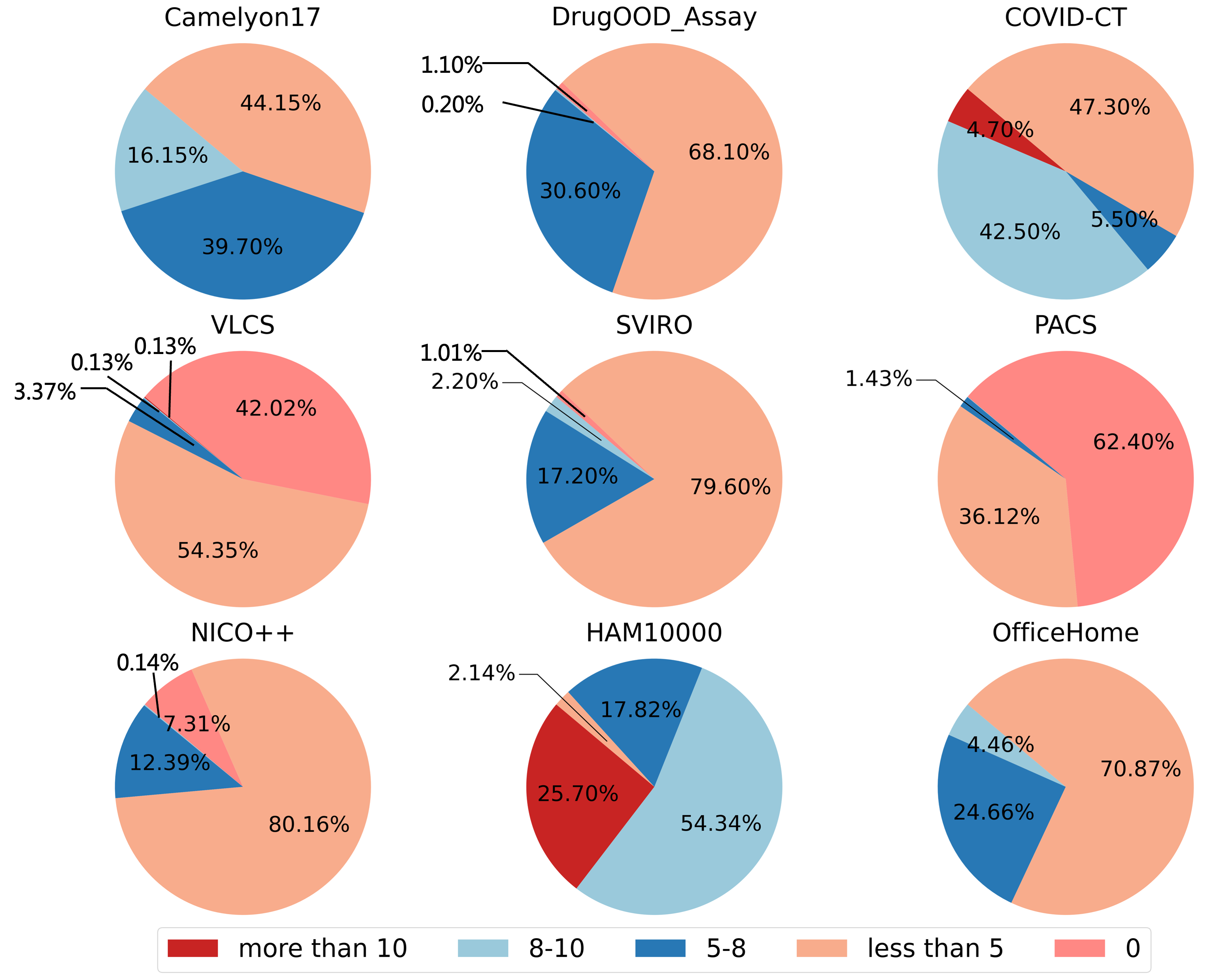}
    \caption{The proportion of numbers of models that predict incorrectly on each dataset. The errors are not centralized in most of the datasets and are distributed in a random scatter.}
    \label{fig:app-error-2}
\end{figure*}

We show the Proportion of categories that MLLMs incorrectly predicted into on OOD generalization and domain-specific datasets in Figure \ref{fig:app-error-1} and the proportion of numbers of models that predict incorrectly on each dataset in Figure \ref{fig:app-error-2}. 
We reveal a lack of consistent, dataset-specific biases. Instead, errors appeared scattered across most datasets. 

\subsection{Imformative and Contextual Prompts for Domain-specific Tasks}
\label{app-new-prompt}
To analyze the impact of semantic misinterpretation within text input on model failure, we use informative and distinct prompts that provide additional scientific context and guidance for each dataset. These prompts aim to address the potential limitations of MLLMs in understanding specific meanings and implications of certain categories within the input prompt, particularly in specialized scientific contexts. We present these prompts for the Camelyon17, HAM10000, and NIH-Chest dataset in Table \ref{tab:new-prompt-camelyon17}, \ref{tab:new-prompt-ham10000}, and \ref{tab:new-prompt-nih}, respectively.

\begin{table*}[th]
\centering
\caption{Imformative and Contextual Prompt for the Camelyon17 dataset as an example.}
\begin{tabular}{l}
\toprule

\textless Image\textgreater \\
Given an image from the Camelyon17 dataset, which focuses on the detection of breast cancer 
\\and involves classifying metastatic and non-metastatic regions in lymph node sections from \\
breast cancer patients, answer the following question using the specified format. \\
Question: What is in the image above? \\
Choices: ['normal', 'tumor']. \\
Normal represents normal lymph node cells, tumor represents metastatic breast cancer \\
lymph nodes. 
Each class has distinctive visual characteristics:\\
1. normal Lymph Nodes: Show preserved architecture with distinct cortical and medullary \\ 
regions. Follicles (sites of immune cell aggregation) and sinuses (spaces within the node) \\
are normally visible.\\
2. tumor: Disruption of normal architecture by tumor cells. Metastatic cells can appear\\
as clusters or diffusely infiltrating the lymph node, often disrupting the normal\\
nodal structure. \\
They might be larger, have irregular shapes, and have nuclei that are larger and more variable\\
in shape.\\
Based on the above information, please answer the question and respond with the\\
following format: \\
---BEGIN FORMAT TEMPLATE-- \\
Answer Choice: [Your Answer Choice Here]\\
---END FORMAT TEMPLATE--\\
Do not deviate from the above format. Repeat the format template for the answer. \\

\bottomrule
\end{tabular}
\label{tab:new-prompt-camelyon17}
\end{table*}

\begin{table*}[th]
\centering
\caption{Imformative and Contextual Prompt for the HAM10000 dataset as an example.}
\begin{tabular}{l}
\toprule

\textless Image\textgreater \\
Given an image from the HAM10000 dataset, which is used in dermatology image analysis for\\
skin lesion classification, answer the following question using the specified format.\\
Question: What is in the image above?\\
Choice : ['actinic keratoses and intraepithelial carcinoma', 'basal cell carcinoma', \\
'benign keratosis-like lesions', 'dermatofibroma', 'melanoma', 'melanocytic nevi', \\
'vascular lesions'].\\
Each class has distinctive visual characteristics:\\
1. Actinic Keratoses and Intraepithelial Carcinoma: Rough, scaly patches, varying in color due \\
to sun exposure.\\
2. Basal Cell Carcinoma: Small, shiny bumps or red patches, resembling sores or scars.\\
3. Benign Keratosis-like Lesions: Common, non-cancerous growths, typically brown or \\
skin-toned, rough bumps.\\
4. Dermatofibroma: Benign, firm, raised nodules, usually brownish to red.\\
5. Melanoma: Asymmetrical, irregularly bordered, multi-colored moles larger than a pencil \\
eraser.\\
6. Melanocytic Nevi: Uniform, round, and brown moles, varying in color and size, usually \\
benign.\\
7. Vascular Lesions: Related to blood vessels, often red or purple discolorations, like \\
birthmarks or spider veins.\\
Based on the above information, please answer the question and respond with the following\\
format: \\
---BEGIN FORMAT TEMPLATE-- \\
Answer Choice: [Your Answer Choice Here]\\
---END FORMAT TEMPLATE--\\
Do not deviate from the above format. Repeat the format template for the answer. \\

\bottomrule
\end{tabular}
\label{tab:new-prompt-ham10000}
\end{table*}

\begin{table*}[th]
\centering
\caption{Imformative and Contextual Prompt for the NIH-Chest dataset as an example.}
\begin{tabular}{l}
\toprule

\textless Image\textgreater \\
Given an image from the NIH-Chest Dataset, used in radiology for diagnosing thoracic diseases,\\
answer the following question using the specified format.\\
Question: What is in the image above? \\
Choice list: ['Cardiomegaly', 'Emphysema', 'Effusion', 'No Finding', 'Fibrosis', 'Hernia', \\
'Tortuous Aorta', 'Infiltration', 'Mass', 'Nodule', 'Atelectasis', 'Pneumothorax', \\
'Pleural Thickening', 'Calcification of the Aorta', 'Pneumonia', 'Pneumomediastinum', \\ 'Subcutaneous Emphysema', 'Pneumoperitoneum', 'Fibrosis', 'Edema',
'Consolidation'] .\\
Each class has distinctive visual characteristics:\\
1. Cardiomegaly: Enlarged heart, increased heart-to-chest size ratio.\\
2. Emphysema: Overinflated air sacs, abnormally clear lung fields.\\
3. Effusion: Fluid in pleural space, whitish area at lung base.\\
4. No Finding: Normal X-ray without pathology.\\
5. Fibrosis: Thickened lung tissue, reticular/linear markings.\\
6. Hernia: Displaced organs, unusual mass/air pocket.\\
7. Tortuous Aorta: Twisted/elongated aorta, irregular outline.\\
8. Infiltration: Increased opacity, indicating fluid/cells in the lungs.\\
9. Mass: Large growth or lesion, well-defined dense area.\\
10. Nodule: Small, round spot, potential tumor/pathology.\\
11. Atelectasis: Partial lung collapse, increased density areas.\\
12. Pneumothorax: Air in pleural space, clear space alongside lung.\\
13. Pleural Thickening: Thickened pleural lining, irregular white line.\\
14. Calcification of Aorta: Hardened aorta, white areas along the path.\\
15. Pneumonia: Lung infection, patchy/diffuse white areas.\\
16. Pneumomediastinum: Air in mediastinum, clear areas around heart/vessels.\\
17. Subcutaneous Emphysema: Air under skin, air pockets outside the lung.\\
18. Pneumoperitoneum: Air in the abdominal cavity, the clear area beneath the diaphragm.\\
19. Edema: Excess fluid in lungs, diffuse cloudiness.\\
20. Consolidation: Lung airspaces filled with material, solid white areas.\\
Based on the above information, please answer the question and respond with the following\\
format: \\
---BEGIN FORMAT TEMPLATE-- \\
Answer Choice: [Your Answer Choice Here]\\
---END FORMAT TEMPLATE--\\
Do not deviate from the above format. Repeat the format template for the answer. \\

\bottomrule
\end{tabular}
\label{tab:new-prompt-nih}
\end{table*}

%% file: paragraphs/app-showcase-zero-shot.tex
\section{Case Demonstration}
\label{sec:app-casestudy}
In this section, we present specific cases of MLLMs on OOD generalization and domain-specific data.
Analyzing both successful and failed predictions, particularly by comparing the confidence and reasoning behind their inferences, offers valuable insights into the internal workings of LLMs and the factors contributing to their successes and limitations. 

\subsection{Case Demonstration of Zero-shot Generalization}
Here we present cases of zero-shot generalization of MLLMs. 
While LLMs demonstrate impressive capabilities in capturing salient features of common objects, they encounter significant limitations in comprehending specific visual domains. 
\textbf{Please note that we omit the detailed prompt and only category selection space for understanding in the following presentation of case studies. All the models are assigned to provide responses to the same textual prompt as shown in Section \ref{sec:zero-shot} and \ref{sec:app-prompt-zero-shot}.}

\begin{figure*}[ht]
    \centering
    \label{fig:}
    \includegraphics[width=1\textwidth]{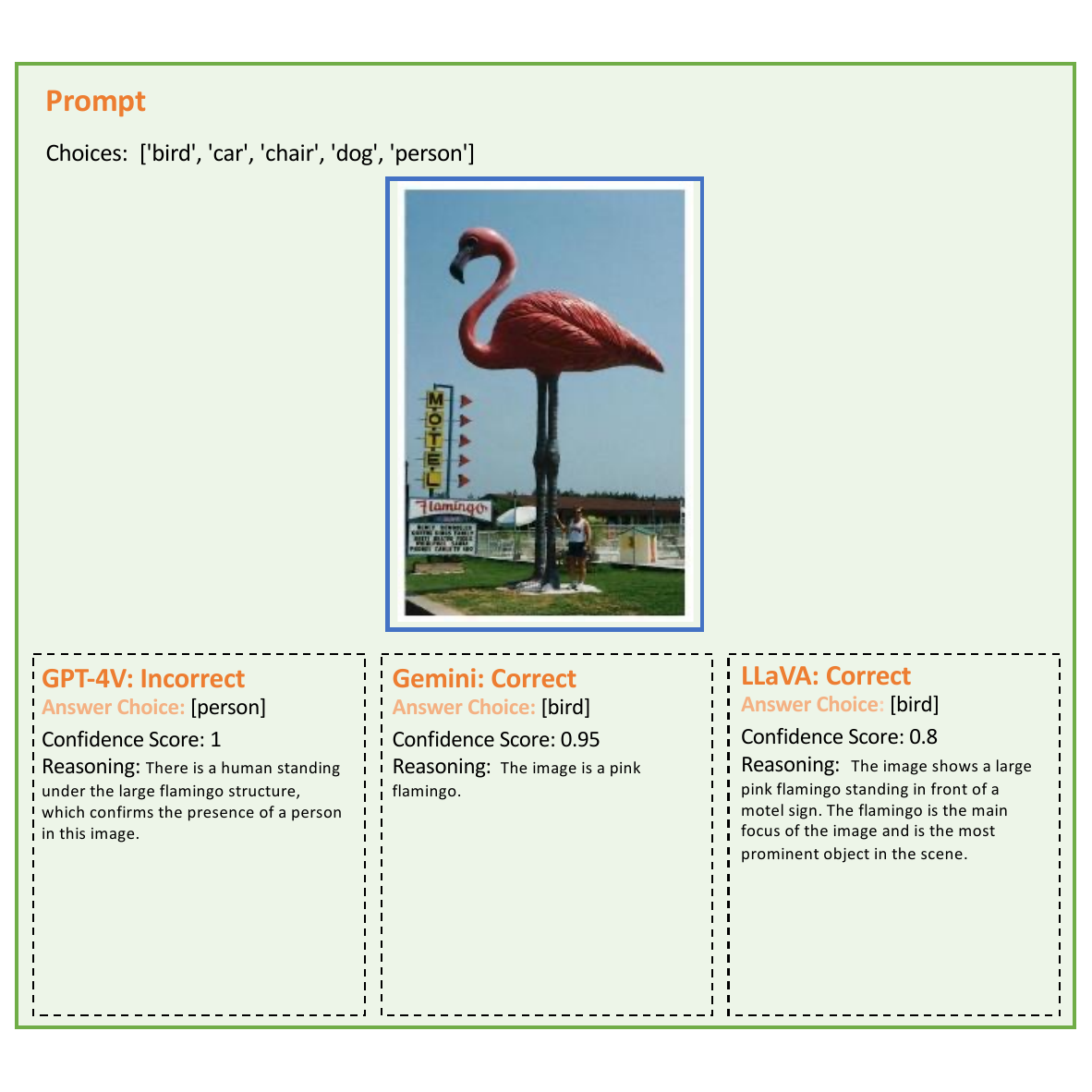}
    \caption{Synthetic and Natural Distribution Shifts in zero-shot generalization: Case 1, analyzing the \textit{bird} category in the \textit{Caltech101} domain of the VLCS dataset. In this case, both Gemini and LLaVA predict correctly, but Gemini gives a higher confidence score and LLaVA gives more detailed explanations such as the main focus of the image. GPT-4V mentions the bird in the image in its explanation, but focuses on the person beneath the giant flamingo, leading to a different category prediction from the label.}
\end{figure*}

\begin{figure*}[ht]
    \centering
    \label{fig:}
    \includegraphics[width=1\textwidth]{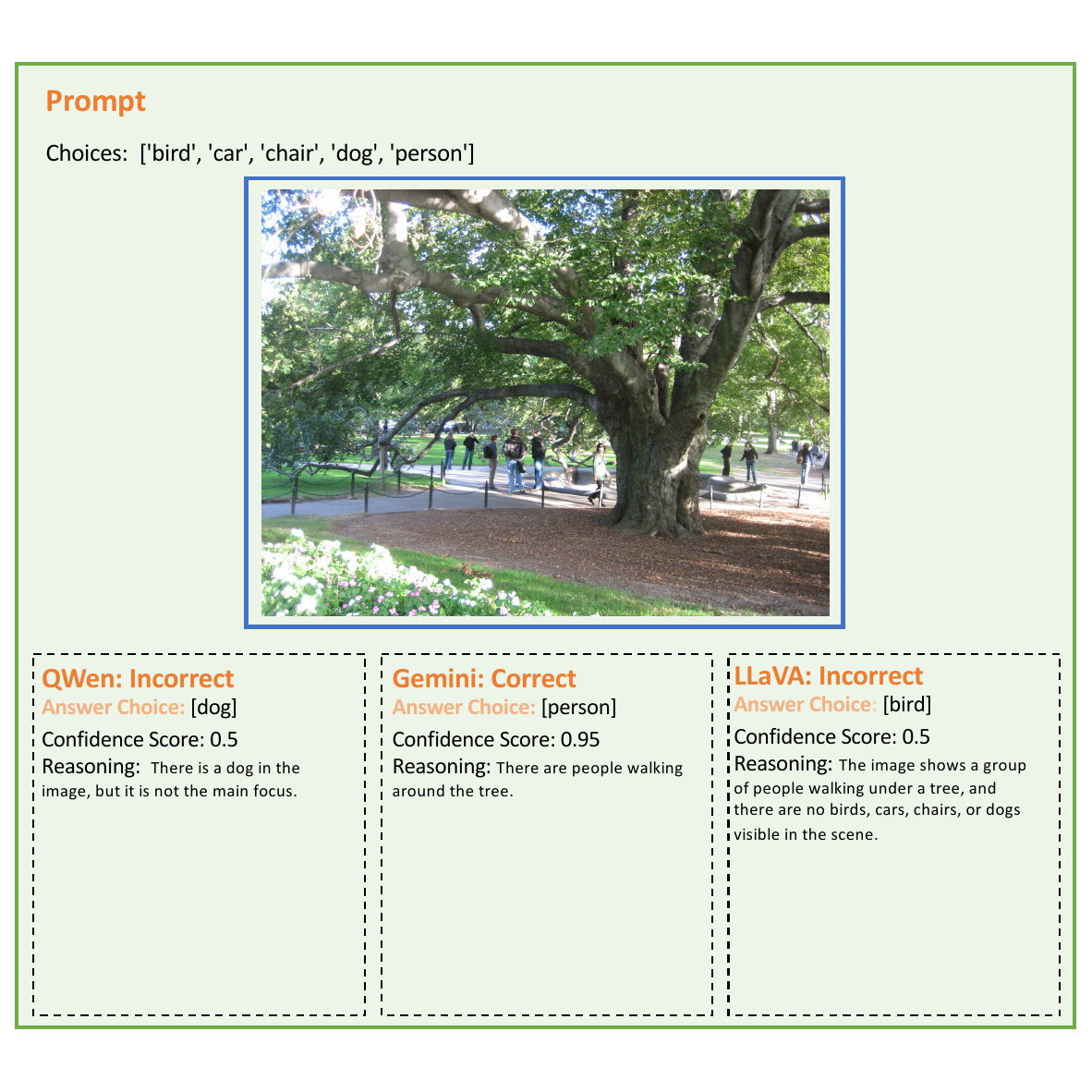}
    \caption{Synthetic and Natural Distribution Shifts in zero-shot generalization: Case 2, analyzing the \textit{person} category in the \textit{LabelMe} domain of the VLCS dataset. In this case, Gemini predicts correctly while QWen and LLaVA predict incorrectly. In the reasoning, Gemini broadly describes the content of the image. QWen notes the presence of a dog in the picture, though it is not the focal point, yet QWen still predicts the category as \textit{dog}. The rationale and choice provided by LLaVA do not align. In its reasoning, it mentions that there are no birds, cars, chairs, or dogs visible in the scene, yet it chooses \textit{bird} in its prediction.}
\end{figure*}

\begin{figure*}[ht]
    \centering
    \label{fig:}
    \includegraphics[width=1\textwidth]{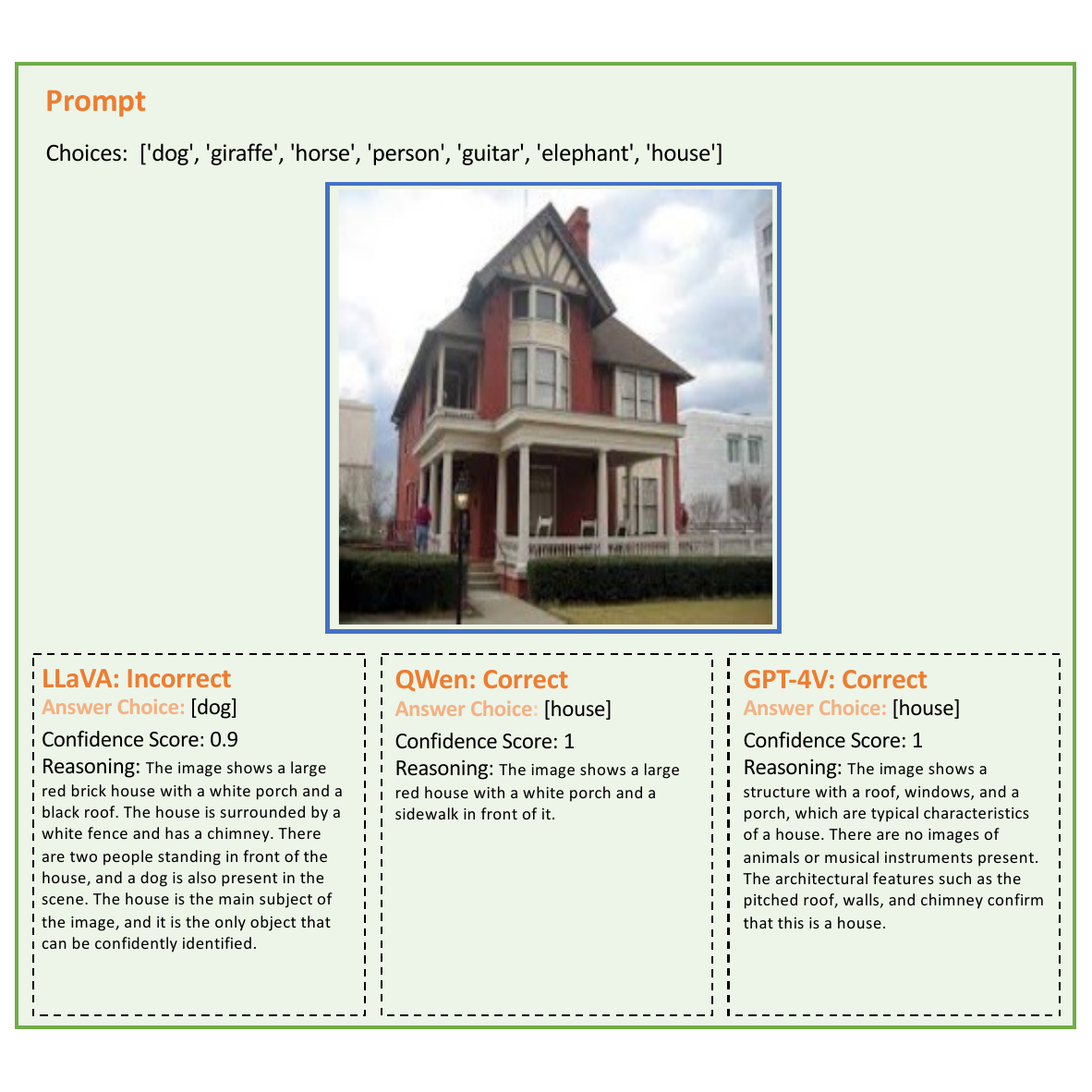}
    \caption{Synthetic and Natural Distribution Shifts in zero-shot generalization: Case 3, analyzing the \textit{house} category in the \textit{Photo} domain of the PACS dataset. In this case, QWen and GPT-4V predict correctly, with a high confidence score of 1.00. GPT-4V gives a more nuanced and reliable reason. The reasoning of LLaVA includes a detailed description of the objects and features present in the image, such as the shape of the house, the colors, and a dog. It notes that `The house is the main subject of the image,' yet the model opts for \textit{dog} for its prediction.}
\end{figure*}

\begin{figure*}[ht]
    \centering
    \label{fig:}
    \includegraphics[width=1\textwidth]{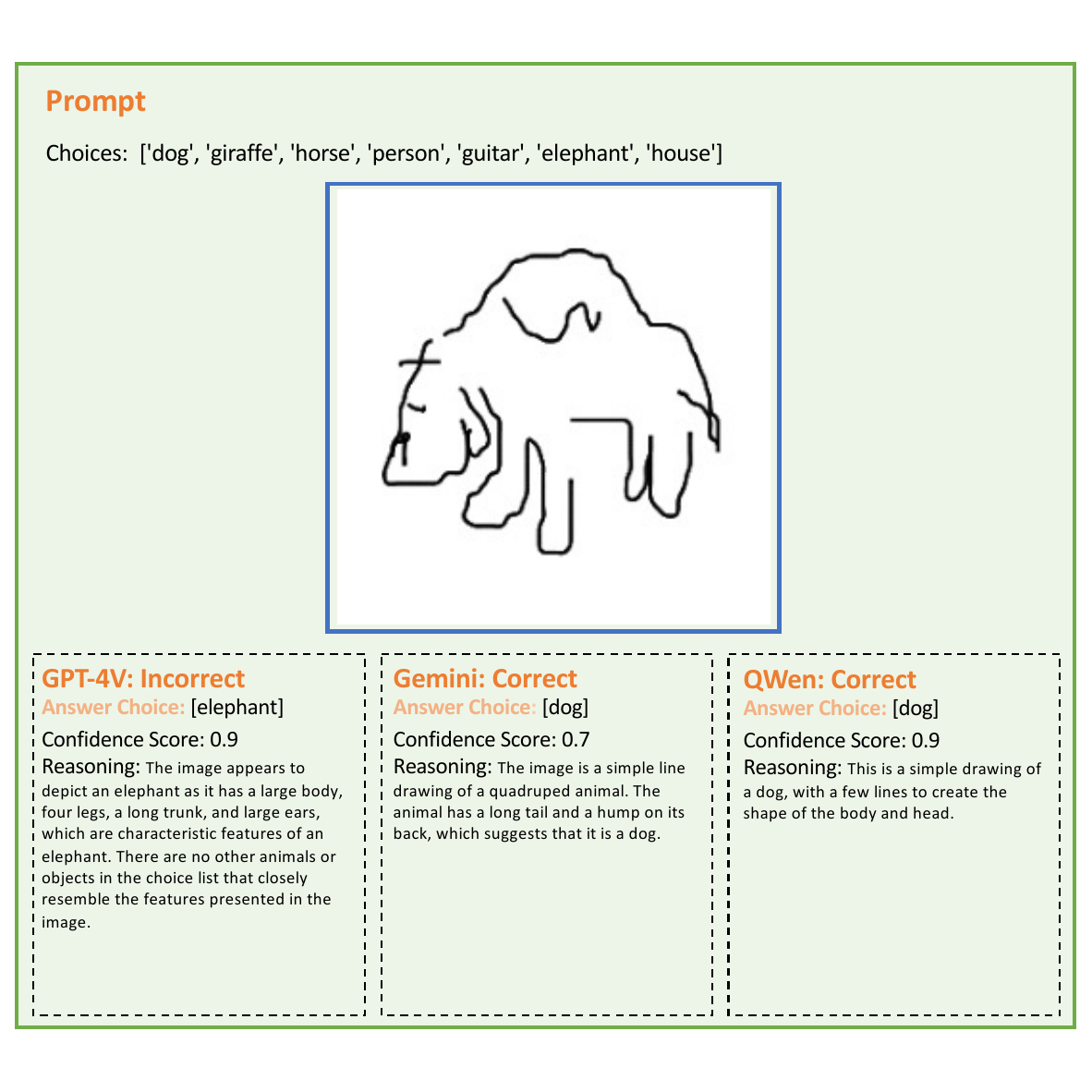}
    \caption{Synthetic and Natural Distribution Shifts in zero-shot generalization: Case 4, analyzing the \textit{dog} category in the \textit{Sketch} domain of the PACS dataset. In this case, both Gemini and QWen make correct predictions, but QWen predicts with a higher confidence level of 0.9. Gemini provides a more detailed and sufficient rationale for its prediction. GPT-4V describes the shapes of objects in the image in its reasoning, leading it to predict the image as depicting an elephant.}
\end{figure*}

\begin{figure*}[ht]
    \centering
    \label{fig:}
    \includegraphics[width=1\textwidth]{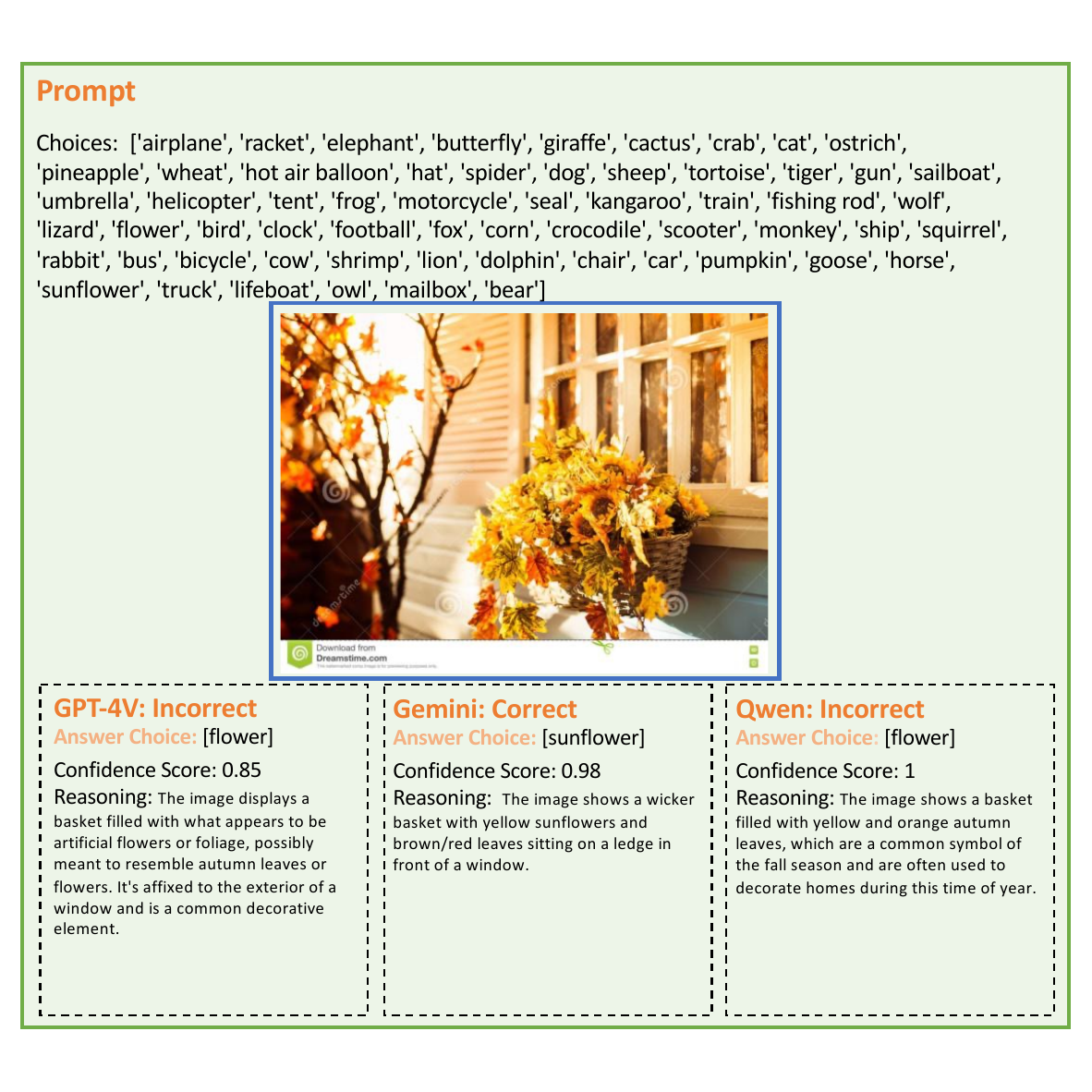}
    \caption{Synthetic and Natural Distribution Shifts in zero-shot generalization: Case 5, analyzing the \textit{sunflower} category in the \textit{autumn} domain of the NICO++ dataset. In this case, Gemini predicts correctly while GPT-4V and QWen predict incorrectly. GPT-4V and QWen both mention in their rationale that the image depicts a decorative element, a basket filled with yellow flowers and orange autumn leaves. However, since `leaves' is not an option, both models predict the subject as \textit{flowers}. Gemini also recognizes the basket in the image and identifies the sunflowers placed within it.}
\end{figure*}

\begin{figure*}[ht]
    \centering
    \label{fig:}
    \includegraphics[width=1\textwidth]{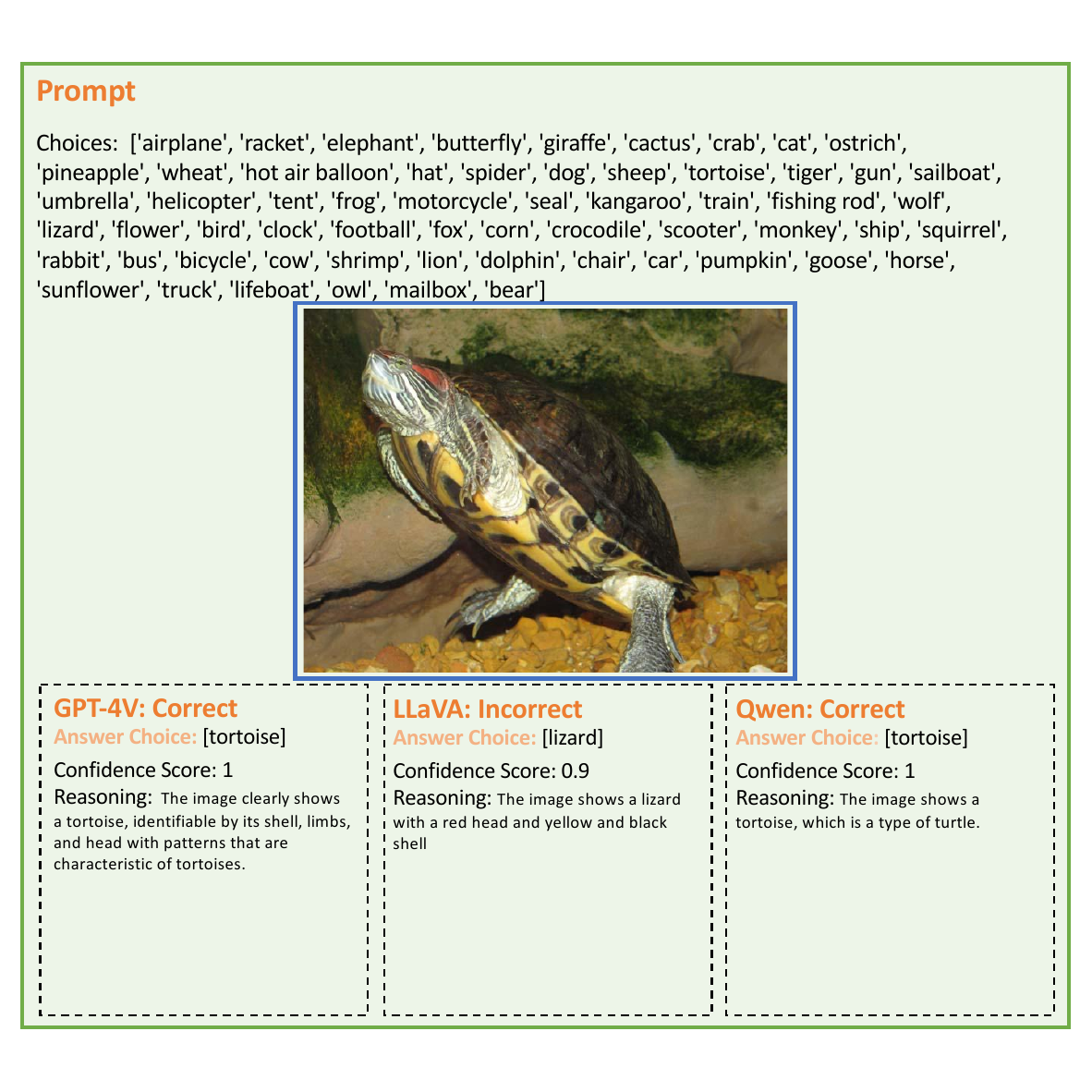}
    \caption{Synthetic and Natural Distribution Shifts in zero-shot generalization: Case 6, analyzing the \textit{tortoise} category in the \textit{autumn} domain of the NICO++ dataset. In this case, both GPT-4V and QWen predict correctly with a confidence score of 1.00. GPT-4V provides a more detailed and substantive rationale, identifying the object in the image as a \textit{tortoise} by its shell, limbs, and head. QWen merely states that the object in the image is a type of turtle, without elaborating further. LLaVA recognizes the object in the image as a lizard.}
\end{figure*}

\begin{figure*}[ht]
    \centering
    \label{fig:}
    \includegraphics[width=1\textwidth]{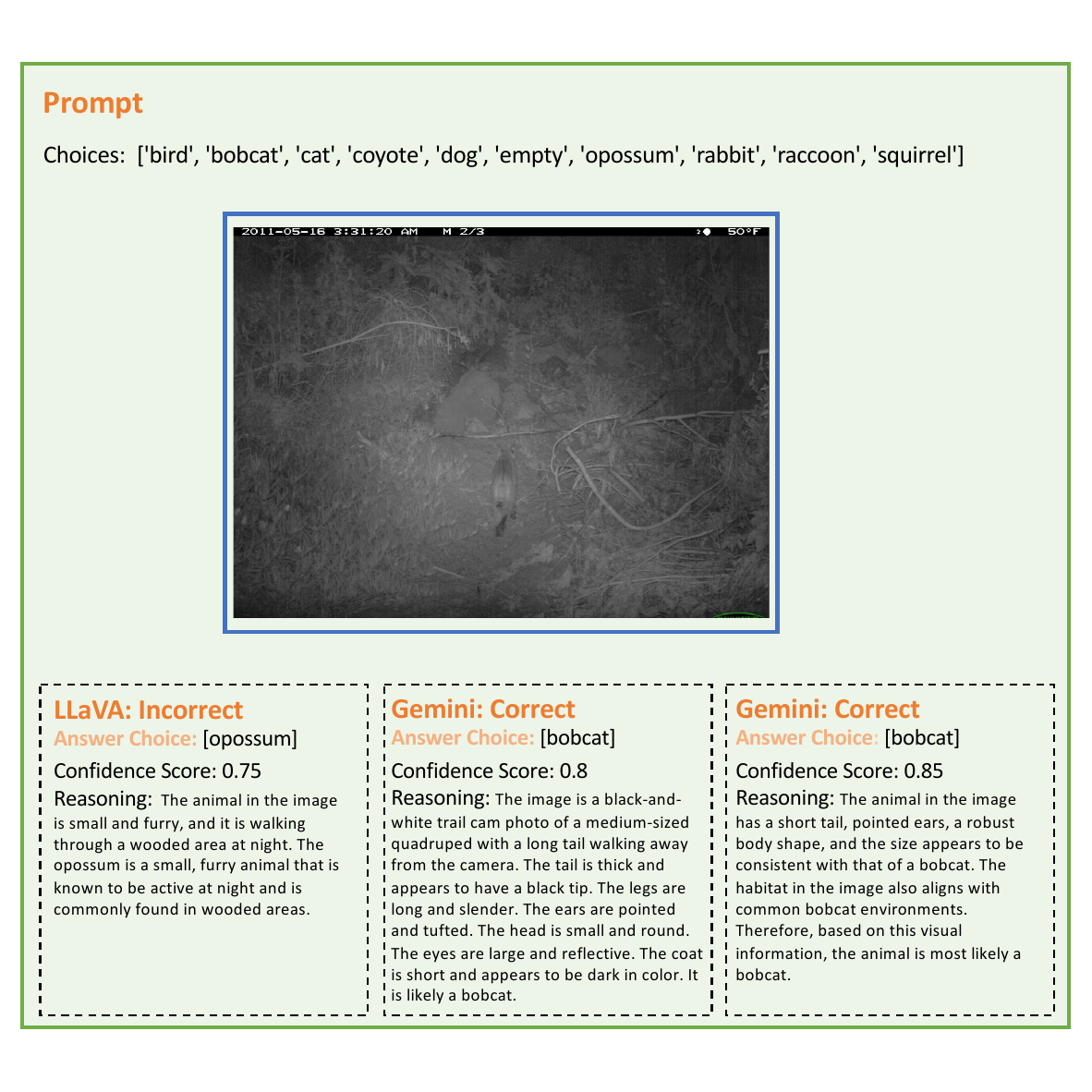}
    \caption{Synthetic and Natural Distribution Shifts in zero-shot generalization: Case 7, analyzing the \textit{bobcat} category in the \textit{location\_100} domain of the TerraInc dataset. In this case, both Gemini and GPT-4V predict correctly, but GPT-4V gives a higher confidence score and Gemini gives more detailed explanations such as long and slender legs, pointed and tufted ears, small and round head, and large and reflective eyes. These are the basic characteristics of a bobcat. LLaVA classifies the object in the image as an opossum, reasoning that its small size, furry appearance, and nocturnal activity lead to this conclusion.}
\end{figure*}

\begin{figure*}[ht]
    \centering
    \label{fig:}
    \includegraphics[width=1\textwidth]{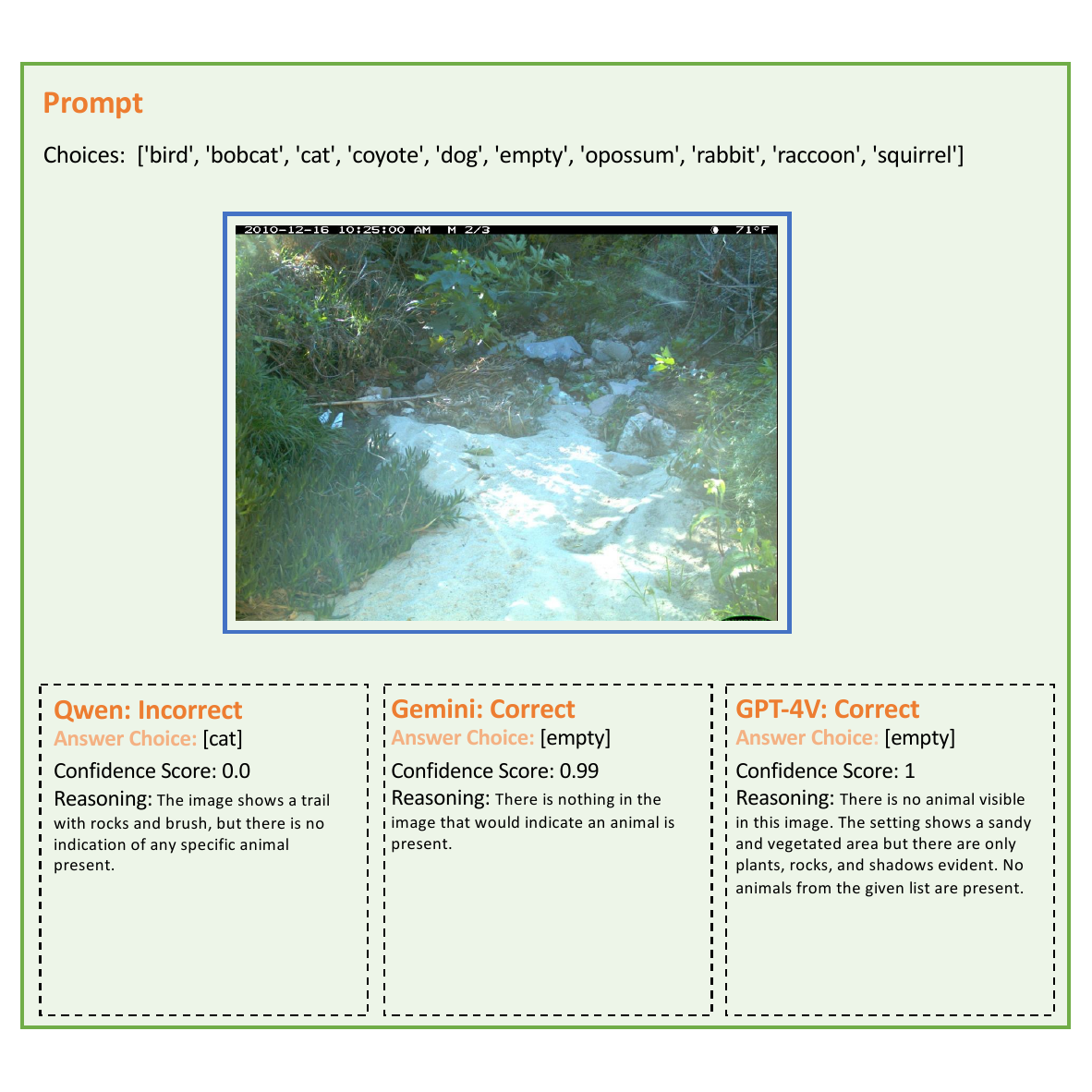}
    \caption{Synthetic and Natural Distribution Shifts in zero-shot generalization: Case 8, analyzing the \textit{empty} category in the \textit{location\_100} domain of the TerraInc dataset. In this case, both Gemini and GPT-4V predict correctly, but GPT-4V gives a higher confidence score of 1.00 as well as more detailed explanations such as only showing a sandy and vegetated area. QWen also believes that there are no animals in the picture, noting that it "only shows a trail with rocks and brush". However, it paradoxically predicts the object category in the image as \textit{cat} with a confidence score of 0.}
\end{figure*}

\begin{figure*}[ht]
    \centering
    \label{fig:}
    \includegraphics[width=1\textwidth]{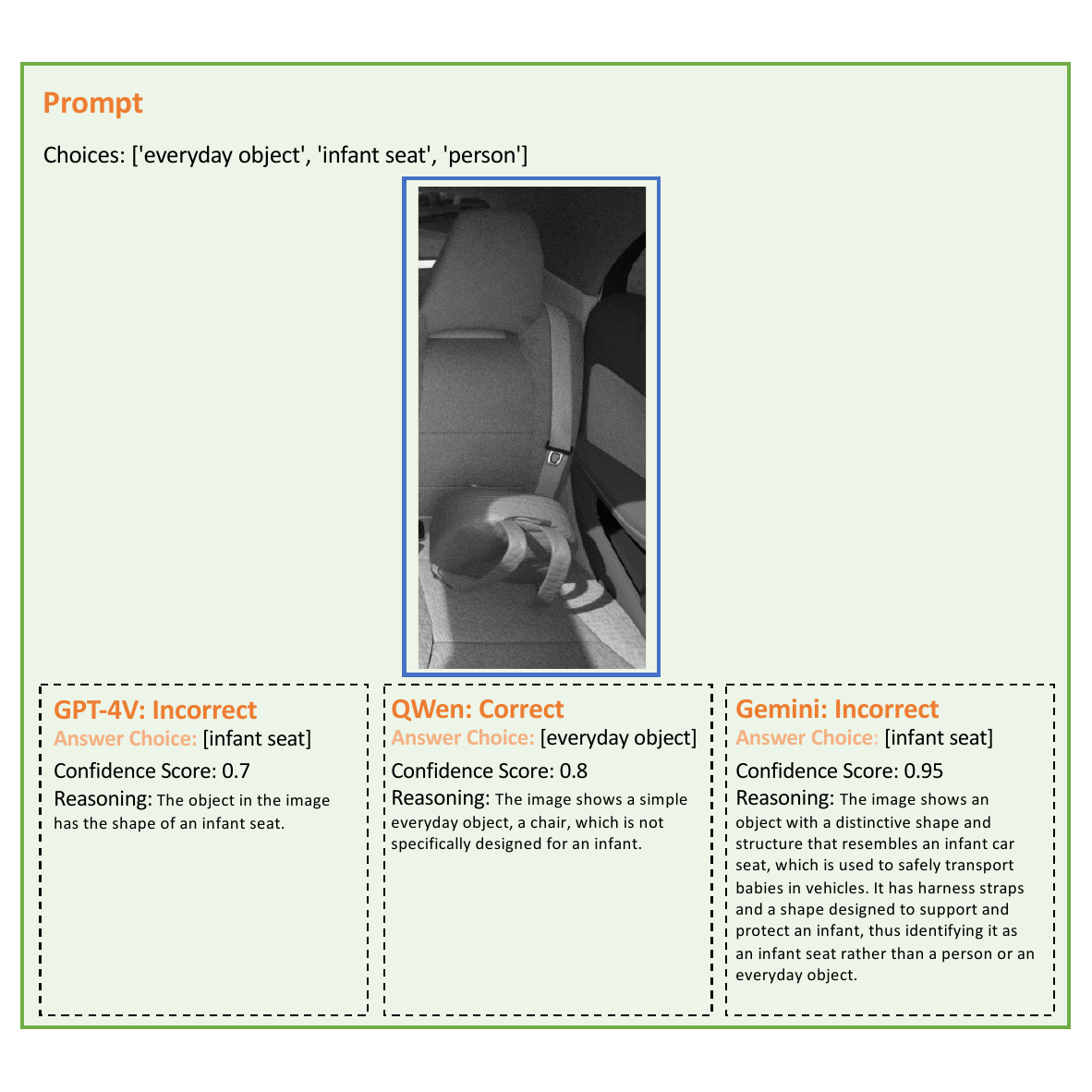}
    \caption{Synthetic and Natural Distribution Shifts in zero-shot generalization: Case 9, analyzing the \textit{everyday object} category in the \textit{aclass} domain of the SVIRO dataset. In this case, QWen predicts correctly while GPT-4V and Gemini predict incorrectly. QWen briefly describes the object in the image as an everyday object rather than an infant seat, a description that is overly simplistic and lacks persuasiveness. GPT-4V believes the image depicts an infant seat. Gemini determines that the object in the image is an infant seat, based on its distinctive shape, structure, and harness straps.}
\end{figure*}

\begin{figure*}[ht]
    \centering
    \label{fig:}
    \includegraphics[width=1\textwidth]{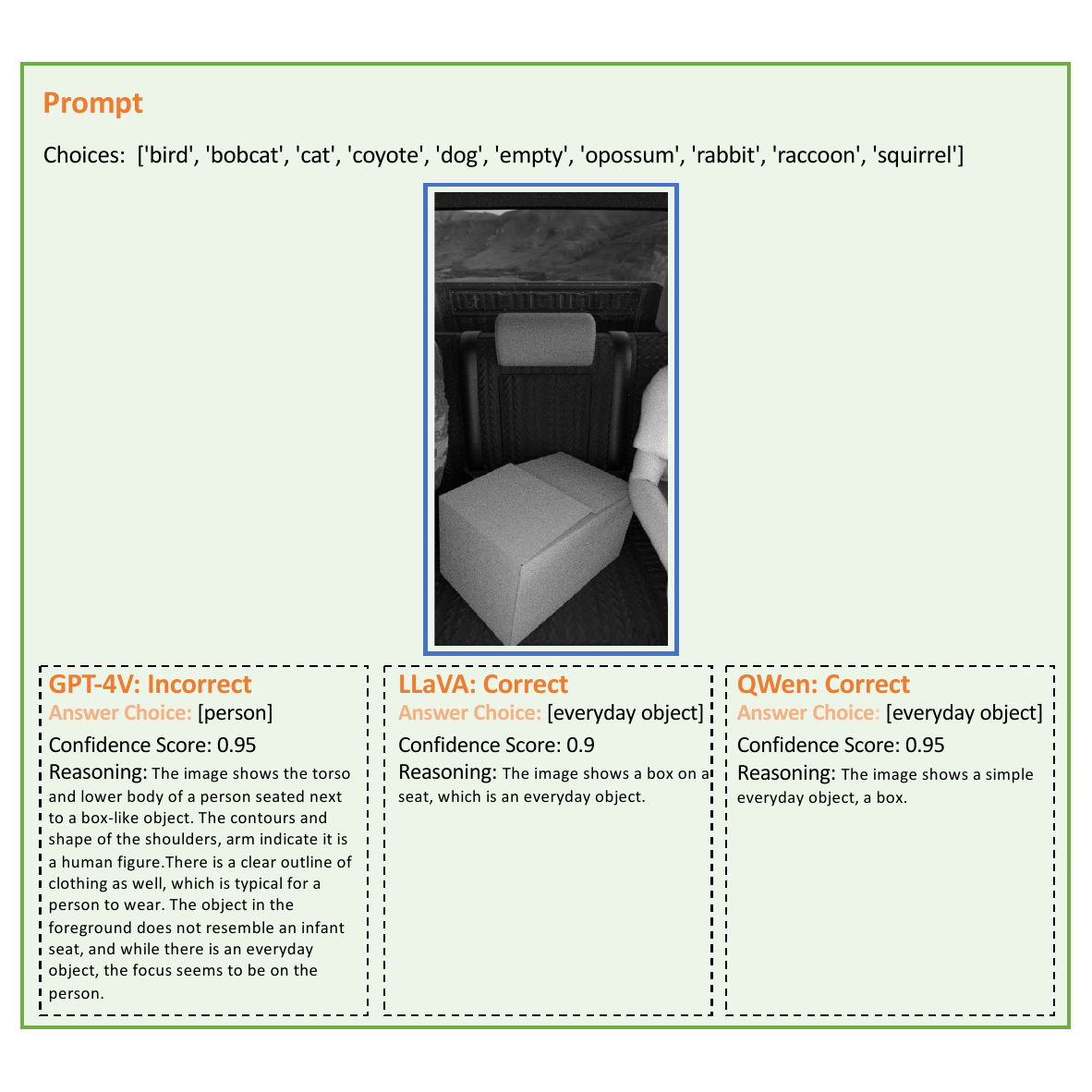}
    \caption{Synthetic and Natural Distribution Shifts in zero-shot generalization: Case 10, analyzing the \textit{everyday object} category in the \textit{hilux} domain of the SVIRO dataset. In this case, both LLaVA and QWen predict correctly, but QWen gives a higher confidence score of 0.95. LLaVA and QWen provide simple descriptions of the content in the image. In contrast, GPT-4V classifies the image as \textit{person} based on the appearance of a segment of a person's arm in the picture. However, GPT-4V's focus is on the box on the seat, not the partial view of the person's arm.}
\end{figure*}

\begin{figure*}[ht]
    \centering
    \label{fig:}
    \includegraphics[width=1\textwidth]{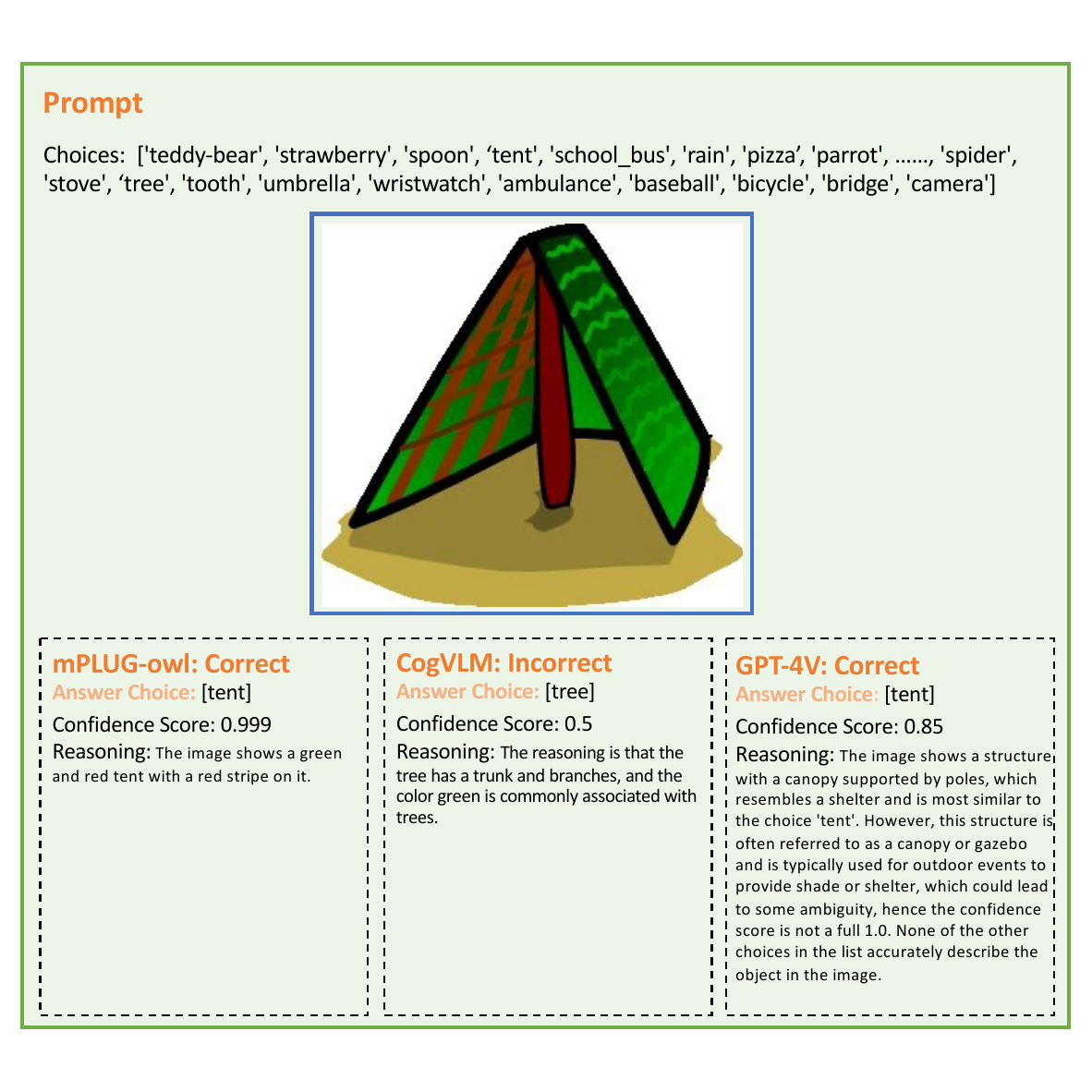}
    \caption{Synthetic and Natural Distribution Shifts in zero-shot generalization: Case 11, analyzing the \textit{tent} category in the \textit{clipart} domain of the DomainNet dataset. In this case, both mPLUG-owl and GPT-4V predict correctly, but mPLUG-owl gives a higher confidence score of 0.999 and GPT-4V gives more detailed explanations. GPT-4V also explains that due to the ambiguity of the concept, its confidence score is not 1. CogVLM mistakenly identifies the green tent and the brown supports as the branches and trunk of a tree.}
\end{figure*}

\begin{figure*}[ht]
    \centering
    \label{fig:}
    \includegraphics[width=1\textwidth]{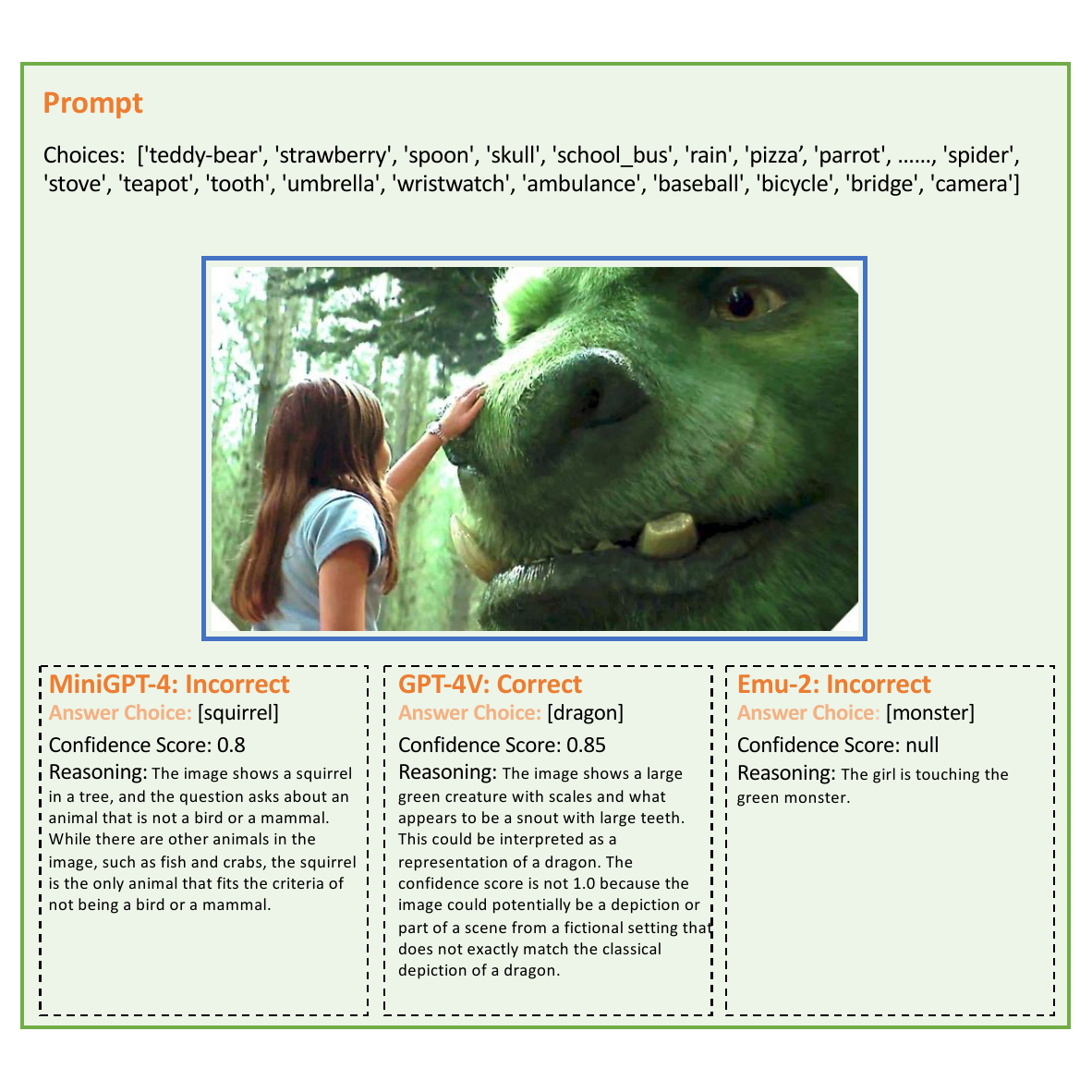}
    \caption{Synthetic and Natural Distribution Shifts in zero-shot generalization: Case 12, analyzing the \textit{dragon} category in the real domain of the DomainNet dataset. In this case, GPT-4V
    predicts correctly while MiniGPT-4 and Emu-2 predict incorrectly. MiniGPT-4 mistakenly identifies the green, furry dragon in the image as a squirrel in a tree. Emu-2 classifies the dragon in the picture as a monster but does not provide a confidence score. GPT-4V gives a reasoning that details specific features of the object in the image, such as its large size and scales, thus determining the object in the image to be a dragon.}
\end{figure*}

\begin{figure*}[ht]
    \centering
    \label{fig:}
    \includegraphics[width=1\textwidth]{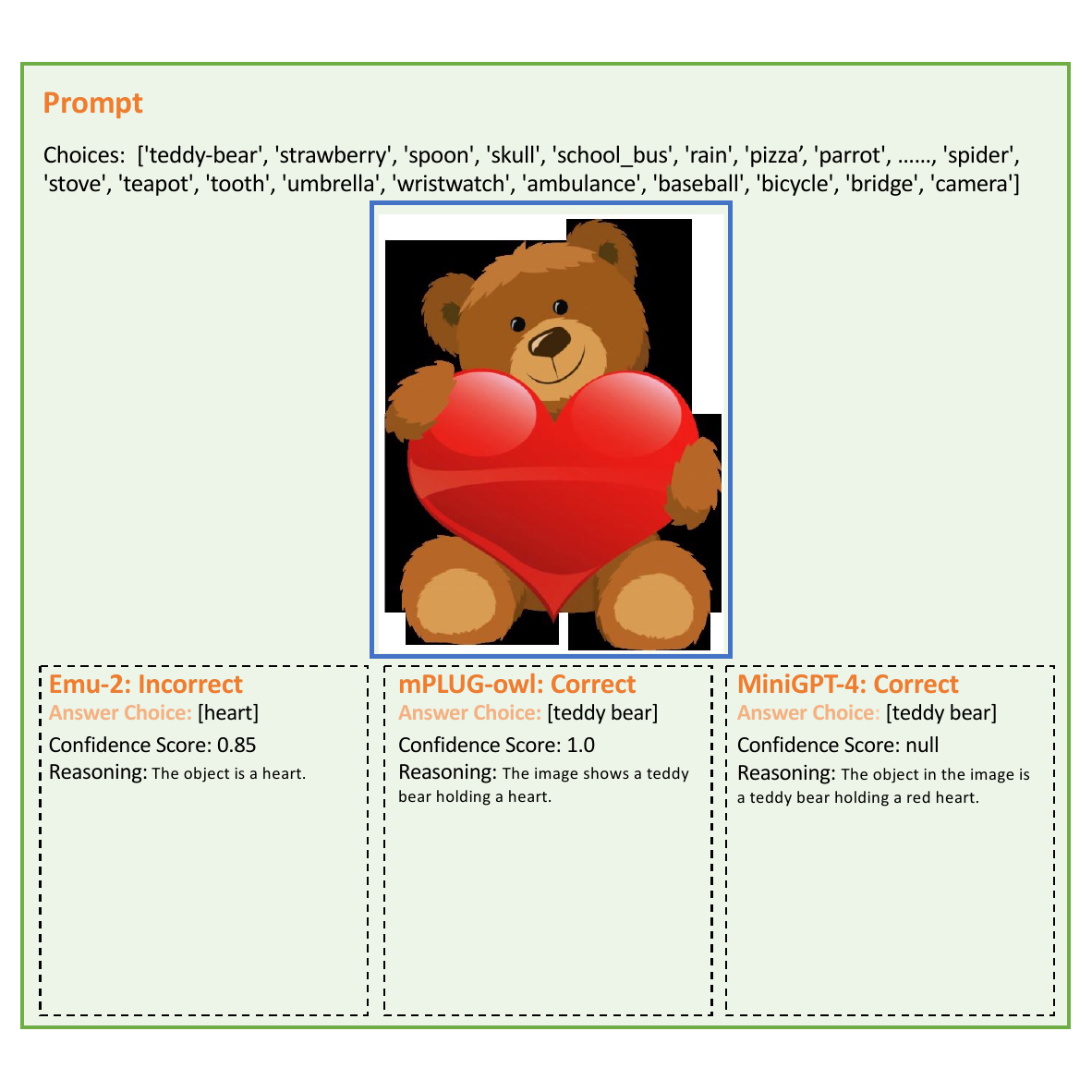}
    \caption{Synthetic and Natural Distribution Shifts in zero-shot generalization: Case 13, analyzing the \textit{teddy-bear} category in the \textit{clipart} domain of the DomainNet dataset. In this case, both mPLUG-owl and MiniGPT-4 predict correctly, but mPLUG-owl gives a higher confidence score with 1.0 and MiniGPT-4 does not provide the confidence score. Emu-2, mPLUG-owl, and MiniGPT-4 all provide accurate descriptions of the image. Emu-2 focuses on the heart held by the teddy bear, leading to an incorrect prediction. }
\end{figure*}

\begin{figure*}[ht]
    \centering
    \label{fig:}
    \includegraphics[width=1\textwidth]{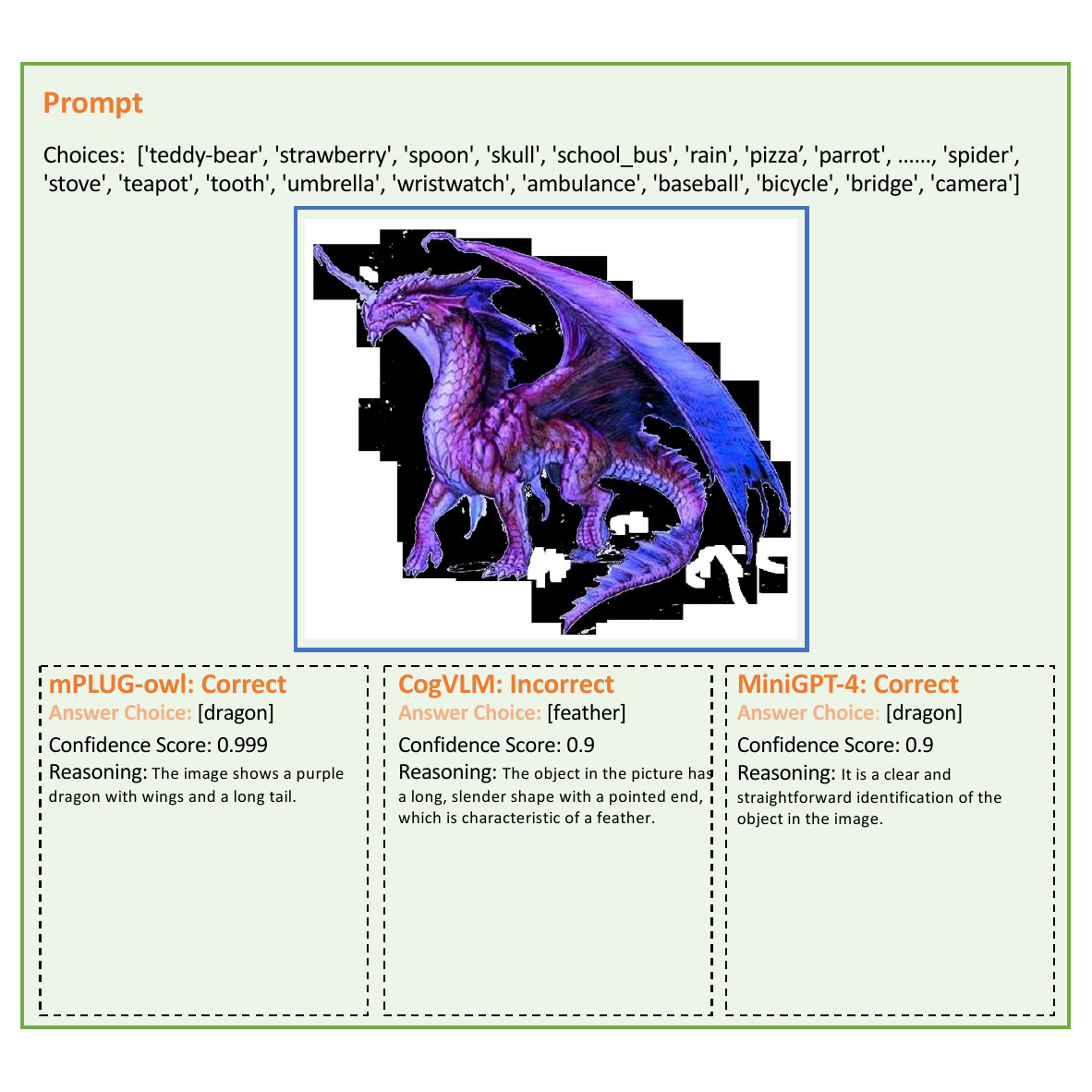}
    \caption{Synthetic and Natural Distribution Shifts in zero-shot generalization: Case 14, analyzing the \textit{dragon} category in the \textit{real} domain of the DomainNet dataset. In this case, both mPLUG-owl and MiniGPT-4 predict correctly, but mPLUG-owl gives a higher confidence score of 0.999. While MiniGPT-4 makes a correct prediction, the rationale it provides is not logical. CogVLM classifies the object in the image as a feather based on the shape of the object in the picture such as a long, slender shape with a pointed end.}
\end{figure*}

\begin{figure*}[ht]
    \centering
    \label{fig:}
    \includegraphics[width=1\textwidth]{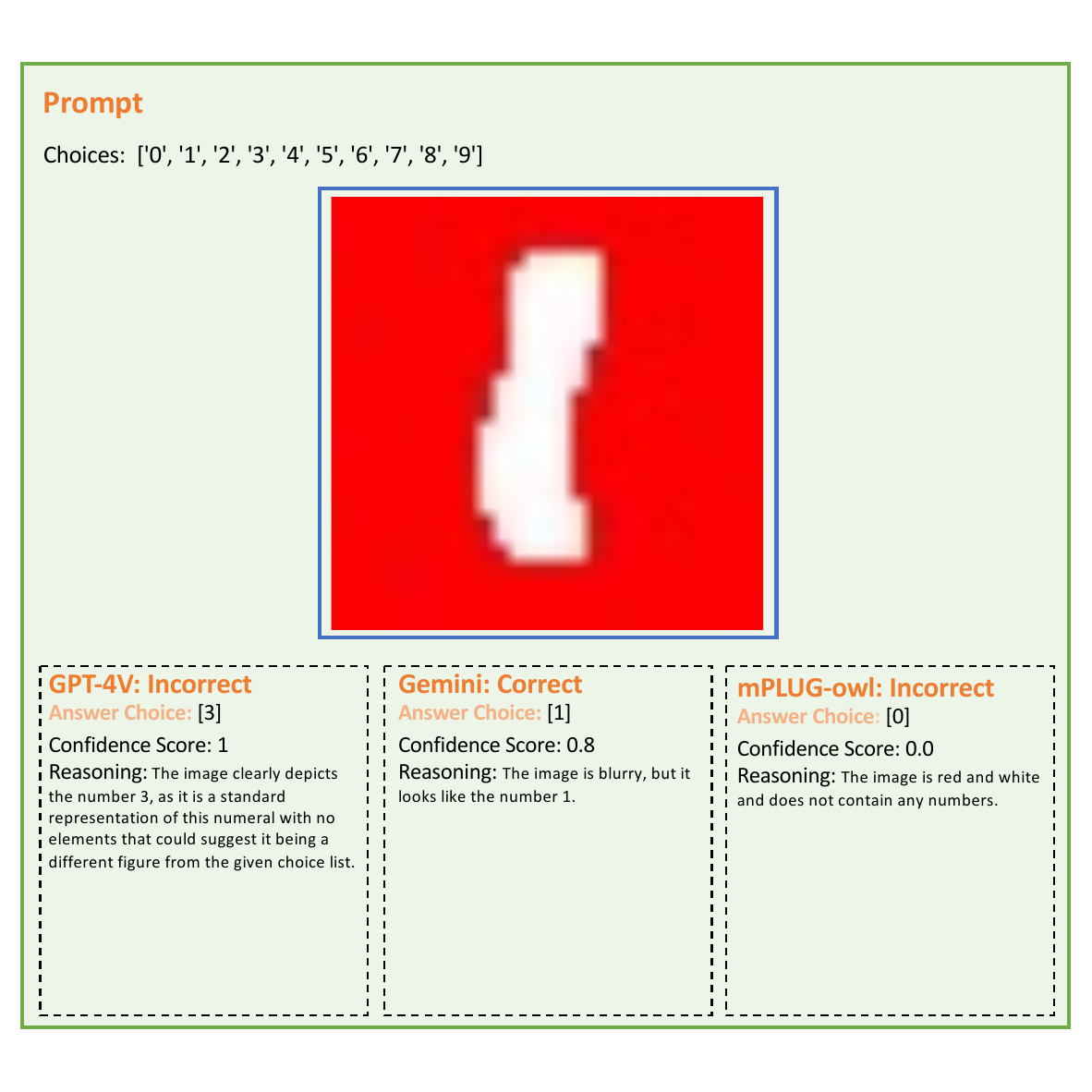}
    \caption{Synthetic and Natural Distribution Shifts in zero-shot generalization: Case 15, analyzing the \textit{1} category in the \textit{test} domain of the CMNIST dataset. In this case, GPT-4V confidently predicts incorrectly, classifying the digit in the image as \textit{3} with a confidence score of 1.0. Despite the image being blurred, Gemini still predicts correctly. mPLUG-owl chose the digit \textit{0} because it believes the image contained no digits.}
\end{figure*}

\begin{figure*}[ht]
    \centering
    \label{fig:}
    \includegraphics[width=1\textwidth]{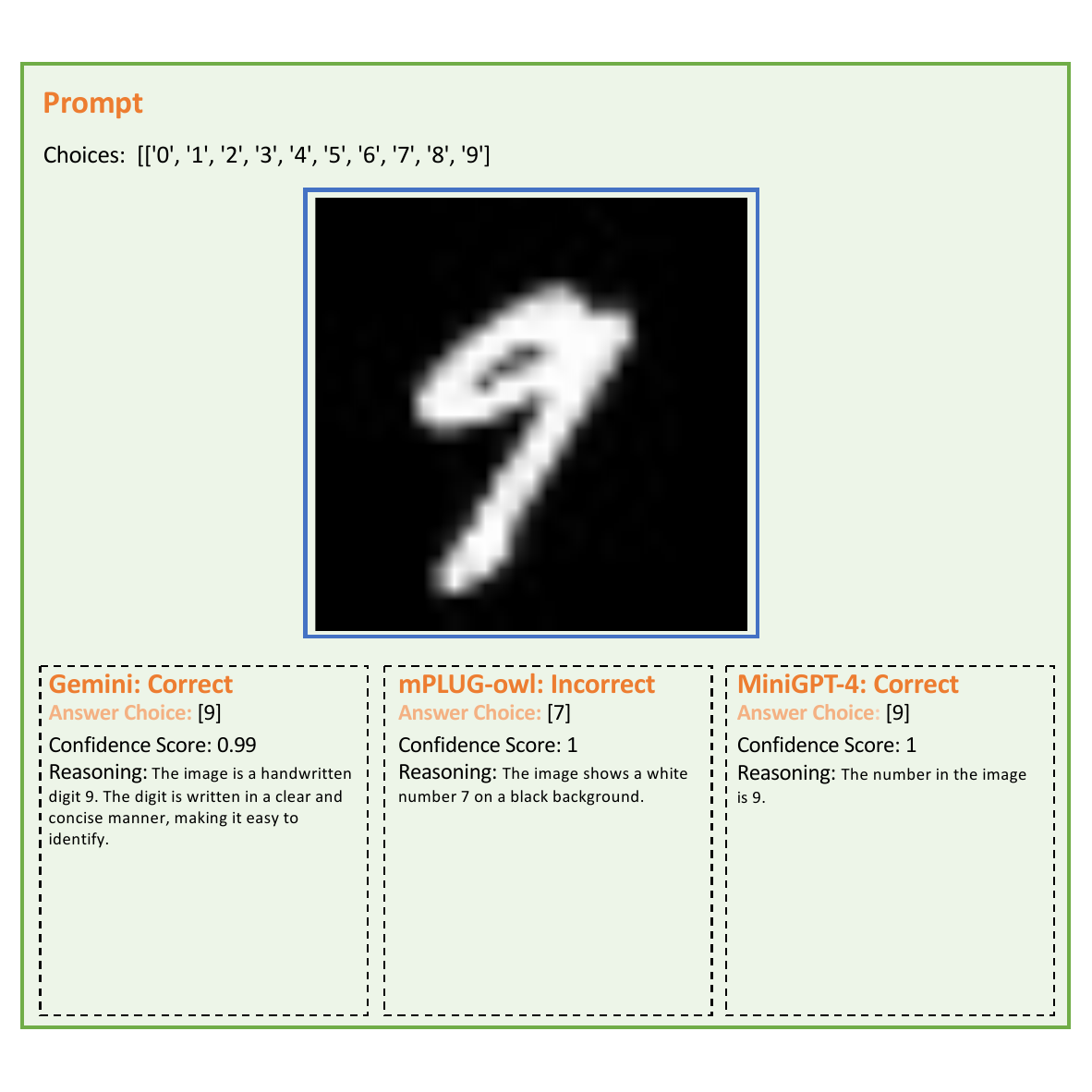}
    \caption{Synthetic and Natural Distribution Shifts in zero-shot generalization: Case 16, analyzing the \textit{9} category in the \textit{test} domain of the RNIST dataset. In this case, Gemini successfully identifies the image as a handwritten digit and predicts it correctly. mPLUG-owl incorrectly predicts the digit as \textit{7}. MiniGPT-4 makes a correct prediction, believing the digit to be \textit{9}, with an insufficient rationale provided.}
\end{figure*}

\begin{figure*}[ht]
    \centering
    \label{fig:}
    \includegraphics[width=1\textwidth]{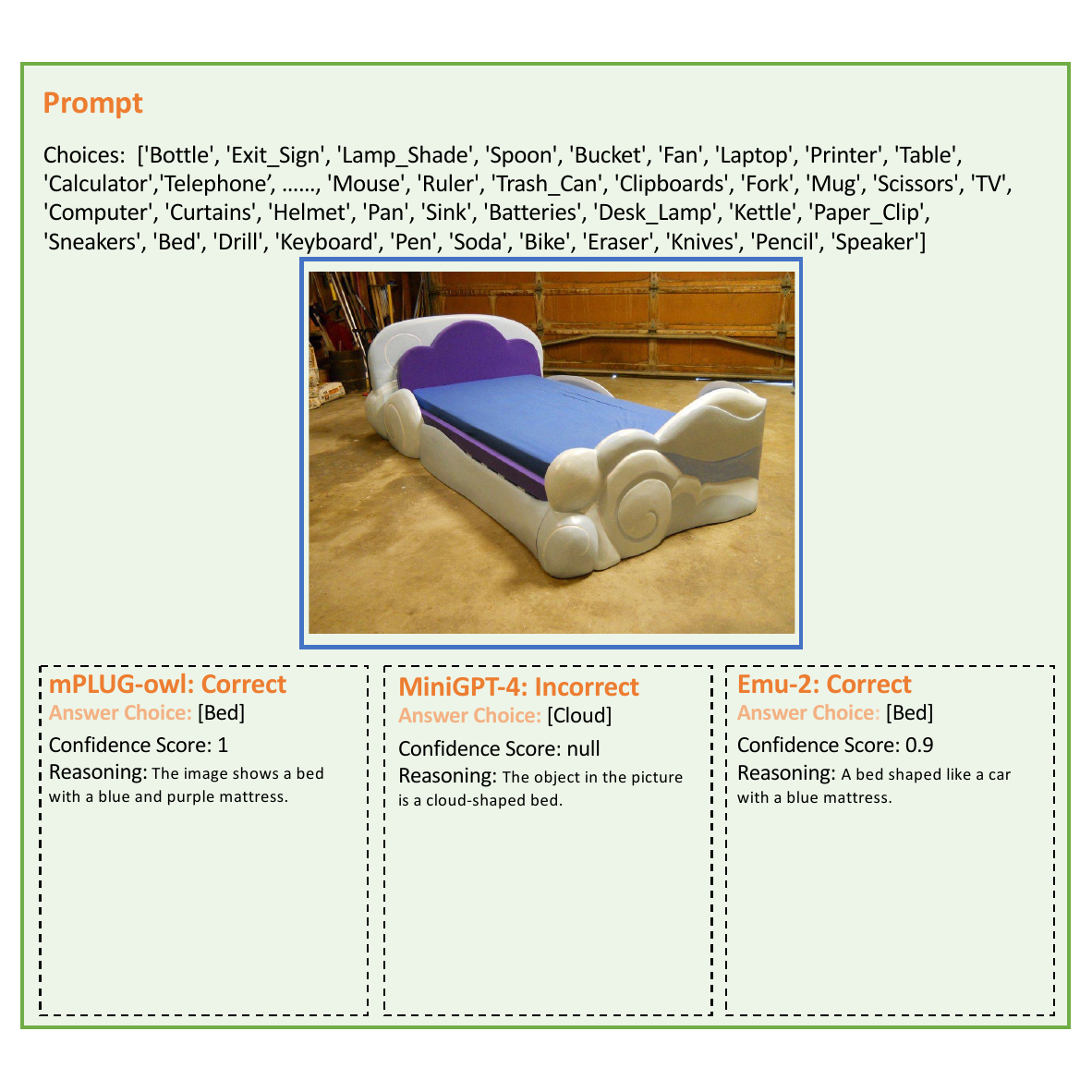}
    \caption{Synthetic and Natural Distribution Shifts in zero-shot generalization: Case 17, analyzing the \textit{Bed} category in the \textit{Art} domain of the OfficeHome dataset. In this case, both mPLUG-owl and Emu-2 predict correctly, but mPLUG-owl gives a higher confidence score of 1.0. The rationales provided by mPLUG-owl, MiniGPT-4, and Emu-2 are general descriptions of the image, which are overall correct. mPLUG-owl describes the color of the bed. MiniGPT-4 believes the image depicts a cloud-shaped bed but incorrectly categorizes the object in the image as a \textit{cloud}. Emu-2 considers the image to be of a car-shaped bed and, although unsure about the shape, correctly predicts the category as a \\textit{bed}.}
\end{figure*}

\begin{figure*}[ht]
    \centering
    \label{fig:}
    \includegraphics[width=1\textwidth]{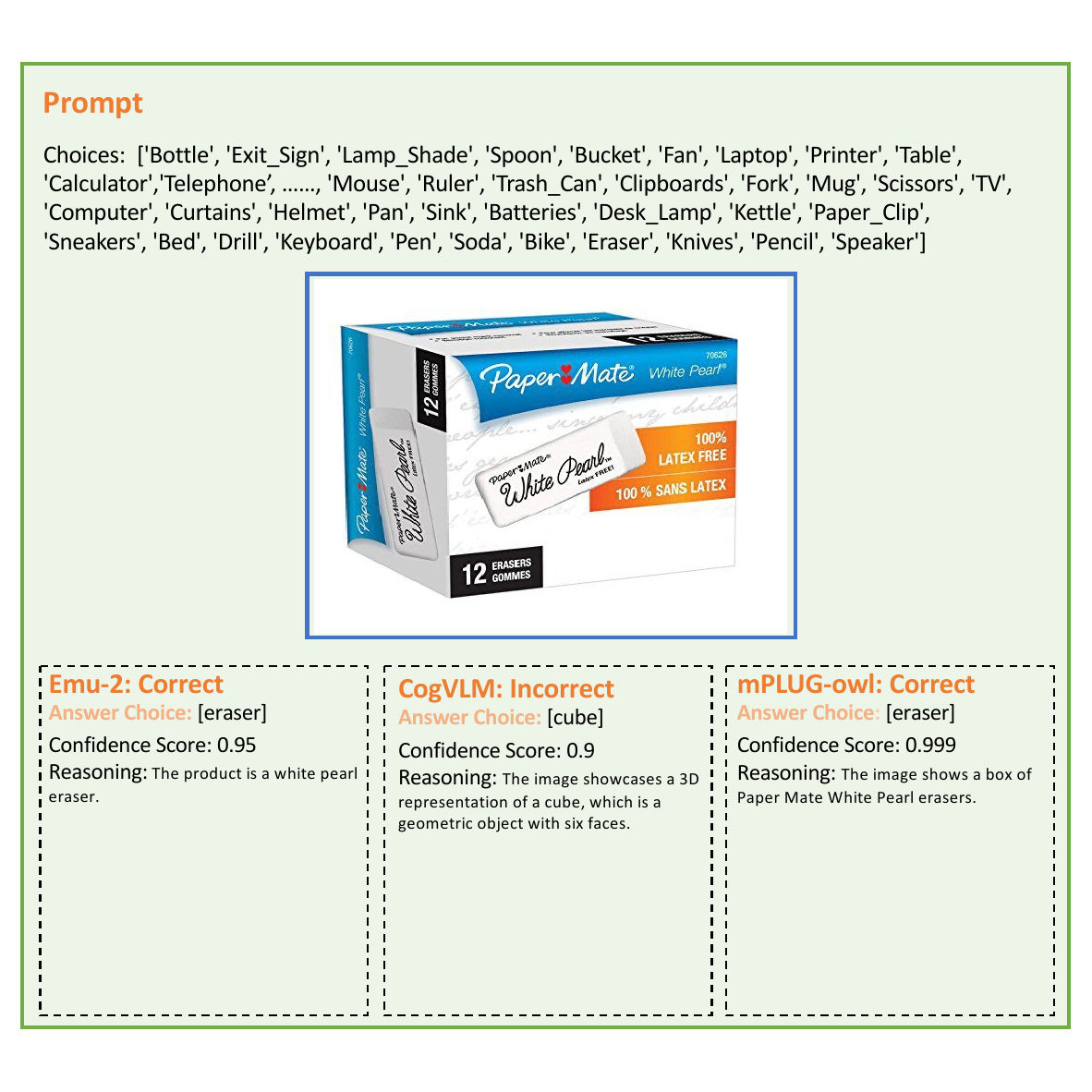}
    \caption{Synthetic and Natural Distribution Shifts in zero-shot generalization: Case 18, analyzing the \textit{Eraser} category in the \textit{Product} domain of the OfficeHome dataset. In this case, both Emu-2 and mPLUG-owl predict correctly, but mPLUG-owl gives a higher confidence score of 0.999. Emu-2 and mPLUG-owl correctly classify the object in the image as an \textit{eraser}, with mPLUG-owl providing a more detailed rationale. The image shows the packaging of the \textit{eraser}, leading CogVLM to incorrectly categorize the object in the image as a cube.}
\end{figure*}

\begin{figure*}[ht]
    \centering
    \label{}
    \includegraphics[width=1\textwidth]{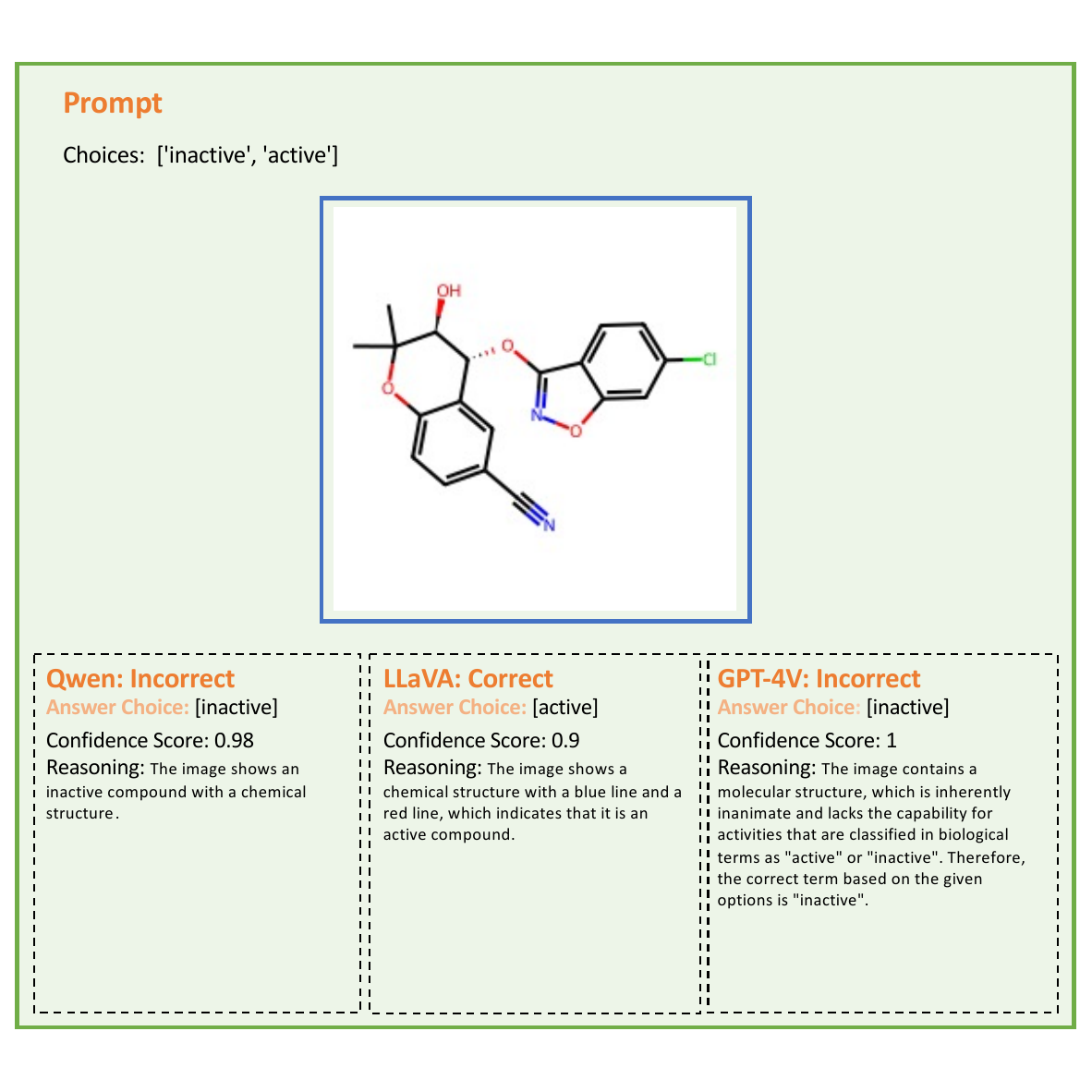}
    \caption{Domain-Specific Distribution Shift in zero-shot generalization: Case 1, analyzing the \textit{active} category in the \textit{ID-25} domain of the DrugOOD\_Assay dataset. In this case, both QWen and GPT-4V erroneously offer predictions with a high degree of confidence, whereas LLaVA delivers accurate predictions also with high confidence. QWen and GPT-4V can identify the image as chemical structures, but they predict the category incorrectly. LLaVA provides accurate predictions, but the reasoning lacks validity. Determining the \textit{active} or \textit{inactive} status of compounds merely based on the bond color is erroneous.}
    \label{fig:app-error-2}
\end{figure*}

\begin{figure*}[ht]
    \centering
    \label{fig:}
    \includegraphics[width=1\textwidth]{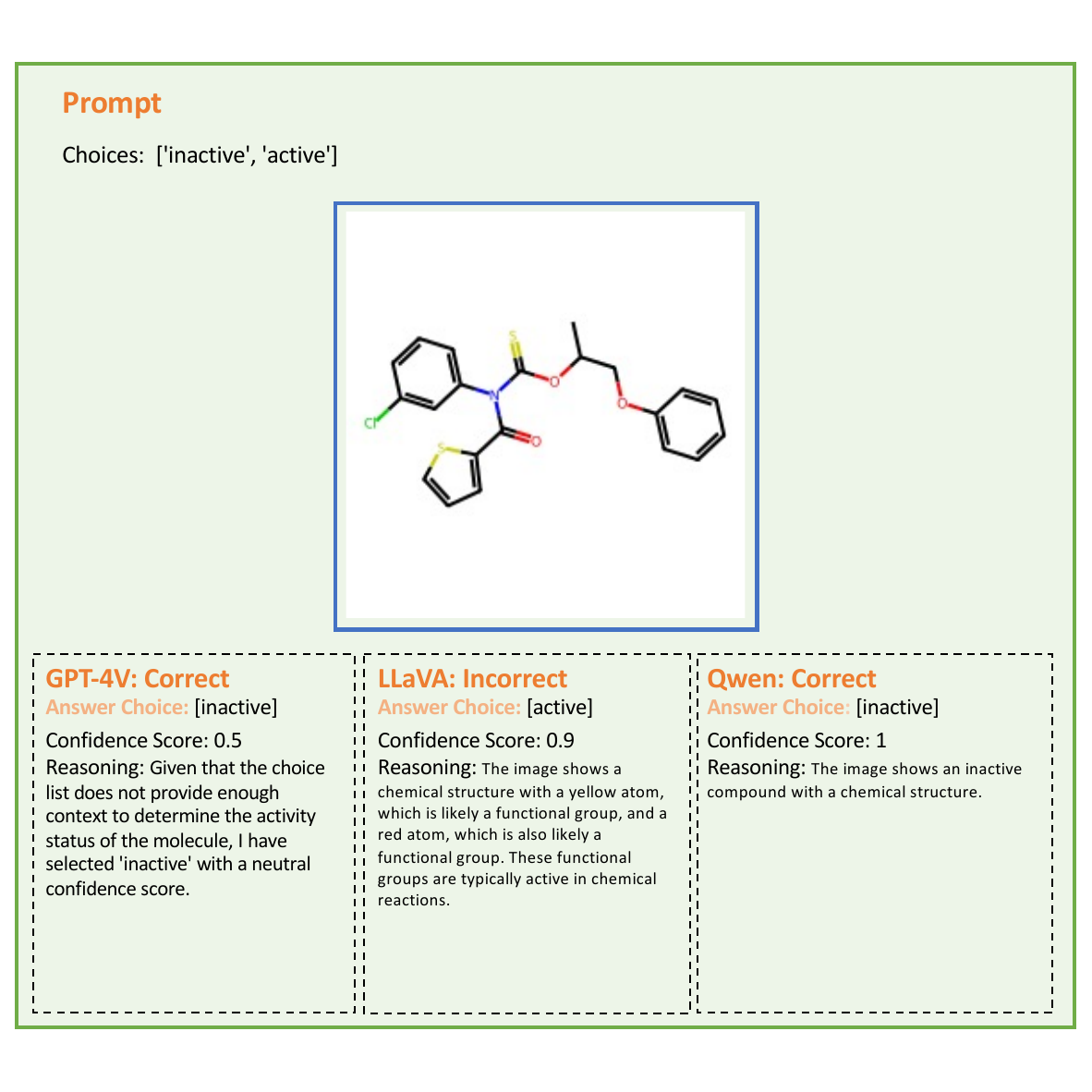}
    \caption{Domain-Specific Distribution Shift in zero-shot generalization: Case 2, analyzing the \textit{inactive} category in the \textit{ID-0} domain of the DrugOOD\_Assay dataset. In this case, LLaVA incorrectly predicts with high confidence. GPT-4V and QWen predict correctly, and Qwen gives a confidence score of 1.00 and GPT-4V correctly predicts with a low confidence score. It is reasonable to assert that although GPT-4V's category prediction is correct, it can not provide an answer with a high degree of confidence due to the lack of additional information, as evidenced by the rationale provided. Despite QWen delivering accurate predictions, its reasoning lacks support and thus seems unreliable. Furthermore, LLaVA's classification of compounds as \textit{active} or \textit{inactive} based on the color of the bonds is incorrect.}
\end{figure*}

\begin{figure*}[ht]
    \centering
    \label{fig:}
    \includegraphics[width=1\textwidth]{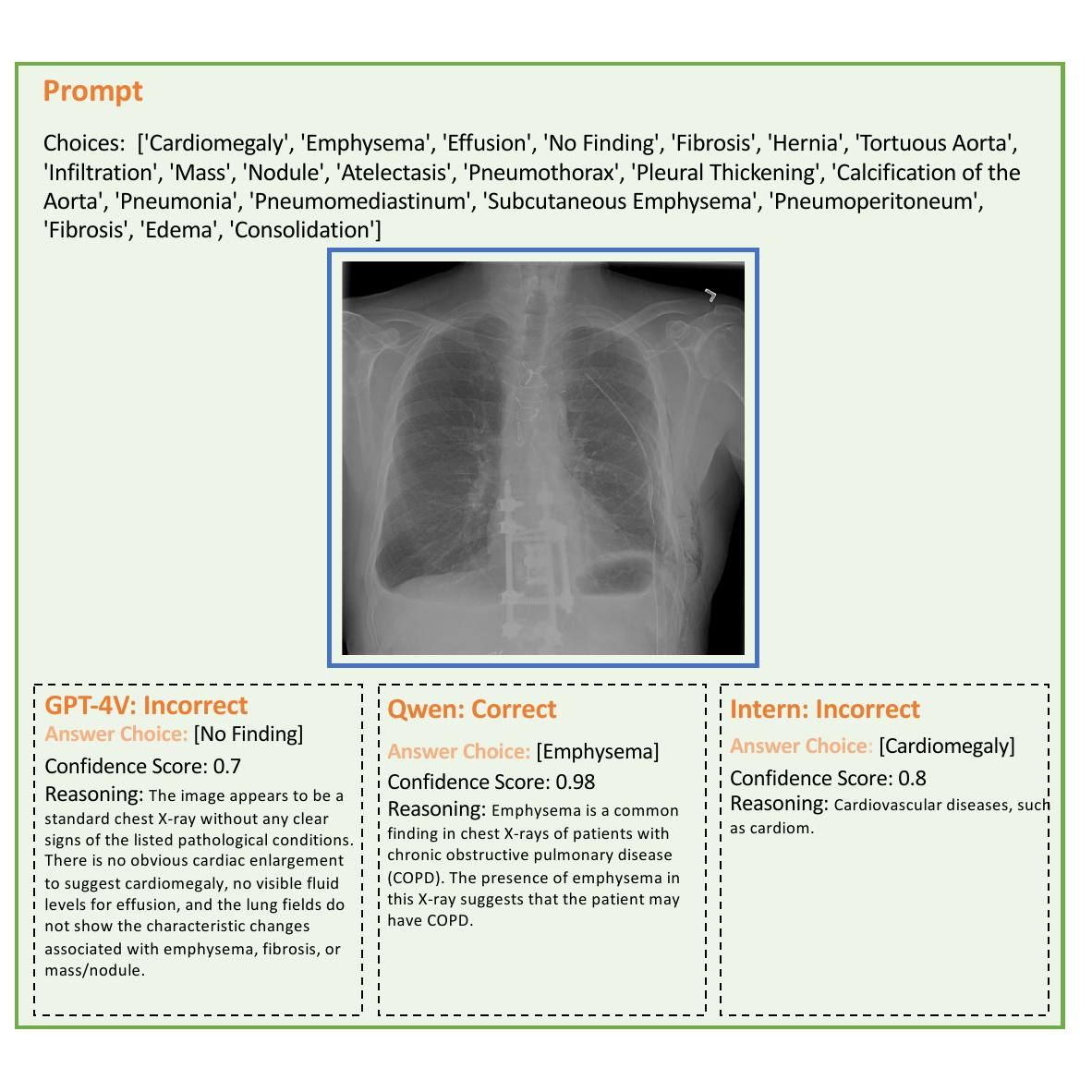}
    \caption{Domain-Specific Distribution Shift in zero-shot generalization: Case 3, analyzing the \textit{Emphysema} category in the \textit{Chest} domain of the NIH-Chest dataset. In this case, QWen predicts correctly while GPT-4V and Intern predict incorrectly. GPT-4V provides a detailed rationale that, despite its intricacy, is incorrect, resulting in a misclassification as \textit{No-Finding}. The reason given by Intern is rough and Intern erroneously predicts a broad category of \textit{Cardiomegaly}. The explanation of QWen lacks detailed justification, merely suggesting that the patient in the X-ray might suffer from emphysema.}
\end{figure*}

\begin{figure*}[ht]
    \centering
    \label{fig:}
    \includegraphics[width=1\textwidth]{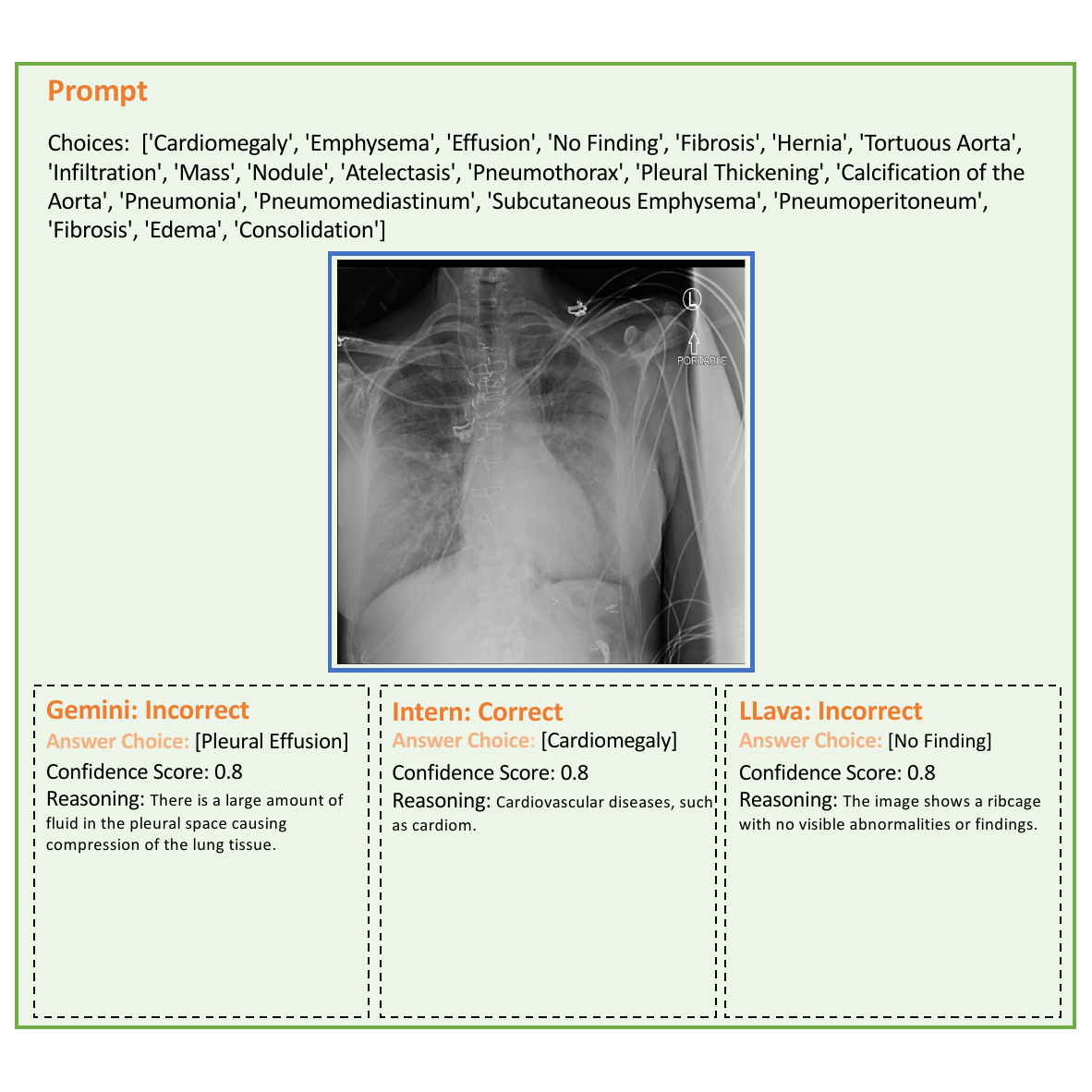}
    \caption{Domain-Specific Distribution Shift in zero-shot generalization: Case 4, analyzing the \textit{Cardiomegaly} category in the \textit{Chest} domain of the NIH-Chest dataset. In this case, Intern predicts correctly while Gemini and LLaVA predict incorrectly. The reasoning given by Gemini is incorrect, however, the rationale provided is detailed and plausible. LLaVA assesses the X-ray as exhibiting no abnormalities. Although Intern's prediction is accurate, the justification offered is not credible, seemingly opting for the most common disease by default.}
\end{figure*}

\begin{figure*}[ht]
    \centering
    \label{fig:}
    \includegraphics[width=1\textwidth]{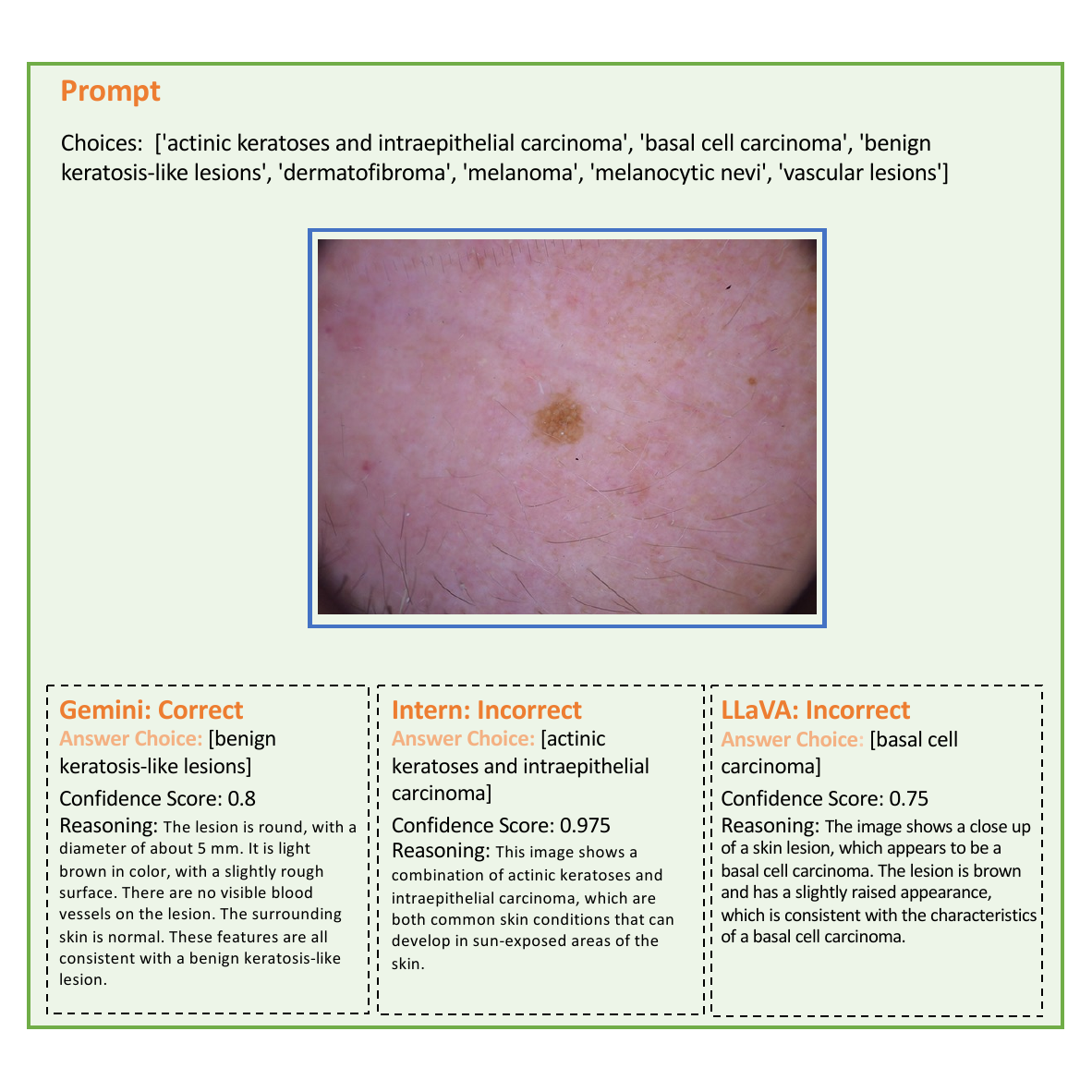}
    \caption{Domain-Specific Distribution Shift in zero-shot generalization: Case 5, analyzing the \textit{benign keratosis-like lesions} category in the \textit{Confocal} domain of the HAM10000 dataset. In this case, Gemini predicts correctly while Intern and LLaVA predict incorrectly. Gemini provides a correct response and detailed specific features of the image, such as a round lesion with a diameter of about 5 mm, light brown in color, with a slightly rough surface. Although LLaVA also roughly describes the image's features, its prediction is incorrect. The rationale provided by Intern is insufficient and wrong.}
\end{figure*}

\begin{figure*}[ht]
    \centering
    \label{fig:}
    \includegraphics[width=1\textwidth]{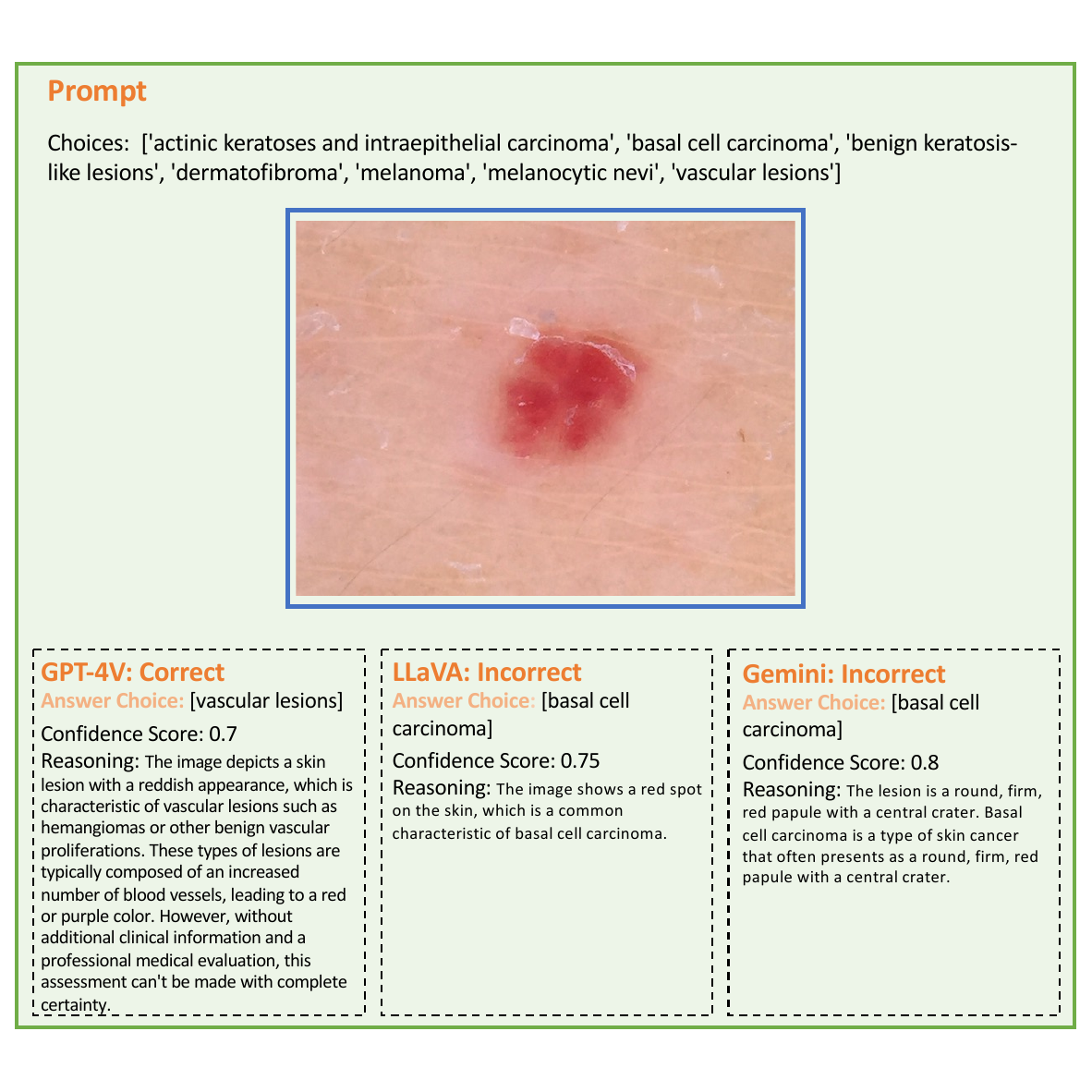}
    \caption{Domain-Specific Distribution Shift in zero-shot generalization: Case 6, analyzing the \textit{vascular lesions} category in the \textit{Consensus} domain of the HAM10000 dataset. In this case, GPT-4V predicts correctly while LLaVA and Gemini predict incorrectly.  GPT-4V primarily describes the features of the image in terms of color and provided an explanation, also rationalizing that the lack of information results in lower confidence, which is a sensible assertion. Although LLaVA and Gemini detail specific features of the image, their category predictions are incorrect.}
\end{figure*}

\begin{figure*}[ht]
    \centering
    \label{fig:}
    \includegraphics[width=1\textwidth]{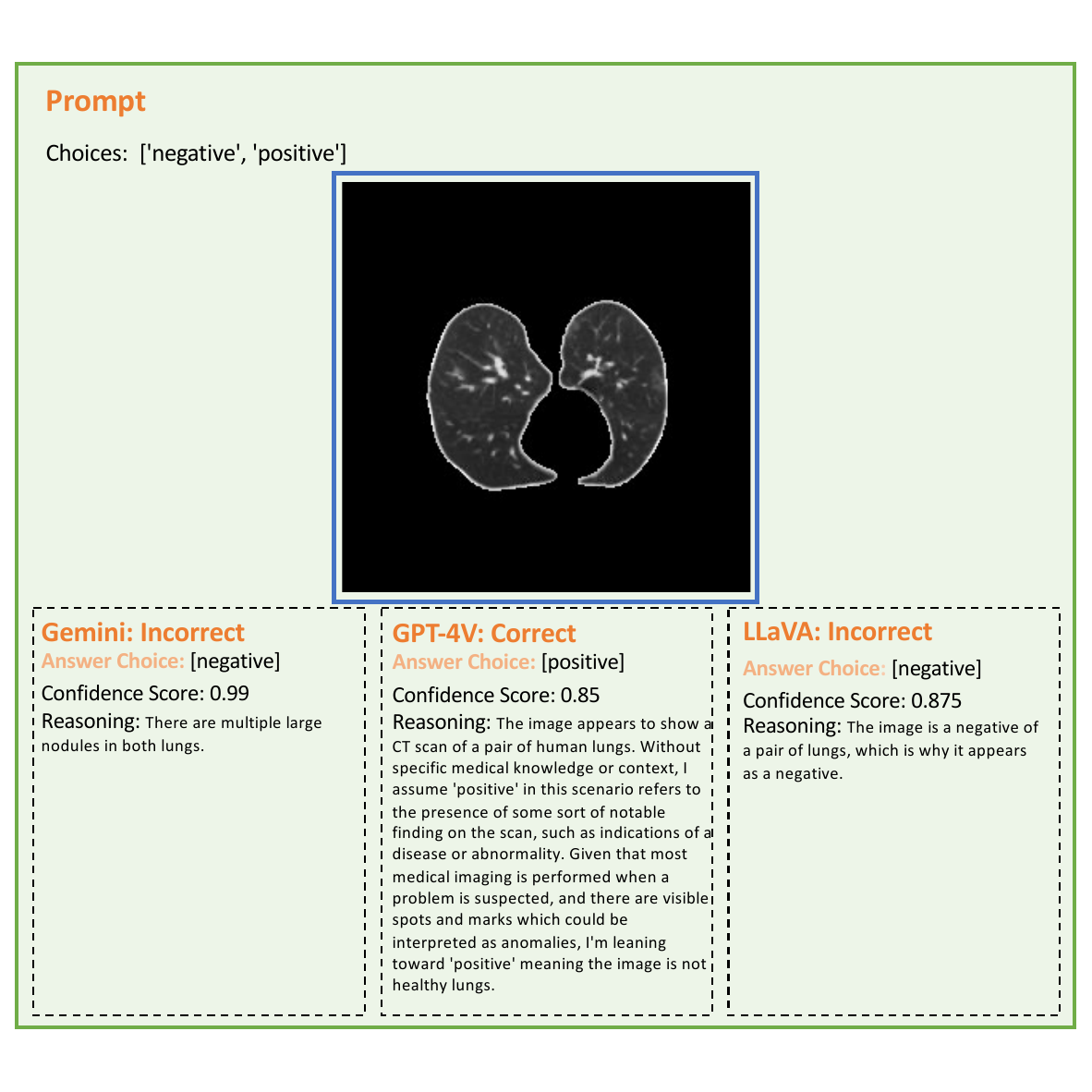}
    \caption{Domain-Specific Distribution Shift in zero-shot generalization: Case 7, analyzing the \textit{positive} category in the \textit{Domain\_0} domain of the CT-XCOV dataset. In this case, \textit{negative} indicates normal and \textit{positivate} indicates suffering from COVID-19. GPT-4V predicts correctly while Gemini and LLaVA predict incorrectly. The reasonings provided by Gemini and LLaVA are vague and do not grasp the specific meanings of the prompt and options. Due to the absence of detailed information about the dataset, their category predictions and the rationale behind them do not align. In its reasoning, GPT-4V elucidates the common meanings of 'negative' and 'positive' within the context of pulmonary CT information and describes the presence of abnormal spots in the CT image, thereby classifying the category as 'positive'.}
\end{figure*}

\begin{figure*}[ht]
    \centering
    \label{fig:}
    \includegraphics[width=1\textwidth]{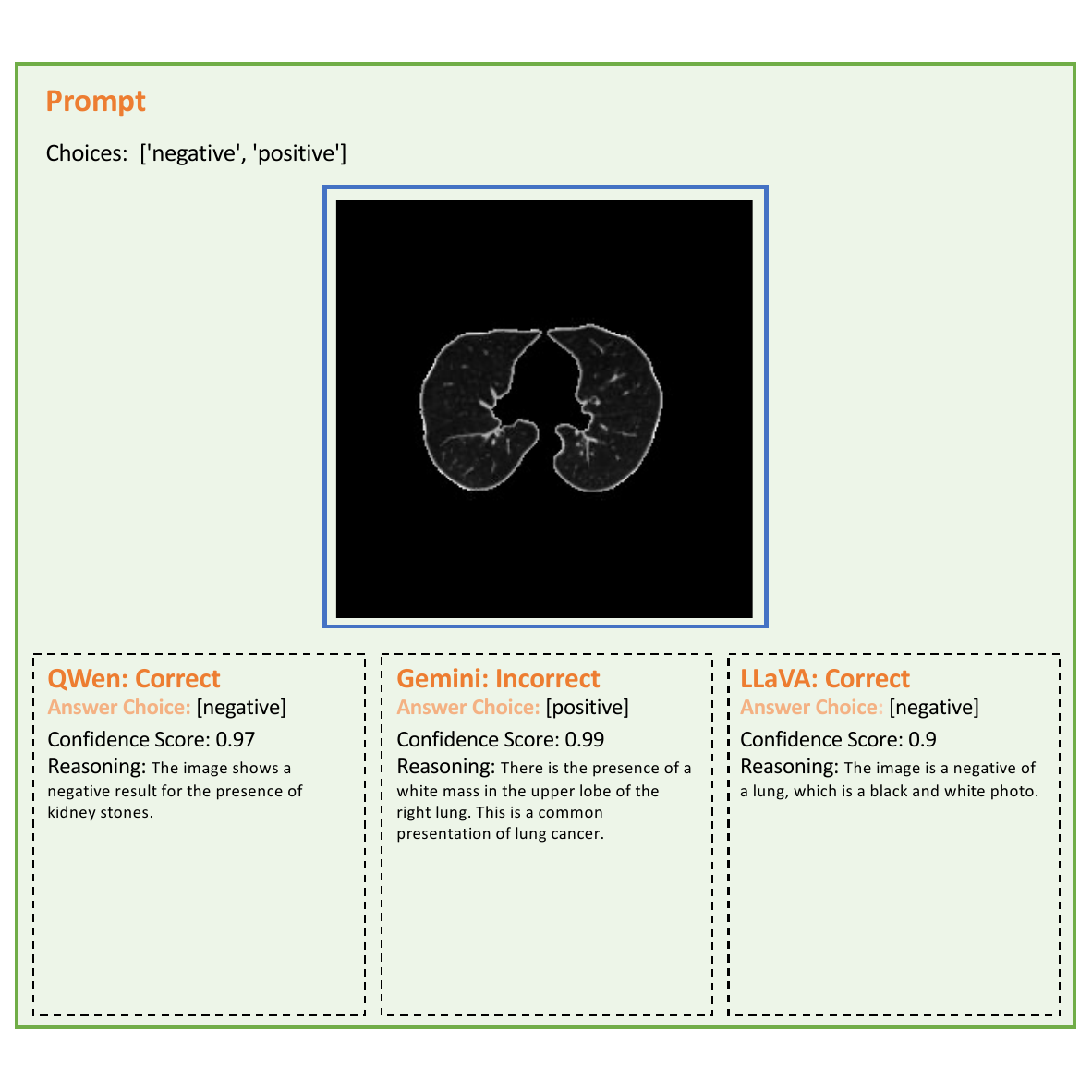}
    \caption{Domain-Specific Distribution Shift in zero-shot generalization: Case 8, analyzing the \textit{negative} category in the \textit{Domain\_0} domain of the CT-XCOV dataset. In this case, both QWen and LLaVA predict correctly, but QWen gives a higher confidence score. Although QWen and LLaVA make correct category predictions, the lack of specific information about the dataset suggests that the predictions and their justifications do not align, with both models providing only cursory reasons. Gemini describes the features of the image more accurately, interpreting it as indicative of lung cancer, but due to missing information, its category prediction is incorrect.}
\end{figure*}

\begin{figure*}[ht]
    \centering
    \label{fig:}
    \includegraphics[width=1\textwidth]{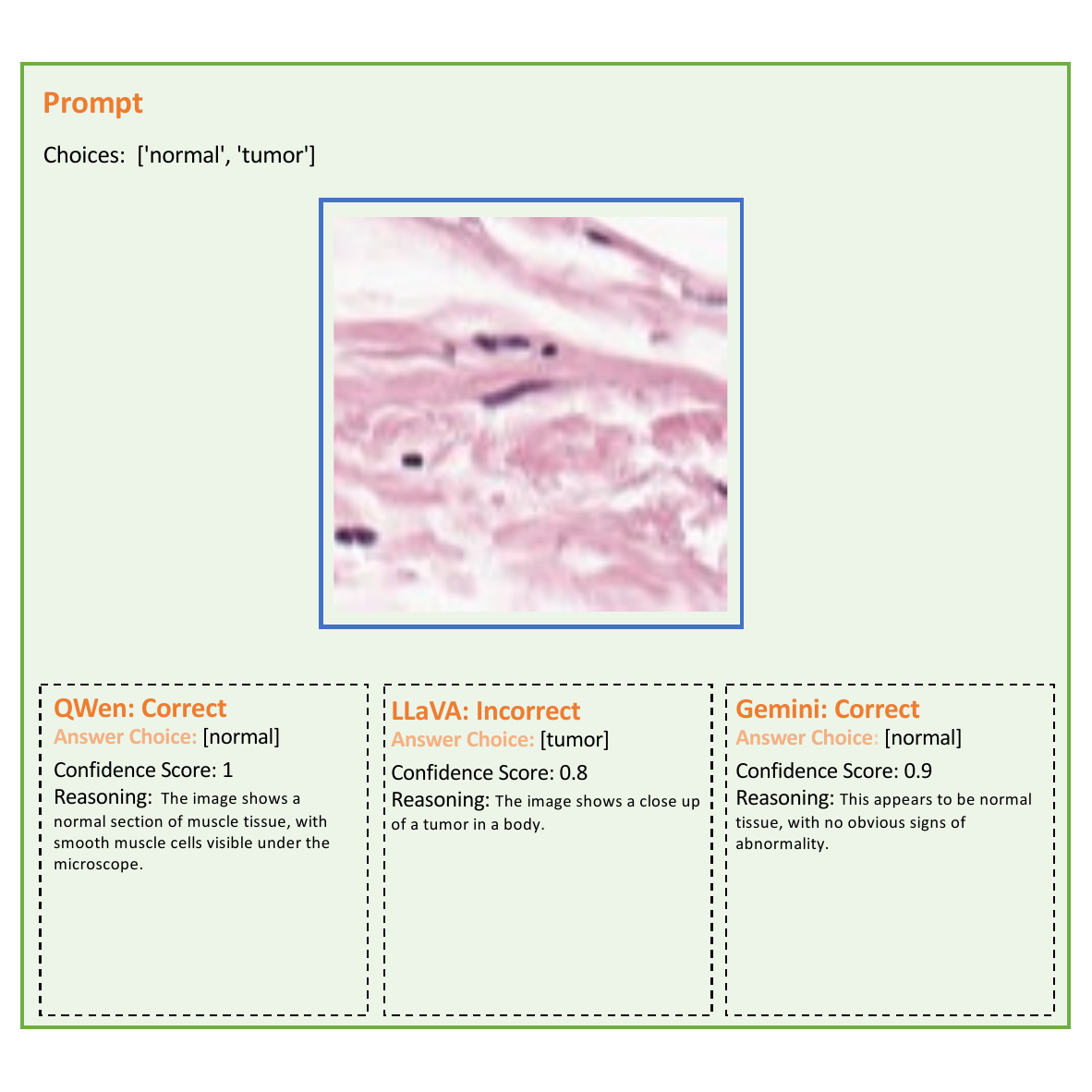}
    \caption{Domain-Specific Distribution Shift in zero-shot generalization: Case 9, analyzing the \textit{normal} category in the \textit{hospital\_0} domain of the Camelyon17 dataset. In this case, both QWen and Gemini predict correctly, where QWen gives a higher confidence score of 1.00. Although QWen and Gemini both make correct predictions with very high confidence, the reasons provided by both models are relatively simplistic and do not elaborate on the rationale. LLaVA makes an incorrect prediction and fails to identify normal cells.}
\end{figure*}

\begin{figure*}[ht]
    \centering
    \label{fig:}
    \includegraphics[width=1\textwidth]{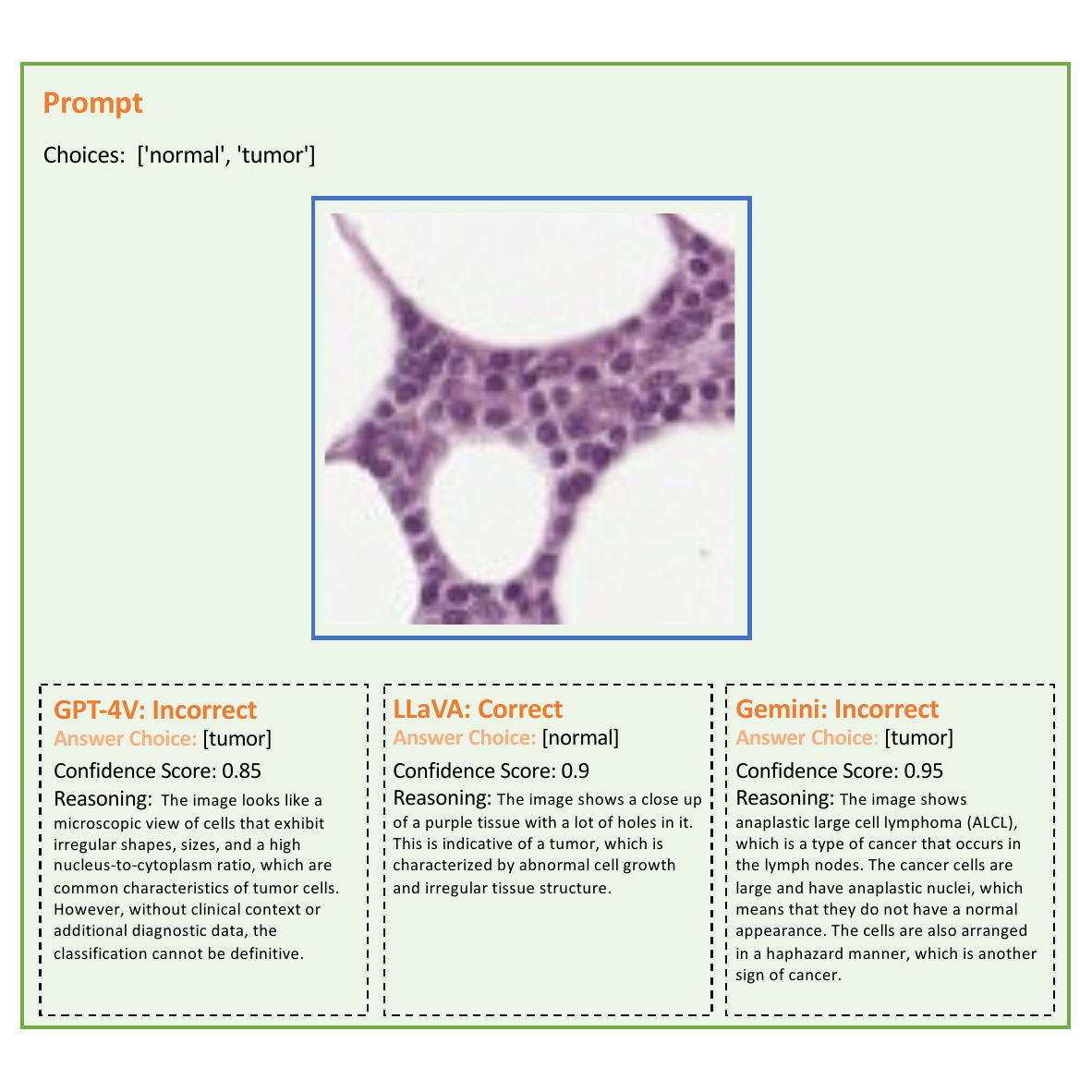}
    \caption{Domain-Specific Distribution Shift in zero-shot generalization: Case 10, analyzing the \textit{normal} category in the \textit{Hospital\_1} domain of the Camelyon17 dataset. In this case, LLaVA predicts correctly while GPT-4V and Gemini predict incorrectly. Although LLaVA makes a correct prediction, the detailed cell features described in the image suggest the other category, tumor, diverging from its given category prediction. Both GPT-4V and Gemini provide detailed descriptions of the image's information and cell features, yet their answer choices are incorrect.}
\end{figure*}

\begin{figure*}[ht]
    \centering
    \label{fig:}
    \includegraphics[width=1\textwidth]{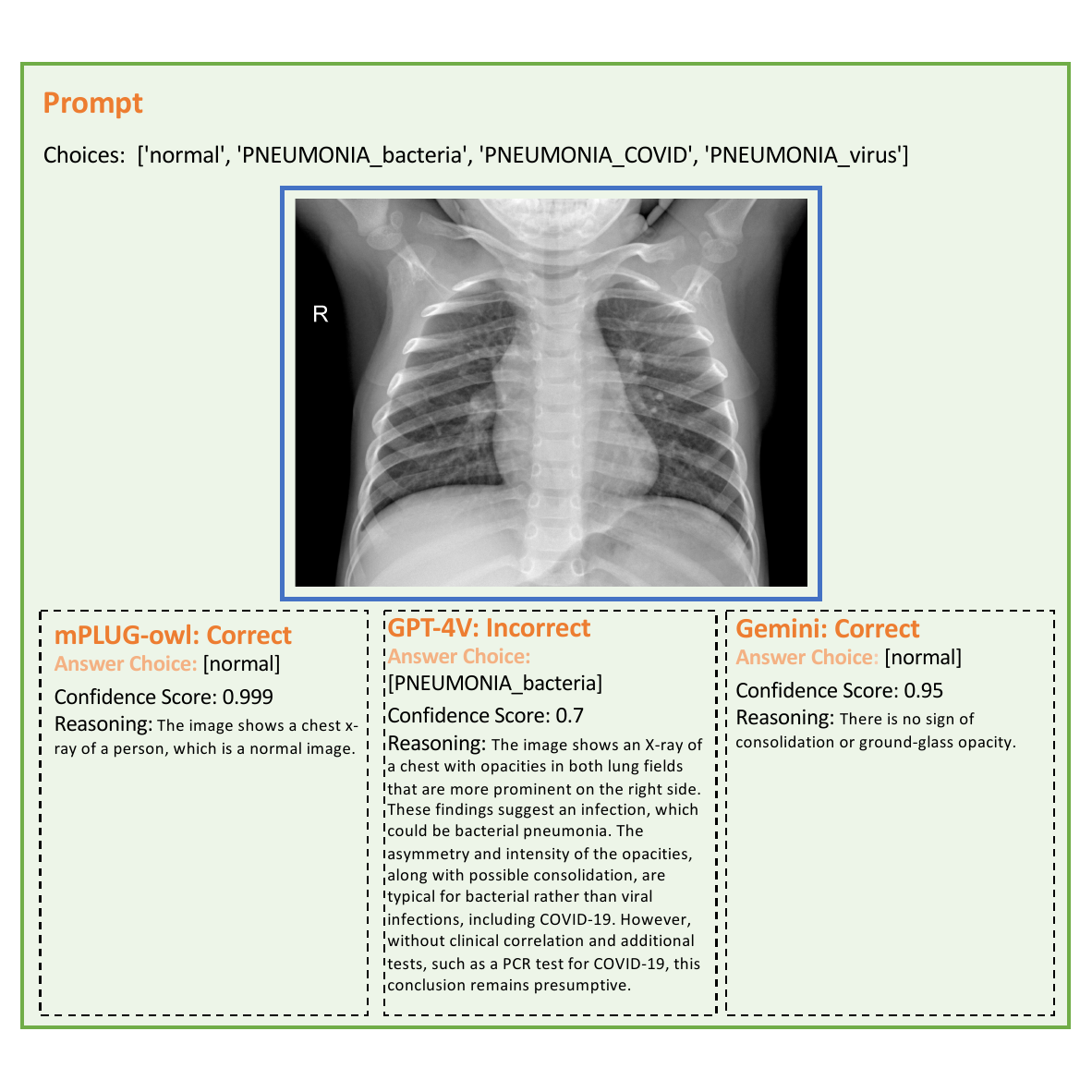}
    \caption{Domain-Specific Distribution Shift in zero-shot generalization: Case 11, analyzing the \textit{normal} category in the \textit{chest\_xray} domain of the XCOVFour dataset. In this case, both mPLUG-owl and Gemini predict incorrectly. mPLUG-owl shows a higher confidence score of 0.999, compared to 0.95 by Gemini. mPLUG-owl and Gemini merely state that the image is a normal X-ray. Although GPT-4V makes an incorrect prediction, it conducts a detailed analysis of the image, and in its rationale, it also emphasizes that `without clinical correlation and additional tests, such as a PCR test for COVID-19, this conclusion remains presumptive.' }
\end{figure*}

\begin{figure*}[ht]
    \centering
    \label{fig:}
    \includegraphics[width=1\textwidth]{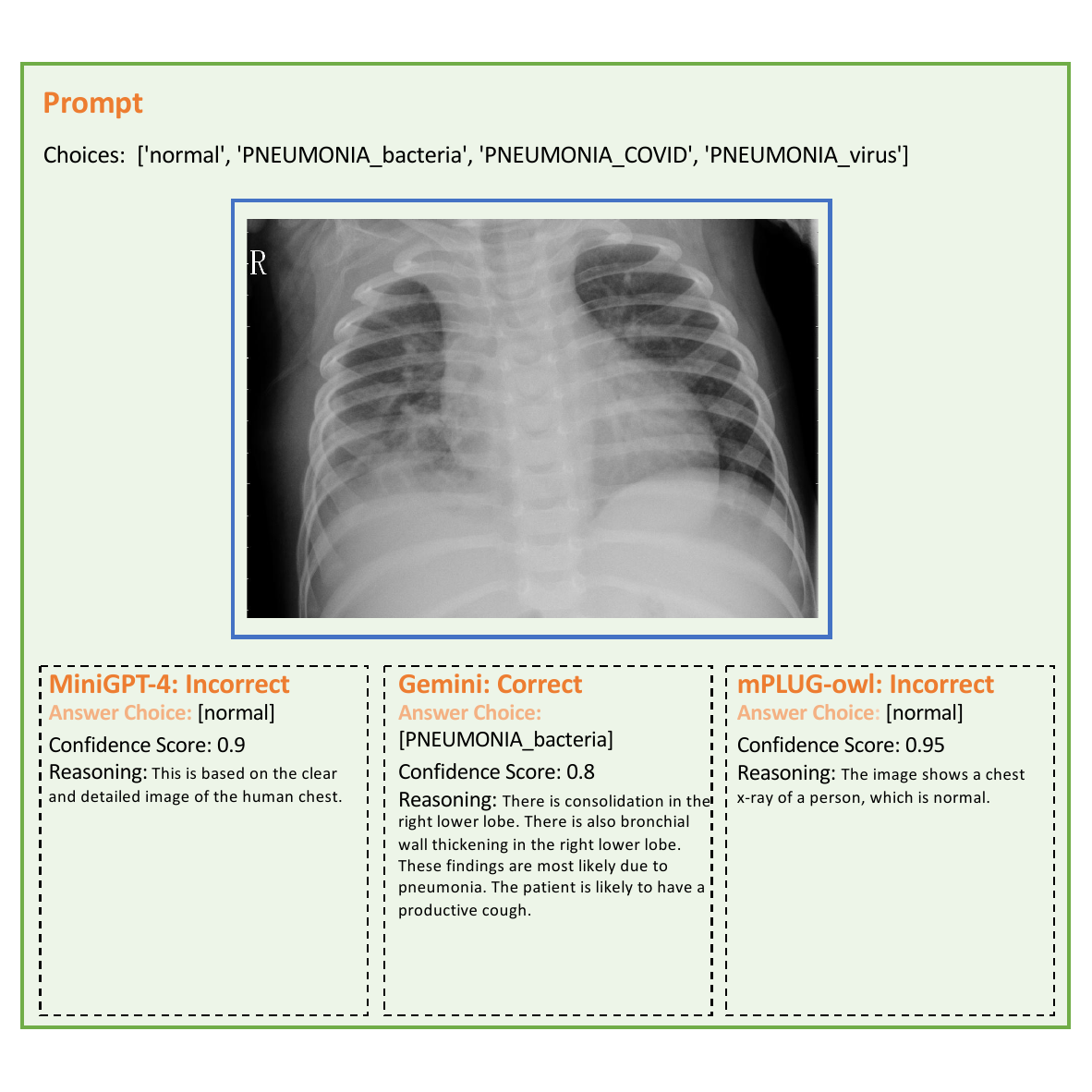}
    \caption{Domain-Specific Distribution Shift in zero-shot generalization: Case 12, analyzing the \textit{PNEUMONIA\_bacteria} category in the \textit{chest\_xray} domain of the XCOVFour dataset. In this case, both MiniGPT-4 and mPLUG-owl incorrectly categorize the provided X-ray as \textit{normal}, with insufficient rationale provided. Gemini conducts a detailed analysis of the chest X-ray, identifying an issue with the right lower lobe, thereby diagnosing the X-ray as \textit{PNEUMONIA\_bacteria}.}
\end{figure*}

%% file: paragraphs/app-showcase-icl.tex
\subsection{Case Demonstration of In-context Learning Generalization}
Here we present cases of ICL generalization of MLLMs.
These cases reveal that while overall ICL significantly enhances the generalization of MLLMs, it does not guarantee performance improvements across all scenarios. Notably, some situations correctly predicted by LLMs at zero-shot were incorrectly predicted after ICL, suggesting potential harm or instability under certain conditions. Additionally, we observed numerous instances of strange effects resulting from ICL, highlighting the need for further investigation into its mechanisms and limitations for specific tasks. These results underscore the importance of understanding the stability and potential risks associated with ICL before its widespread adoption.

\textbf{Please note that we omit the detailed prompt and only category selection space for understanding in the following presentation of case studies. All the models are assigned to provide responses to the same textual prompt as shown in Section \ref{sec:icl} and \ref{sec:app-prompt-icl}.}

\begin{figure*}[ht]
    \centering
    \label{fig:}
    \includegraphics[width=1\textwidth]{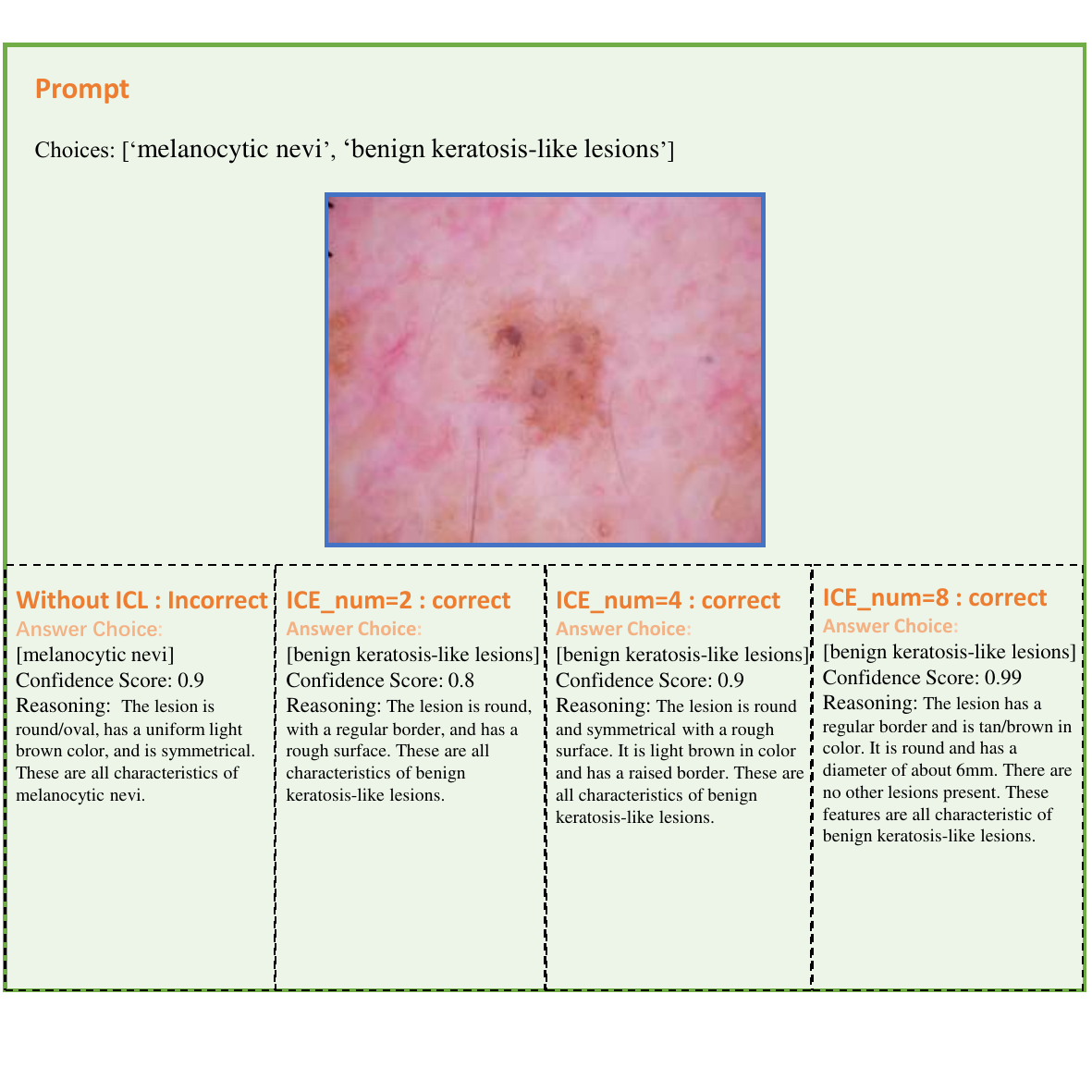}
    \caption{
    Domain-Specific Distribution Shift in in-context learning generalization: Case 1, analyzing the \textit{benign keratosis-like lesions} category in the \textit{Histopathology} domain of the HAM10000 dataset. ICE are sampled from the target distribution and the responses are from Gemini.
    In this case, Gemini predicts the example as \textit{melanocytic nevi} incorrectly without ICL. With 2, 4, and 8 ICE, Gemini returns correct predictions with increasing confidence. This shows the significant improvement significant enhancements brought about by ICL.}
\end{figure*}

\begin{figure*}[ht]
    \centering
    \label{fig:}
    \includegraphics[width=1\textwidth]{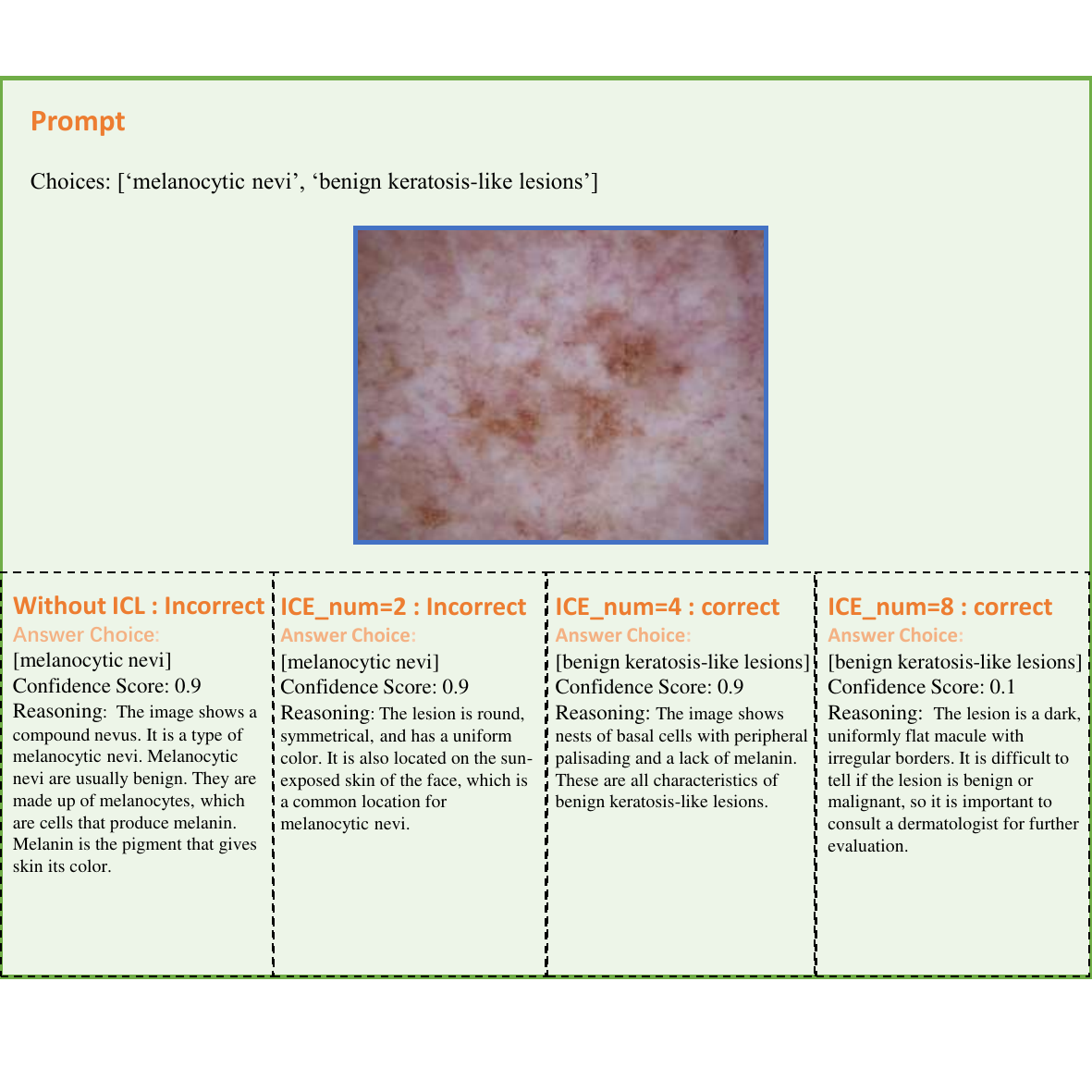}
    \caption{Domain-Specific Distribution Shift in in-context learning generalization: Case 2, analyzing the \textit{benign keratosis-like lesions} category in the \textit{Histopathology} domain of the HAM10000 dataset. ICE are sampled from the target distribution and the responses are from Gemini. 
    In this case, Gemini predicts the example is \textit{melanocytic nevi} incorrectly with high confidence when we test with 0 or 2 ICE. When the number of ICE is 4, Gemini predicts correctly with a confidence score of 0.9 and gives rational reasoning. When the number of ICE is 8, Gemini predicts correctly with a confidence score of 0.1 and the reasoning seems unreliable. Thus the effect of ICE can be unstable.}
\end{figure*}

\begin{figure*}[ht]
    \centering
    \label{fig:}
    \includegraphics[width=1\textwidth]{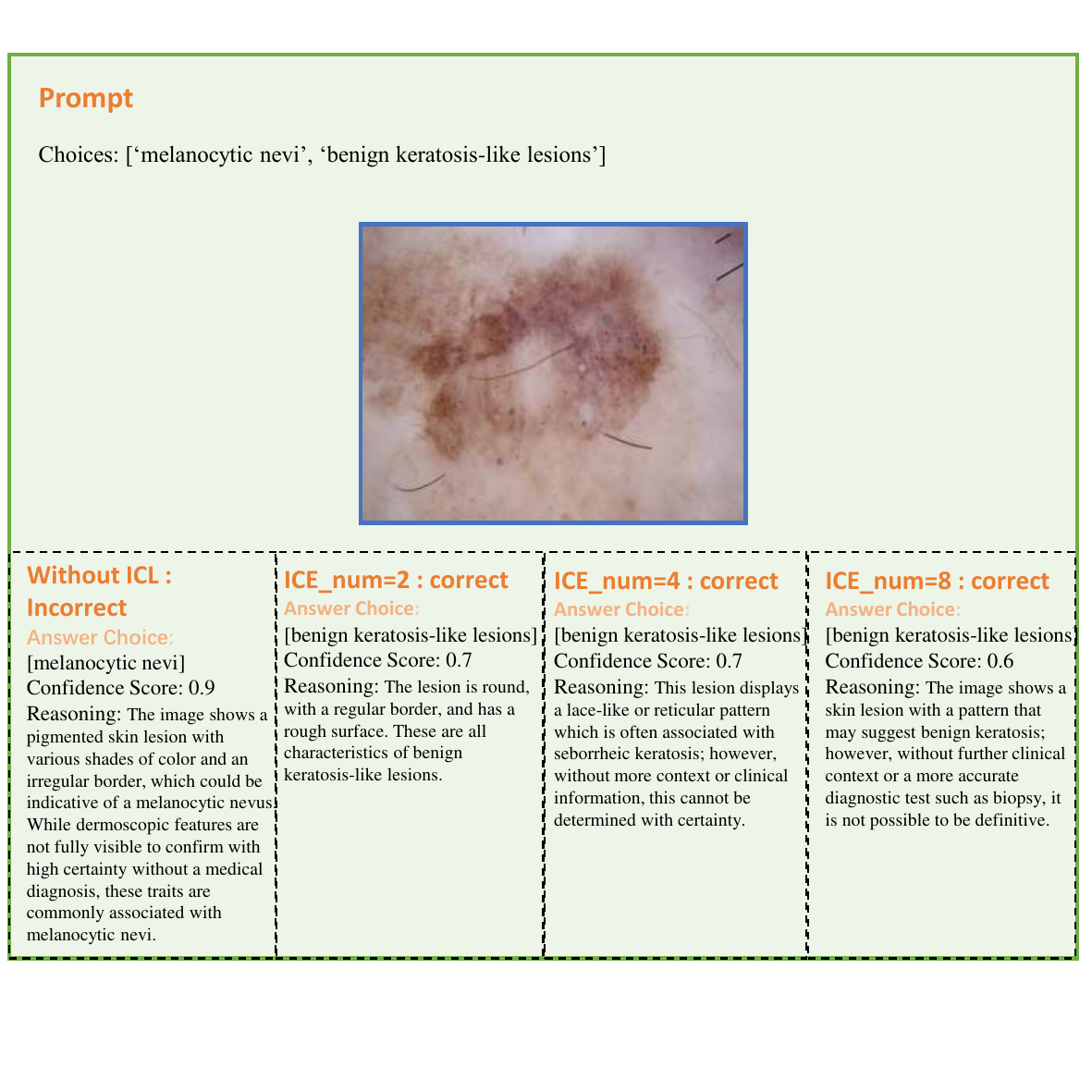}
    \caption{Domain-Specific Distribution Shift in in-context learning generalization: Case 3, analyzing the \textit{benign keratosis-like lesions} category in the \textit{Consensus} domain of the HAM10000 dataset. ICE are sampled from the target distribution and the responses are from GPT-4V. 
    In this case, GPT-4V gives the answer choice of \textit{melanocytic nevi} with a high confidence score of 0.9 without ICL. When tested with 2, 4, and 8 ICE, GPT-4V returns the correct choice. The reasoning seems unconfident and is reflected by the confidence score. We consider this to be a significant improvement to return a lower confidence score when making predictions that are not very rigorous.}
\end{figure*}

\begin{figure*}[ht]
    \centering
    \label{fig:}
    \includegraphics[width=1\textwidth]{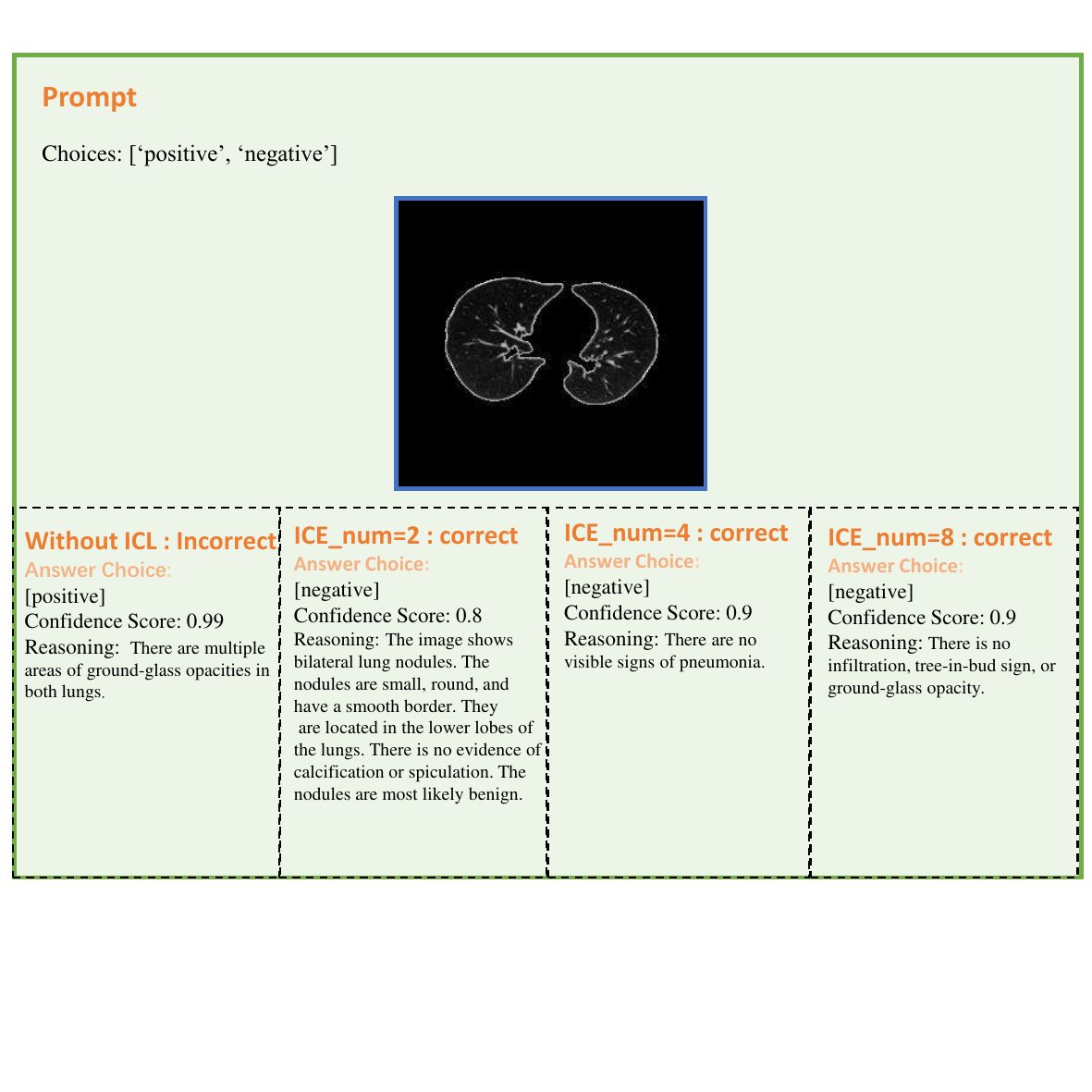}
    \caption{
    Domain-Specific Distribution Shift in in-context learning generalization: Case 4, analyzing the \textit{negative} category of the CT-XCOV dataset.
    ICE are sampled from the target distribution and the responses are from GPT-4V. 
    In this case, GPT-4V gives the answer choice of \textit{positive} with a high confidence score of 0.9 without ICL. When tested with 2, 4, and 8 ICE, GPT-4V returns the correct choice and the reasoning seems rational.}
\end{figure*}

\begin{figure*}[ht]
    \centering
    \label{fig:}
    \includegraphics[width=1\textwidth]{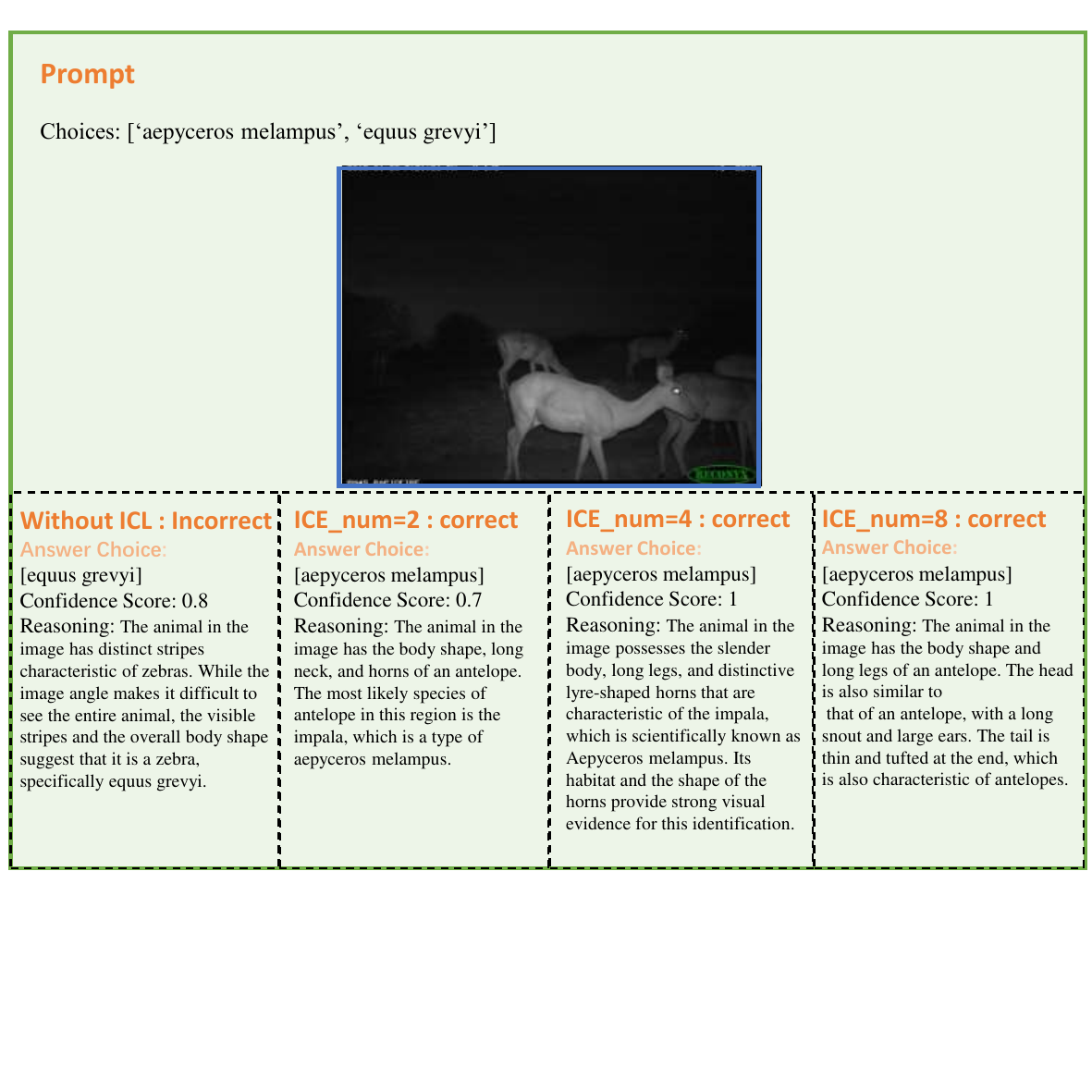}
    \caption{
    Domain-Specific Distribution Shift in in-context learning generalization: Case 5, analyzing the \textit{aepyceros melampus} category in the \textit{locatioin\_1} domain of the iWildCam dataset.
    ICE are sampled from the target distribution and the responses are from Gemini. In this case, Gemini classifies the sample as \textit{equus grevyi} incorrectly without ICE and as \textit{aepyceros melampus} correctly with 2, 4, and 8 ICE. Notably, more ICE from the target distribution leads to a higher confidence score.
    }
\end{figure*}

\begin{figure*}[ht]
    \centering
    \label{fig:}
    \includegraphics[width=1\textwidth]{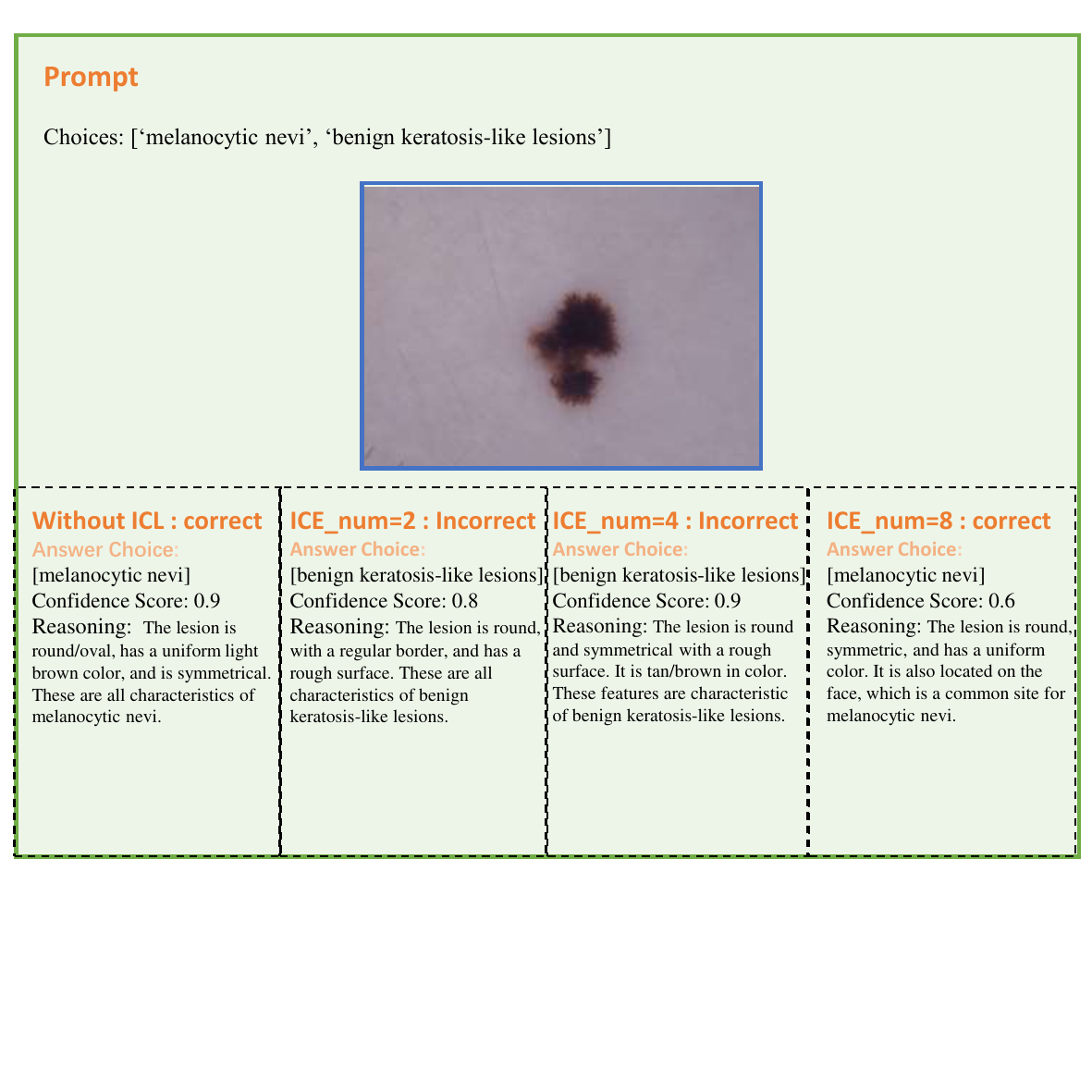}
    \caption{Domain-Specific Distribution Shift in in-context learning generalization: Case 6, analyzing the \textit{melanocytic nevi} category in the \textit{Consensus} domain of the HAM10000 dataset.
    The ICE are sampled from another domain to explore the impact of domain shifts. The responses are from GPT-4V. In this case, GPT-4V gives the wrong answer when ICE number is 2 and 4. This suggests that the domain shift between ICE and test samples could introduce performance degradation and instability. And when the ICE number is 8, GPT-4V returns a correct answer with a lower confidence score of 0.6.
    }
\end{figure*}

\begin{figure*}[ht]
    \centering
    \label{fig:}
    \includegraphics[width=1\textwidth]{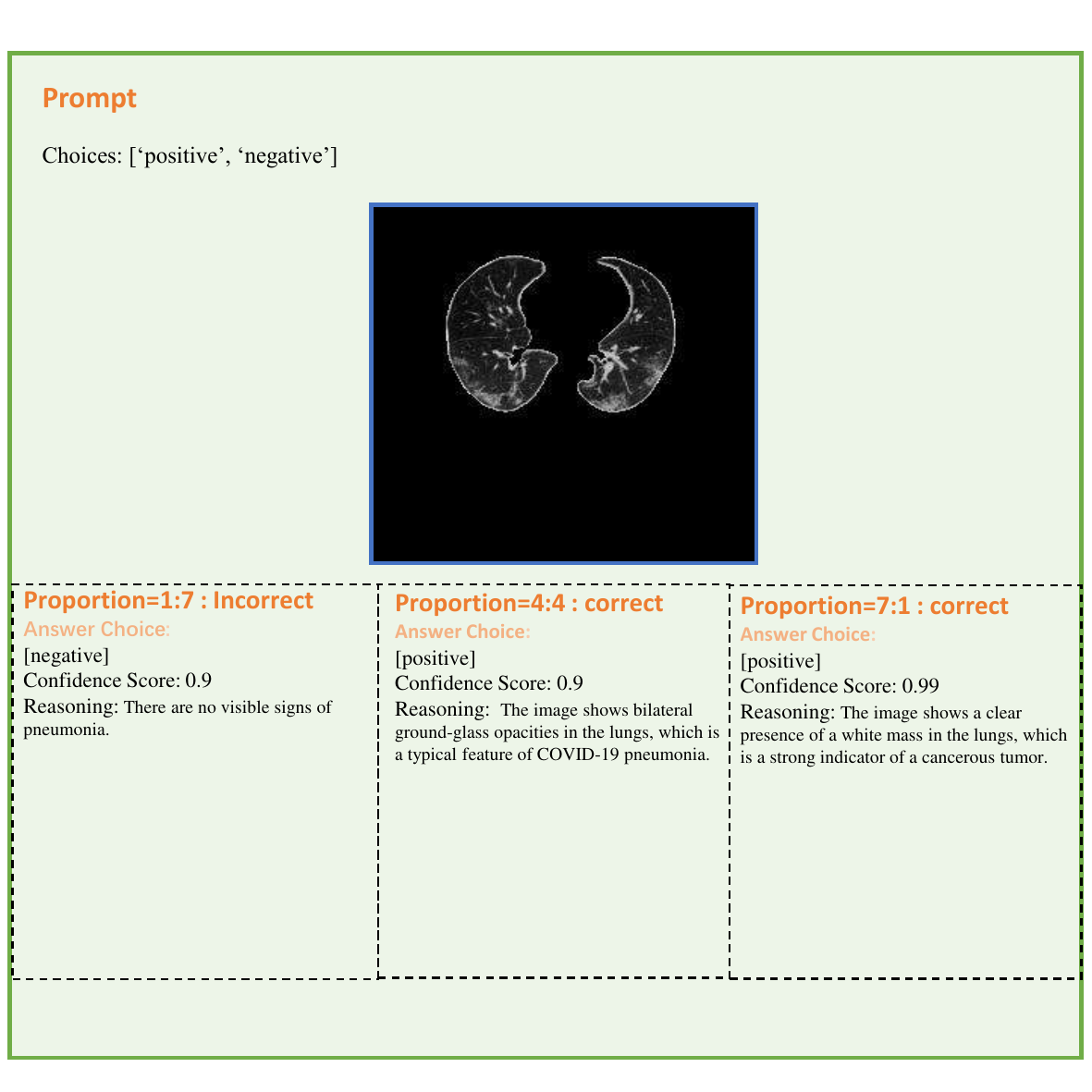}
    \caption{
    Domain-Specific Distribution Shift in in-context learning generalization: Case 7, analyzing the \textit{positive} category of the CT-XCOV dataset.
    ICE are sampled from the target domain yet the ratio of categories is not aligned with that in test samples to explore the impact of label shifts between ICE and test samples. The responses are from GPT-4V. When the ratio of \textit{positive} to \textit{negative} is 1:7, GPT-4V believes that the test sample is a negative sample with no visible signs of pneumonia. When the ratio is 4:4, GPT-4V returns a \textit{positive} and believes the sample contains a typical feature of COVID-19 pneumonia. When the ratio is 7:1, the answer choice is \textit{positive} and the confidence reaches 0.99. However, the reasoning seems that GPT-4V believes that there is a cancerous tumor in the sample. This suggests that the label shifts between ICE and test samples can introduce serious misinformation and unreliable preferences.
    }
\end{figure*}

\begin{figure*}[ht]
    \centering
    \label{fig:}
    \includegraphics[width=1\textwidth]{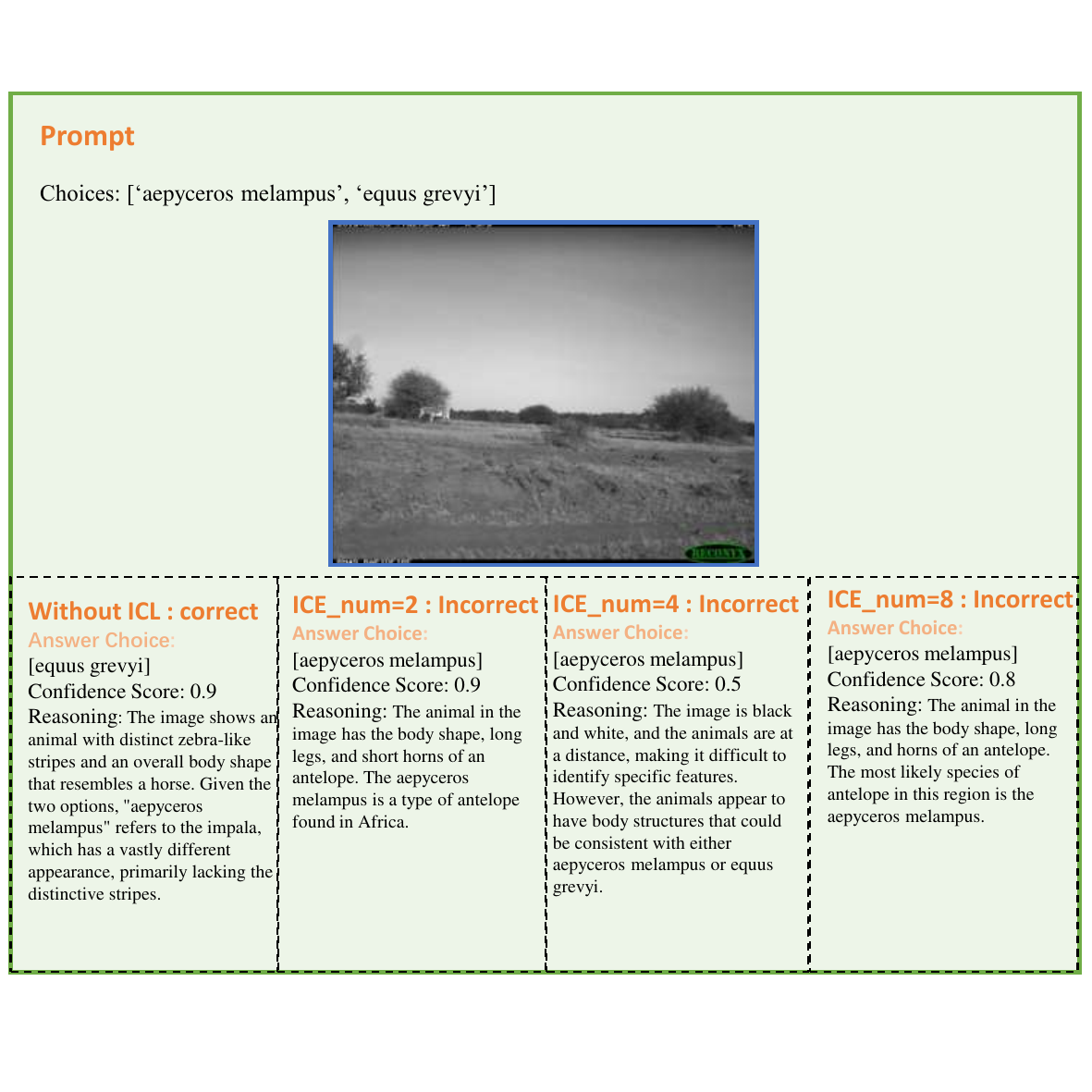}
    \caption{
    Domain-Specific Distribution Shift in in-context learning generalization: Case 8, analyzing the \textit{equus grevyi} category in the \textit{locatioin\_1} domain of the iWildCam dataset.
    There are spurious correlation shifts between ICE and test samples, i.e., the \textit{equus grevyi} in test samples are sampled from the \textit{locatioin\_1} domain and the \textit{aepyceros melampus} are sampled from the \textit{locatioin\_23} domain, while the opposite in ICE. The responses are from Gemini. In this case, Gemini returns the correct prediction and detailed and rational reasoning without ICE. However, with ICE, hallucinations in both prediction and justification are significantly exacerbated. This suggests that spurious correlation shifts between ICE and test samples can introduce severe performance degradation.}
\end{figure*}

%% file: main.bbl
\begin{thebibliography}{99}
\providecommand{\natexlab}[1]{#1}
\providecommand{\url}[1]{\texttt{#1}}
\expandafter\ifx\csname urlstyle\endcsname\relax
  \providecommand{\doi}[1]{doi: #1}\else
  \providecommand{\doi}{doi: \begingroup \urlstyle{rm}\Url}\fi

\bibitem[Ahuja and Lopez-Paz(2023)]{ahuja2023closer}
Kartik Ahuja and David Lopez-Paz.
\newblock A closer look at in-context learning under distribution shifts, 2023.

\bibitem[Alayrac et~al.(2022)Alayrac, Donahue, Luc, Miech, Barr, and et~al.]{alayrac2022flamingo}
Jean-Baptiste Alayrac, Jeff Donahue, Pauline Luc, Antoine Miech, Iain Barr, and et~al.
\newblock Flamingo: a visual language model for few-shot learning.
\newblock \emph{Advances in Neural Information Processing Systems}, 35:\penalty0 23716--23736, 2022.

\bibitem[Arjovsky et~al.(2019{\natexlab{a}})Arjovsky, Bottou, Gulrajani, and Lopez-Paz]{Arjovsky2019InvariantRM}
Mart{\'i}n Arjovsky, L{\'e}on Bottou, Ishaan Gulrajani, and David Lopez-Paz.
\newblock Invariant risk minimization.
\newblock \emph{ArXiv}, abs/1907.02893, 2019{\natexlab{a}}.
\newblock URL \url{https://api.semanticscholar.org/CorpusID:195820364}.

\bibitem[Arjovsky et~al.(2019{\natexlab{b}})Arjovsky, Bottou, Gulrajani, and Lopez-Paz]{arjovsky2019invariant}
Martin Arjovsky, L{\'e}on Bottou, Ishaan Gulrajani, and David Lopez-Paz.
\newblock Invariant risk minimization.
\newblock \emph{arXiv preprint arXiv:1907.02893}, 2019{\natexlab{b}}.

\bibitem[Arpit et~al.(2021)Arpit, Wang, Zhou, and Xiong]{Arpit2021EnsembleOA}
Devansh Arpit, Huan Wang, Yingbo Zhou, and Caiming Xiong.
\newblock Ensemble of averages: Improving model selection and boosting performance in domain generalization.
\newblock \emph{ArXiv}, abs/2110.10832, 2021.
\newblock URL \url{https://api.semanticscholar.org/CorpusID:239049452}.

\bibitem[Awadalla et~al.(2023)Awadalla, Gao, Gardner, Hessel, Hanafy, and et~al.]{awadalla2023openflamingo}
Anas Awadalla, Irena Gao, Josh Gardner, Jack Hessel, Yusuf Hanafy, and et~al.
\newblock Openflamingo: An open-source framework for training large autoregressive vision-language models, 2023.

\bibitem[Azizzadenesheli et~al.(2019)Azizzadenesheli, Liu, Yang, and Anandkumar]{azizzadenesheli2018regularized}
Kamyar Azizzadenesheli, Anqi Liu, Fanny Yang, and Animashree Anandkumar.
\newblock Regularized learning for domain adaptation under label shifts.
\newblock In \emph{International Conference on Learning Representations}, 2019.

\bibitem[Bai et~al.(2023{\natexlab{a}})Bai, Bai, Yang, Wang, Tan, and et~al.]{bai2023qwen}
Jinze Bai, Shuai Bai, Shusheng Yang, Shijie Wang, Sinan Tan, and et~al.
\newblock Qwen-vl: A frontier large vision-language model with versatile abilities.
\newblock \emph{arXiv preprint arXiv:2308.12966}, 2023{\natexlab{a}}.

\bibitem[Bai et~al.(2023{\natexlab{b}})Bai, Yang, Bai, Wang, Zhang, Lin, Wang, Zhou, and Zhou]{bai2023touchstone}
Shuai Bai, Shusheng Yang, Jinze Bai, Peng Wang, Xingxuan Zhang, Junyang Lin, Xinggang Wang, Chang Zhou, and Jingren Zhou.
\newblock Touchstone: Evaluating vision-language models by language models.
\newblock \emph{arXiv preprint arXiv:2308.16890}, 2023{\natexlab{b}}.

\bibitem[Bar et~al.(2022)Bar, Gandelsman, Darrell, Globerson, and Efros]{bar2022visual}
Amir Bar, Yossi Gandelsman, Trevor Darrell, Amir Globerson, and Alexei Efros.
\newblock Visual prompting via image inpainting.
\newblock \emph{Advances in Neural Information Processing Systems}, 35:\penalty0 25005--25017, 2022.

\bibitem[Beery et~al.(2018)Beery, Van~Horn, and Perona]{beery2018recognition}
Sara Beery, Grant Van~Horn, and Pietro Perona.
\newblock Recognition in terra incognita.
\newblock In \emph{Proceedings of the European conference on computer vision (ECCV)}, pages 456--473, 2018.

\bibitem[Beery et~al.(2021)Beery, Agarwal, Cole, and Birodkar]{beery2021iwildcam}
Sara Beery, Arushi Agarwal, Elijah Cole, and Vighnesh Birodkar.
\newblock The iwildcam 2021 competition dataset.
\newblock \emph{arXiv preprint arXiv:2105.03494}, 2021.

\bibitem[Brown et~al.(2020)Brown, Mann, Ryder, Subbiah, Kaplan, and et~al.]{brown2020language}
Tom~B. Brown, Benjamin Mann, Nick Ryder, Melanie Subbiah, Jared Kaplan, and et~al.
\newblock Language models are few-shot learners, 2020.

\bibitem[Carlucci et~al.(2019)Carlucci, D'Innocente, Bucci, Caputo, and Tommasi]{Carlucci2019DomainGB}
Fabio~Maria Carlucci, Antonio D'Innocente, Silvia Bucci, Barbara Caputo, and Tatiana Tommasi.
\newblock Domain generalization by solving jigsaw puzzles.
\newblock \emph{2019 IEEE/CVF Conference on Computer Vision and Pattern Recognition (CVPR)}, pages 2224--2233, 2019.
\newblock URL \url{https://api.semanticscholar.org/CorpusID:81978372}.

\bibitem[Cha et~al.(2021)Cha, Chun, Lee, Cho, Park, and et~al.]{Cha2021SWADDG}
Junbum Cha, Sanghyuk Chun, Kyungjae Lee, Han-Cheol Cho, Seunghyun Park, and et~al.
\newblock Swad: Domain generalization by seeking flat minima.
\newblock In \emph{Neural Information Processing Systems}, 2021.
\newblock URL \url{https://api.semanticscholar.org/CorpusID:235367622}.

\bibitem[Chan et~al.(2022)Chan, Dasgupta, Kim, Kumaran, Lampinen, and Hill]{chan2022transformers}
Stephanie C.~Y. Chan, Ishita Dasgupta, Junkyung Kim, Dharshan Kumaran, Andrew~K. Lampinen, and Felix Hill.
\newblock Transformers generalize differently from information stored in context vs in weights, 2022.

\bibitem[Cherti et~al.(2023)Cherti, Beaumont, Wightman, Wortsman, Ilharco, Gordon, Schuhmann, Schmidt, and Jitsev]{cherti2023reproducible}
Mehdi Cherti, Romain Beaumont, Ross Wightman, Mitchell Wortsman, Gabriel Ilharco, Cade Gordon, Christoph Schuhmann, Ludwig Schmidt, and Jenia Jitsev.
\newblock Reproducible scaling laws for contrastive language-image learning.
\newblock In \emph{Proceedings of the IEEE/CVF Conference on Computer Vision and Pattern Recognition}, pages 2818--2829, 2023.

\bibitem[Cho et~al.(2023)Cho, Nam, Kim, Yang, and Kwak]{cho2023promptstyler}
Junhyeong Cho, Gilhyun Nam, Sungyeon Kim, Hunmin Yang, and Suha Kwak.
\newblock Promptstyler: Prompt-driven style generation for source-free domain generalization.
\newblock In \emph{Proceedings of the IEEE/CVF International Conference on Computer Vision}, pages 15702--15712, 2023.

\bibitem[Chowdhery et~al.(2022)Chowdhery, Narang, Devlin, Bosma, Mishra, and et~al.]{chowdhery2022palm}
Aakanksha Chowdhery, Sharan Narang, Jacob Devlin, Maarten Bosma, Gaurav Mishra, and et~al.
\newblock Palm: Scaling language modeling with pathways, 2022.

\bibitem[Christie et~al.(2018)Christie, Fendley, Wilson, and Mukherjee]{christie2018functional}
Gordon Christie, Neil Fendley, James Wilson, and Ryan Mukherjee.
\newblock Functional map of the world.
\newblock In \emph{Proceedings of the IEEE Conference on Computer Vision and Pattern Recognition}, pages 6172--6180, 2018.

\bibitem[Chu et~al.(2022)Chu, Jin, Zhu, Wang, Wang, Zhang, and Mei]{Chu2022DNADG}
Xu~Chu, Yujie Jin, Wenwu Zhu, Yasha Wang, Xin Wang, Shanghang Zhang, and Hong Mei.
\newblock Dna: Domain generalization with diversified neural averaging.
\newblock In \emph{International Conference on Machine Learning}, 2022.
\newblock URL \url{https://api.semanticscholar.org/CorpusID:250340736}.

\bibitem[Cruz et~al.(2020)Cruz, Wasenmuller, Beise, Stifter, and Stricker]{cruz2020sviro}
Steve Dias~Da Cruz, Oliver Wasenmuller, Hans-Peter Beise, Thomas Stifter, and Didier Stricker.
\newblock Sviro: Synthetic vehicle interior rear seat occupancy dataset and benchmark.
\newblock In \emph{Proceedings of the IEEE/CVF Winter Conference on Applications of Computer Vision}, pages 973--982, 2020.

\bibitem[Dai et~al.(2023)Dai, Li, Li, Tiong, Zhao, and et~al.]{dai2023instructblip}
W~Dai, J~Li, D~Li, AMH Tiong, J~Zhao, and et~al.
\newblock Instructblip: Towards general-purpose vision-language models with instruction tuning. arxiv 2023.
\newblock \emph{arXiv preprint arXiv:2305.06500}, 2, 2023.

\bibitem[Deng(2012)]{deng2012mnist}
Li~Deng.
\newblock The mnist database of handwritten digit images for machine learning research [best of the web].
\newblock \emph{IEEE signal processing magazine}, 29\penalty0 (6):\penalty0 141--142, 2012.

\bibitem[Dosovitskiy et~al.(2021)Dosovitskiy, Beyer, Kolesnikov, Weissenborn, Zhai, and et~al.]{dosovitskiy2021an}
Alexey Dosovitskiy, Lucas Beyer, Alexander Kolesnikov, Dirk Weissenborn, Xiaohua Zhai, and et~al.
\newblock An image is worth 16x16 words: Transformers for image recognition at scale.
\newblock In \emph{International Conference on Learning Representations}, 2021.
\newblock URL \url{https://openreview.net/forum?id=YicbFdNTTy}.

\bibitem[Dou et~al.(2019)Dou, de~Castro, Kamnitsas, and Glocker]{Dou2019DomainGV}
Qi~Dou, Daniel~Coelho de~Castro, Konstantinos Kamnitsas, and Ben Glocker.
\newblock Domain generalization via model-agnostic learning of semantic features.
\newblock In \emph{Neural Information Processing Systems}, 2019.
\newblock URL \url{https://api.semanticscholar.org/CorpusID:202768984}.

\bibitem[Driess et~al.(2023)Driess, Xia, Sajjadi, Lynch, Chowdhery, and et~al.]{driess2023palme}
Danny Driess, Fei Xia, Mehdi S.~M. Sajjadi, Corey Lynch, Aakanksha Chowdhery, and et~al.
\newblock Palm-e: An embodied multimodal language model, 2023.

\bibitem[Elbouknify et~al.(2023)Elbouknify, Bouhoute, Fardousse, Berrada, and Badri]{elbouknify2023ct}
Ismail Elbouknify, Afaf Bouhoute, Khalid Fardousse, Ismail Berrada, and Abdelmajid Badri.
\newblock Ct-xcov: a ct-scan based explainable framework for covid-19 diagnosis.
\newblock In \emph{2023 10th International Conference on Wireless Networks and Mobile Communications (WINCOM)}, pages 1--8. IEEE, 2023.

\bibitem[Fang et~al.(2013)Fang, Xu, and Rockmore]{fang2013unbiased}
Chen Fang, Ye~Xu, and Daniel~N Rockmore.
\newblock Unbiased metric learning: On the utilization of multiple datasets and web images for softening bias.
\newblock In \emph{Proceedings of the IEEE International Conference on Computer Vision}, pages 1657--1664, 2013.

\bibitem[Foret et~al.(2020)Foret, Kleiner, Mobahi, and Neyshabur]{foret2020sharpness}
Pierre Foret, Ariel Kleiner, Hossein Mobahi, and Behnam Neyshabur.
\newblock Sharpness-aware minimization for efficiently improving generalization.
\newblock \emph{arXiv preprint arXiv:2010.01412}, 2020.

\bibitem[Fu et~al.(2023)Fu, Chen, Shen, Qin, Zhang, Lin, Yang, Zheng, Li, Sun, et~al.]{fu2023mme}
Chaoyou Fu, Peixian Chen, Yunhang Shen, Yulei Qin, Mengdan Zhang, Xu~Lin, Jinrui Yang, Xiawu Zheng, Ke~Li, Xing Sun, et~al.
\newblock Mme: A comprehensive evaluation benchmark for multimodal large language models.
\newblock \emph{arXiv preprint arXiv:2306.13394}, 2023.

\bibitem[Gao et~al.(2023)Gao, Han, Zhang, Lin, Geng, Zhou, Zhang, Lu, He, Yue, et~al.]{gao2023llama}
Peng Gao, Jiaming Han, Renrui Zhang, Ziyi Lin, Shijie Geng, Aojun Zhou, Wei Zhang, Pan Lu, Conghui He, Xiangyu Yue, et~al.
\newblock Llama-adapter v2: Parameter-efficient visual instruction model.
\newblock \emph{arXiv preprint arXiv:2304.15010}, 2023.

\bibitem[Ghifary et~al.(2015)Ghifary, Kleijn, Zhang, and Balduzzi]{ghifary2015domain}
Muhammad Ghifary, W~Bastiaan Kleijn, Mengjie Zhang, and David Balduzzi.
\newblock Domain generalization for object recognition with multi-task autoencoders.
\newblock In \emph{Proceedings of the IEEE international conference on computer vision}, pages 2551--2559, 2015.

\bibitem[Gulrajani and Lopez-Paz(2020)]{gulrajani2020search}
Ishaan Gulrajani and David Lopez-Paz.
\newblock In search of lost domain generalization.
\newblock \emph{arXiv preprint arXiv:2007.01434}, 2020.

\bibitem[Han et~al.(2021)Han, He, Li, Wei, Wang, and Yin]{han2021semi}
Zhongyi Han, Rundong He, Tianyang Li, Benzheng Wei, Jian Wang, and Yilong Yin.
\newblock Semi-supervised screening of covid-19 from positive and unlabeled data with constraint non-negative risk estimator.
\newblock In \emph{Information Processing in Medical Imaging: 27th International Conference, IPMI 2021, Virtual Event, June 28--June 30, 2021, Proceedings 27}, pages 611--623. Springer, 2021.

\bibitem[Han et~al.(2023)Han, Zhou, He, Wang, Xie, Wu, Yin, Khan, Yao, Liu, et~al.]{han2023well}
Zhongyi Han, Guanglin Zhou, Rundong He, Jindong Wang, Xing Xie, Tailin Wu, Yilong Yin, Salman Khan, Lina Yao, Tongliang Liu, et~al.
\newblock How well does gpt-4v (ision) adapt to distribution shifts? a preliminary investigation.
\newblock \emph{arXiv preprint arXiv:2312.07424}, 2023.

\bibitem[Hu et~al.(2019)Hu, Zhang, Chen, and Chan]{hu2019domain}
Shoubo Hu, Kun Zhang, Zhitang Chen, and Laiwan Chan.
\newblock Domain generalization via multidomain discriminant analysis, 2019.

\bibitem[Ji et~al.(2022)Ji, Zhang, Wu, Wu, Huang, and et~al.]{ji2022drugood}
Yuanfeng Ji, Lu~Zhang, Jiaxiang Wu, Bingzhe Wu, Long-Kai Huang, and et~al.
\newblock Drugood: Out-of-distribution (ood) dataset curator and benchmark for ai-aided drug discovery--a focus on affinity prediction problems with noise annotations.
\newblock \emph{arXiv preprint arXiv:2201.09637}, 2022.

\bibitem[Kirsch et~al.(2024)Kirsch, Harrison, Sohl-Dickstein, and Metz]{kirsch2024generalpurpose}
Louis Kirsch, James Harrison, Jascha Sohl-Dickstein, and Luke Metz.
\newblock General-purpose in-context learning by meta-learning transformers, 2024.

\bibitem[Koh et~al.(2021)Koh, Sagawa, Marklund, Xie, Zhang, and et~al.]{koh2021wilds}
Pang~Wei Koh, Shiori Sagawa, Henrik Marklund, Sang~Michael Xie, Marvin Zhang, and et~al.
\newblock Wilds: A benchmark of in-the-wild distribution shifts, 2021.

\bibitem[Lewis et~al.(2020)Lewis, Perez, Piktus, Petroni, Karpukhin, and et~al.]{lewis2020retrieval}
Patrick Lewis, Ethan Perez, Aleksandra Piktus, Fabio Petroni, Vladimir Karpukhin, and et~al.
\newblock Retrieval-augmented generation for knowledge-intensive nlp tasks.
\newblock \emph{Advances in Neural Information Processing Systems}, 33:\penalty0 9459--9474, 2020.

\bibitem[Li et~al.(2023{\natexlab{a}})Li, Zhang, Chen, Wang, Yang, and Liu]{li2023otter}
Bo~Li, Yuanhan Zhang, Liangyu Chen, Jinghao Wang, Jingkang Yang, and Ziwei Liu.
\newblock Otter: A multi-modal model with in-context instruction tuning, 2023{\natexlab{a}}.

\bibitem[Li et~al.(2017)Li, Yang, Song, and Hospedales]{li2017deeper}
Da~Li, Yongxin Yang, Yi-Zhe Song, and Timothy~M Hospedales.
\newblock Deeper, broader and artier domain generalization.
\newblock In \emph{Proceedings of the IEEE international conference on computer vision}, pages 5542--5550, 2017.

\bibitem[Li et~al.(2019)Li, Zhang, Yang, Liu, Song, and Hospedales]{Li2019EpisodicTF}
Da~Li, Jianshu Zhang, Yongxin Yang, Cong Liu, Yi-Zhe Song, and Timothy~M. Hospedales.
\newblock Episodic training for domain generalization.
\newblock \emph{2019 IEEE/CVF International Conference on Computer Vision (ICCV)}, pages 1446--1455, 2019.
\newblock URL \url{https://api.semanticscholar.org/CorpusID:59553457}.

\bibitem[Li et~al.(2023{\natexlab{b}})Li, Li, Savarese, and Hoi]{li2023blip}
Junnan Li, Dongxu Li, Silvio Savarese, and Steven Hoi.
\newblock Blip-2: Bootstrapping language-image pre-training with frozen image encoders and large language models.
\newblock \emph{arXiv preprint arXiv:2301.12597}, 2023{\natexlab{b}}.

\bibitem[Li et~al.(2023{\natexlab{c}})Li, Ren, Jiang, Shen, Zhang, and Li]{Li2023SIMPLESM}
Ziyue Li, Kan Ren, Xinyang Jiang, Yifei Shen, Haipeng Zhang, and Dongsheng Li.
\newblock Simple: Specialized model-sample matching for domain generalization.
\newblock In \emph{International Conference on Learning Representations}, 2023{\natexlab{c}}.
\newblock URL \url{https://api.semanticscholar.org/CorpusID:259298845}.

\bibitem[Lin et~al.(2022)Lin, Zhu, Tan, and Cui]{lin2022zin}
Yong Lin, Shengyu Zhu, Lu~Tan, and Peng Cui.
\newblock Zin: When and how to learn invariance without environment partition?
\newblock \emph{Advances in Neural Information Processing Systems}, 35:\penalty0 24529--24542, 2022.

\bibitem[Liu et~al.(2023{\natexlab{a}})Liu, Li, Li, and Lee]{liu2023improved}
Haotian Liu, Chunyuan Li, Yuheng Li, and Yong~Jae Lee.
\newblock Improved baselines with visual instruction tuning, 2023{\natexlab{a}}.

\bibitem[Liu et~al.(2023{\natexlab{b}})Liu, Li, Wu, and Lee]{liu2023visual}
Haotian Liu, Chunyuan Li, Qingyang Wu, and Yong~Jae Lee.
\newblock Visual instruction tuning.
\newblock \emph{arXiv preprint arXiv:2304.08485}, 2023{\natexlab{b}}.

\bibitem[Lu et~al.(2023)Lu, Bansal, Xia, Liu, Li, Hajishirzi, Cheng, Chang, Galley, and Gao]{lu2023mathvista}
Pan Lu, Hritik Bansal, Tony Xia, Jiacheng Liu, Chunyuan Li, Hannaneh Hajishirzi, Hao Cheng, Kai-Wei Chang, Michel Galley, and Jianfeng Gao.
\newblock Mathvista: Evaluating math reasoning in visual contexts with gpt-4v, bard, and other large multimodal models.
\newblock \emph{arXiv e-prints}, pages arXiv--2310, 2023.

\bibitem[Mancini et~al.(2018{\natexlab{a}})Mancini, Bul{\`o}, Caputo, and Ricci]{Mancini2018RobustPC}
Massimiliano Mancini, Samuel~Rota Bul{\`o}, Barbara Caputo, and Elisa Ricci.
\newblock Robust place categorization with deep domain generalization.
\newblock \emph{IEEE Robotics and Automation Letters}, 3:\penalty0 2093--2100, 2018{\natexlab{a}}.
\newblock URL \url{https://api.semanticscholar.org/CorpusID:4245797}.

\bibitem[Mancini et~al.(2018{\natexlab{b}})Mancini, Bulò, Caputo, and Ricci]{Mancini18}
Massimiliano Mancini, Samuel~Rota Bulò, Barbara Caputo, and Elisa Ricci.
\newblock Best sources forward: Domain generalization through source-specific nets.
\newblock In \emph{2018 25th IEEE International Conference on Image Processing (ICIP)}, pages 1353--1357, 2018{\natexlab{b}}.
\newblock \doi{10.1109/ICIP.2018.8451318}.

\bibitem[Mendez et~al.(2019)Mendez, Gaulton, Bento, Chambers, De~Veij, F{\'e}lix, Magari{\~n}os, Mosquera, Mutowo, Nowotka, et~al.]{mendez2019chembl}
David Mendez, Anna Gaulton, A~Patr{\'\i}cia Bento, Jon Chambers, Marleen De~Veij, Eloy F{\'e}lix, Mar{\'\i}a~Paula Magari{\~n}os, Juan~F Mosquera, Prudence Mutowo, Micha{\l} Nowotka, et~al.
\newblock Chembl: towards direct deposition of bioassay data.
\newblock \emph{Nucleic acids research}, 47\penalty0 (D1):\penalty0 D930--D940, 2019.

\bibitem[of~Health et~al.(2017)]{national2017nih}
National~Institutes of~Health et~al.
\newblock Nih clinical center provides one of the largest publicly available chest x-ray datasets to scientific community, 2017.

\bibitem[OpenAI(2023)]{openai2023gpt4}
OpenAI.
\newblock Gpt-4 technical report, 2023.

\bibitem[Peng et~al.(2023{\natexlab{a}})Peng, Li, He, Galley, and Gao]{peng2023instruction}
Baolin Peng, Chunyuan Li, Pengcheng He, Michel Galley, and Jianfeng Gao.
\newblock Instruction tuning with gpt-4, 2023{\natexlab{a}}.

\bibitem[Peng et~al.(2019)Peng, Bai, Xia, Huang, Saenko, and Wang]{peng2019moment}
Xingchao Peng, Qinxun Bai, Xide Xia, Zijun Huang, Kate Saenko, and Bo~Wang.
\newblock Moment matching for multi-source domain adaptation.
\newblock In \emph{Proceedings of the IEEE/CVF international conference on computer vision}, pages 1406--1415, 2019.

\bibitem[Peng et~al.(2023{\natexlab{b}})Peng, Wang, Dong, Hao, Huang, Ma, and Wei]{peng2023kosmos}
Zhiliang Peng, Wenhui Wang, Li~Dong, Yaru Hao, Shaohan Huang, Shuming Ma, and Furu Wei.
\newblock Kosmos-2: Grounding multimodal large language models to the world.
\newblock \emph{arXiv preprint arXiv:2306.14824}, 2023{\natexlab{b}}.

\bibitem[Piratla et~al.(2020)Piratla, Netrapalli, and Sarawagi]{Piratla2020EfficientDG}
Vihari Piratla, Praneeth Netrapalli, and Sunita Sarawagi.
\newblock Efficient domain generalization via common-specific low-rank decomposition.
\newblock In \emph{International Conference on Machine Learning}, 2020.
\newblock URL \url{https://api.semanticscholar.org/CorpusID:214714229}.

\bibitem[Qiao et~al.(2020)Qiao, Zhao, and Peng]{Qiao2020LearningTL}
Fengchun Qiao, Long Zhao, and Xi~Peng.
\newblock Learning to learn single domain generalization.
\newblock \emph{2020 IEEE/CVF Conference on Computer Vision and Pattern Recognition (CVPR)}, pages 12553--12562, 2020.
\newblock URL \url{https://api.semanticscholar.org/CorpusID:214713888}.

\bibitem[Radford et~al.(2021)Radford, Kim, Hallacy, Ramesh, Goh, Agarwal, Sastry, Askell, Mishkin, Clark, et~al.]{radford2021learning}
Alec Radford, Jong~Wook Kim, Chris Hallacy, Aditya Ramesh, Gabriel Goh, Sandhini Agarwal, Girish Sastry, Amanda Askell, Pamela Mishkin, Jack Clark, et~al.
\newblock Learning transferable visual models from natural language supervision.
\newblock In \emph{International conference on machine learning}, pages 8748--8763. PMLR, 2021.

\bibitem[Ram{\'e} et~al.(2022{\natexlab{a}})Ram{\'e}, Ahuja, Zhang, Cord, Bottou, and Lopez-Paz]{Ram2022RecyclingDM}
Alexandre Ram{\'e}, Kartik Ahuja, Jianyu Zhang, Matthieu Cord, L{\'e}on Bottou, and David Lopez-Paz.
\newblock Recycling diverse models for out-of-distribution generalization.
\newblock \emph{ArXiv}, abs/2212.10445, 2022{\natexlab{a}}.
\newblock URL \url{https://api.semanticscholar.org/CorpusID:263876108}.

\bibitem[Ram{\'e} et~al.(2022{\natexlab{b}})Ram{\'e}, Kirchmeyer, Rahier, Rakotomamonjy, Gallinari, and Cord]{Ram2022DiverseWA}
Alexandre Ram{\'e}, Matthieu Kirchmeyer, Thibaud Rahier, Alain Rakotomamonjy, Patrick Gallinari, and Matthieu Cord.
\newblock Diverse weight averaging for out-of-distribution generalization.
\newblock \emph{ArXiv}, abs/2205.09739, 2022{\natexlab{b}}.
\newblock URL \url{https://api.semanticscholar.org/CorpusID:248887264}.

\bibitem[Ram{\'e} et~al.(2023)Ram{\'e}, Ahuja, Zhang, Cord, Bottou, and Lopez-Paz]{rame2023model}
Alexandre Ram{\'e}, Kartik Ahuja, Jianyu Zhang, Matthieu Cord, L{\'e}on Bottou, and David Lopez-Paz.
\newblock Model ratatouille: Recycling diverse models for out-of-distribution generalization.
\newblock In \emph{International Conference on Machine Learning}, pages 28656--28679. PMLR, 2023.

\bibitem[Seo et~al.(2019)Seo, Suh, Kim, Han, and Han]{Seo2019LearningTO}
Seonguk Seo, Yumin Suh, Dongwan Kim, Jongwoo Han, and Bohyung Han.
\newblock Learning to optimize domain specific normalization for domain generalization.
\newblock \emph{ArXiv}, abs/1907.04275, 2019.
\newblock URL \url{https://api.semanticscholar.org/CorpusID:195848051}.

\bibitem[Sun et~al.(2023)Sun, Cui, Zhang, Zhang, Yu, Luo, Wang, Rao, Liu, Huang, et~al.]{sun2023generative}
Quan Sun, Yufeng Cui, Xiaosong Zhang, Fan Zhang, Qiying Yu, Zhengxiong Luo, Yueze Wang, Yongming Rao, Jingjing Liu, Tiejun Huang, et~al.
\newblock Generative multimodal models are in-context learners.
\newblock \emph{arXiv preprint arXiv:2312.13286}, 2023.

\bibitem[Sun et~al.(2022)Sun, Meng, Liu, et~al.]{sun2022camelyon}
Tianbo Sun, Tong Meng, Yutong Liu, et~al.
\newblock Camelyon 17 challenge: A comparison of traditional machine learning (svm) with the deep learning method.
\newblock \emph{Wireless Communications and Mobile Computing}, 2022, 2022.

\bibitem[Team et~al.(2023)Team, Anil, Borgeaud, Wu, Alayrac, and et~al.]{team2023gemini}
Gemini Team, Rohan Anil, Sebastian Borgeaud, Yonghui Wu, Jean-Baptiste Alayrac, and et~al.
\newblock Gemini: a family of highly capable multimodal models.
\newblock \emph{arXiv preprint arXiv:2312.11805}, 2023.

\bibitem[Touvron et~al.(2023)Touvron, Lavril, Izacard, Martinet, Lachaux, and et~al.]{touvron2023llama}
Hugo Touvron, Thibaut Lavril, Gautier Izacard, Xavier Martinet, Marie-Anne Lachaux, and et~al.
\newblock Llama: Open and efficient foundation language models, 2023.

\bibitem[Tschandl et~al.(2018)Tschandl, Rosendahl, and Kittler]{tschandl2018ham10000}
Philipp Tschandl, Cliff Rosendahl, and Harald Kittler.
\newblock The ham10000 dataset, a large collection of multi-source dermatoscopic images of common pigmented skin lesions.
\newblock \emph{Scientific data}, 5\penalty0 (1):\penalty0 1--9, 2018.

\bibitem[Tsimpoukelli et~al.(2021)Tsimpoukelli, Menick, Cabi, Eslami, Vinyals, and Hill]{tsimpoukelli2021multimodal}
Maria Tsimpoukelli, Jacob Menick, Serkan Cabi, S.~M.~Ali Eslami, Oriol Vinyals, and Felix Hill.
\newblock Multimodal few-shot learning with frozen language models, 2021.

\bibitem[Venkateswara et~al.(2017)Venkateswara, Eusebio, Chakraborty, and Panchanathan]{venkateswara2017deep}
Hemanth Venkateswara, Jose Eusebio, Shayok Chakraborty, and Sethuraman Panchanathan.
\newblock Deep hashing network for unsupervised domain adaptation.
\newblock In \emph{Proceedings of the IEEE conference on computer vision and pattern recognition}, pages 5018--5027, 2017.

\bibitem[Wang et~al.(2023{\natexlab{a}})Wang, Hu, Hou, Chen, Zheng, Wang, Yang, Huang, Ye, Geng, et~al.]{wang2023robustness}
Jindong Wang, Xixu Hu, Wenxin Hou, Hao Chen, Runkai Zheng, Yidong Wang, Linyi Yang, Haojun Huang, Wei Ye, Xiubo Geng, et~al.
\newblock On the robustness of chatgpt: An adversarial and out-of-distribution perspective.
\newblock \emph{arXiv preprint arXiv:2302.12095}, 2023{\natexlab{a}}.

\bibitem[Wang et~al.(2023{\natexlab{b}})Wang, Lv, Yu, Hong, Qi, Wang, Ji, Yang, Zhao, Song, et~al.]{wang2023cogvlm}
Weihan Wang, Qingsong Lv, Wenmeng Yu, Wenyi Hong, Ji~Qi, Yan Wang, Junhui Ji, Zhuoyi Yang, Lei Zhao, Xixuan Song, et~al.
\newblock Cogvlm: Visual expert for pretrained language models.
\newblock \emph{arXiv preprint arXiv:2311.03079}, 2023{\natexlab{b}}.

\bibitem[Wang et~al.(2023{\natexlab{c}})Wang, Wang, Cao, Shen, and Huang]{wang2023images}
Xinlong Wang, Wen Wang, Yue Cao, Chunhua Shen, and Tiejun Huang.
\newblock Images speak in images: A generalist painter for in-context visual learning, 2023{\natexlab{c}}.

\bibitem[Wang et~al.(2023{\natexlab{d}})Wang, Zhang, Cao, Wang, Shen, and Huang]{wang2023seggpt}
Xinlong Wang, Xiaosong Zhang, Yue Cao, Wen Wang, Chunhua Shen, and Tiejun Huang.
\newblock Seggpt: Segmenting everything in context, 2023{\natexlab{d}}.

\bibitem[Wei et~al.(2022)Wei, Tay, Bommasani, Raffel, Zoph, and et~al.]{wei2022emergent}
Jason Wei, Yi~Tay, Rishi Bommasani, Colin Raffel, Barret Zoph, and et~al.
\newblock Emergent abilities of large language models, 2022.

\bibitem[Wei et~al.(2023{\natexlab{a}})Wei, Wang, Schuurmans, Bosma, Ichter, and et~al.]{wei2023chainofthought}
Jason Wei, Xuezhi Wang, Dale Schuurmans, Maarten Bosma, Brian Ichter, and et~al.
\newblock Chain-of-thought prompting elicits reasoning in large language models, 2023{\natexlab{a}}.

\bibitem[Wei et~al.(2023{\natexlab{b}})Wei, Wei, Tay, Tran, Webson, Lu, and et~al.]{wei2023larger}
Jerry Wei, Jason Wei, Yi~Tay, Dustin Tran, Albert Webson, Yifeng Lu, and et~al.
\newblock Larger language models do in-context learning differently, 2023{\natexlab{b}}.

\bibitem[Wortsman et~al.(2022)Wortsman, Ilharco, Gadre, Roelofs, Gontijo-Lopes, and et~al.]{Wortsman2022ModelSA}
Mitchell Wortsman, Gabriel Ilharco, Samir~Yitzhak Gadre, Rebecca Roelofs, Raphael Gontijo-Lopes, and et~al.
\newblock Model soups: averaging weights of multiple fine-tuned models improves accuracy without increasing inference time.
\newblock \emph{ArXiv}, abs/2203.05482, 2022.
\newblock URL \url{https://api.semanticscholar.org/CorpusID:247362886}.

\bibitem[Xu et~al.(2021{\natexlab{a}})Xu, Zhang, Zhang, Wang, and Tian]{Xu2021AFF}
Qinwei Xu, Ruipeng Zhang, Ya~Zhang, Yanfeng Wang, and Qi~Tian.
\newblock A fourier-based framework for domain generalization.
\newblock \emph{2021 IEEE/CVF Conference on Computer Vision and Pattern Recognition (CVPR)}, pages 14378--14387, 2021{\natexlab{a}}.
\newblock URL \url{https://api.semanticscholar.org/CorpusID:235166569}.

\bibitem[Xu et~al.(2021{\natexlab{b}})Xu, Cui, Shen, Zhang, and Zhang]{Xu2021WhySL}
Renzhe Xu, Peng Cui, Zheyan Shen, Xingxuan Zhang, and Tong Zhang.
\newblock Why stable learning works? a theory of covariate shift generalization.
\newblock \emph{ArXiv}, abs/2111.02355, 2021{\natexlab{b}}.
\newblock URL \url{https://api.semanticscholar.org/CorpusID:241035549}.

\bibitem[Yang et~al.(2023)Yang, Li, Lin, Wang, Lin, and et~al.]{yang2023dawn}
Zhengyuan Yang, Linjie Li, Kevin Lin, Jianfeng Wang, Chung-Ching Lin, and et~al.
\newblock The dawn of lmms: Preliminary explorations with gpt-4v (ision).
\newblock \emph{arXiv preprint arXiv:2309.17421}, 9\penalty0 (1):\penalty0 1, 2023.

\bibitem[Ye et~al.(2023)Ye, Xu, Xu, Ye, Yan, Zhou, Wang, Hu, Shi, Shi, et~al.]{ye2023mplug}
Qinghao Ye, Haiyang Xu, Guohai Xu, Jiabo Ye, Ming Yan, Yiyang Zhou, Junyang Wang, Anwen Hu, Pengcheng Shi, Yaya Shi, et~al.
\newblock mplug-owl: Modularization empowers large language models with multimodality.
\newblock \emph{arXiv preprint arXiv:2304.14178}, 2023.

\bibitem[Yuan et~al.(2023)Yuan, Chen, Cui, Gao, Zou, Cheng, Ji, Liu, and Sun]{yuan2023revisiting}
Lifan Yuan, Yangyi Chen, Ganqu Cui, Hongcheng Gao, Fangyuan Zou, Xingyi Cheng, Heng Ji, Zhiyuan Liu, and Maosong Sun.
\newblock Revisiting out-of-distribution robustness in nlp: Benchmark, analysis, and llms evaluations.
\newblock \emph{arXiv preprint arXiv:2306.04618}, 2023.

\bibitem[Zhang et~al.(2021{\natexlab{a}})Zhang, Zhang, Liu, Weller, Scholkopf, and Xing]{Zhang2021TowardsPD}
Hanlin Zhang, Yi-Fan Zhang, Weiyang Liu, Adrian Weller, Bernhard Scholkopf, and Eric~P. Xing.
\newblock Towards principled disentanglement for domain generalization.
\newblock \emph{2022 IEEE/CVF Conference on Computer Vision and Pattern Recognition (CVPR)}, pages 8014--8024, 2021{\natexlab{a}}.
\newblock URL \url{https://api.semanticscholar.org/CorpusID:244715116}.

\bibitem[Zhang et~al.(2023{\natexlab{a}})Zhang, Wang, Cao, Xu, Ouyang, Zhao, Ding, Zhang, Duan, Yan, et~al.]{zhang2023internlm}
Pan Zhang, Xiaoyi Dong~Bin Wang, Yuhang Cao, Chao Xu, Linke Ouyang, Zhiyuan Zhao, Shuangrui Ding, Songyang Zhang, Haodong Duan, Hang Yan, et~al.
\newblock Internlm-xcomposer: A vision-language large model for advanced text-image comprehension and composition.
\newblock \emph{arXiv preprint arXiv:2309.15112}, 2023{\natexlab{a}}.

\bibitem[Zhang et~al.(2023{\natexlab{b}})Zhang, Han, Liu, Gao, Zhou, and et~al.]{zhang2023llamaadapter}
Renrui Zhang, Jiaming Han, Chris Liu, Peng Gao, Aojun Zhou, and et~al.
\newblock Llama-adapter: Efficient fine-tuning of language models with zero-init attention, 2023{\natexlab{b}}.

\bibitem[Zhang et~al.(2022)Zhang, Roller, Goyal, Artetxe, Chen, and et~al.]{zhang2022opt}
Susan Zhang, Stephen Roller, Naman Goyal, Mikel Artetxe, Moya Chen, and et~al.
\newblock Opt: Open pre-trained transformer language models, 2022.

\bibitem[Zhang et~al.(2021{\natexlab{b}})Zhang, Cui, Xu, Zhou, He, and Shen]{Zhang2021DeepSL}
Xingxuan Zhang, Peng Cui, Renzhe Xu, Linjun Zhou, Yue He, and Zheyan Shen.
\newblock Deep stable learning for out-of-distribution generalization.
\newblock \emph{2021 IEEE/CVF Conference on Computer Vision and Pattern Recognition (CVPR)}, pages 5368--5378, 2021{\natexlab{b}}.
\newblock URL \url{https://api.semanticscholar.org/CorpusID:233289530}.

\bibitem[Zhang et~al.(2023{\natexlab{c}})Zhang, He, Xu, Yu, Shen, and Cui]{zhang2023nico++}
Xingxuan Zhang, Yue He, Renzhe Xu, Han Yu, Zheyan Shen, and Peng Cui.
\newblock Nico++: Towards better benchmarking for domain generalization.
\newblock In \emph{Proceedings of the IEEE/CVF Conference on Computer Vision and Pattern Recognition}, pages 16036--16047, 2023{\natexlab{c}}.

\bibitem[Zhang et~al.(2023{\natexlab{d}})Zhang, Xu, Yu, Dong, Tian, and Cui]{Zhang_2023_ICCV}
Xingxuan Zhang, Renzhe Xu, Han Yu, Yancheng Dong, Pengfei Tian, and Peng Cui.
\newblock Flatness-aware minimization for domain generalization.
\newblock In \emph{Proceedings of the IEEE/CVF International Conference on Computer Vision (ICCV)}, pages 5189--5202, October 2023{\natexlab{d}}.

\bibitem[Zhang et~al.(2023{\natexlab{e}})Zhang, Xu, Yu, Zou, and Cui]{zhang2023gradient}
Xingxuan Zhang, Renzhe Xu, Han Yu, Hao Zou, and Peng Cui.
\newblock Gradient norm aware minimization seeks first-order flatness and improves generalization.
\newblock In \emph{Proceedings of the IEEE/CVF Conference on Computer Vision and Pattern Recognition}, pages 20247--20257, 2023{\natexlab{e}}.

\bibitem[Zhang et~al.(2023{\natexlab{f}})Zhang, Wang, Liang, Zhang, Wang, and et~al.]{zhang2023free}
YiFan Zhang, Xue Wang, Jian Liang, Zhang Zhang, Liang Wang, and et~al.
\newblock Free lunch for domain adversarial training: Environment label smoothing, 2023{\natexlab{f}}.

\bibitem[Zhang et~al.(2023{\natexlab{g}})Zhang, Zhou, and Liu]{zhang2023makes}
Yuanhan Zhang, Kaiyang Zhou, and Ziwei Liu.
\newblock What makes good examples for visual in-context learning?
\newblock \emph{arXiv preprint arXiv:2301.13670}, 2023{\natexlab{g}}.

\bibitem[Zhou et~al.(2020{\natexlab{a}})Zhou, Yang, Hospedales, and Xiang]{Zhou2020DeepDI}
Kaiyang Zhou, Yongxin Yang, Timothy~M. Hospedales, and Tao Xiang.
\newblock Deep domain-adversarial image generation for domain generalisation.
\newblock \emph{ArXiv}, abs/2003.06054, 2020{\natexlab{a}}.
\newblock URL \url{https://api.semanticscholar.org/CorpusID:212718063}.

\bibitem[Zhou et~al.(2020{\natexlab{b}})Zhou, Yang, Hospedales, and Xiang]{Zhou2020LearningTG}
Kaiyang Zhou, Yongxin Yang, Timothy~M. Hospedales, and Tao Xiang.
\newblock Learning to generate novel domains for domain generalization.
\newblock \emph{ArXiv}, abs/2007.03304, 2020{\natexlab{b}}.
\newblock URL \url{https://api.semanticscholar.org/CorpusID:220380867}.

\bibitem[Zhu et~al.(2023{\natexlab{a}})Zhu, Chen, Shen, Li, and Elhoseiny]{zhu2023minigpt}
Deyao Zhu, Jun Chen, Xiaoqian Shen, Xiang Li, and Mohamed Elhoseiny.
\newblock Minigpt-4: Enhancing vision-language understanding with advanced large language models.
\newblock \emph{arXiv preprint arXiv:2304.10592}, 2023{\natexlab{a}}.

\bibitem[Zhu et~al.(2023{\natexlab{b}})Zhu, Chen, Shen, Li, and Elhoseiny]{zhu2023minigpt4}
Deyao Zhu, Jun Chen, Xiaoqian Shen, Xiang Li, and Mohamed Elhoseiny.
\newblock Minigpt-4: Enhancing vision-language understanding with advanced large language models, 2023{\natexlab{b}}.

\end{thebibliography}
